\documentclass[sigconf]{acmart}
\AtBeginDocument{%
  }

\setcopyright{acmlicensed}
\copyrightyear{2025}
\acmYear{2025}
\acmDOI{}
\acmPrice{}
\acmISBN{}
\acmConference[DAI '25]{The Seventh International Conference on Distributed Artificial Intelligence}{Nov 21--24, 2025}{London, UK}

\title[Trajectory Q-Learning]{Is Risk-Sensitive Reinforcement Learning Properly Resolved?}

\author{Ruiwen Zhou}
\email{skyriver@sjtu.edu.cn}
\affiliation{%
  \institution{Shanghai Jiao Tong University}
  \city{Shanghai}
  \country{China}
}

\author{Minghuan Liu}
\email{minghuanliu@sjtu.edu.cn}
\affiliation{%
  \institution{Shanghai Jiao Tong University}
  \city{Shanghai}
  \country{China}
}

\author{Kan Ren}
\email{renkan@shanghaitech.edu.cn}
\affiliation{%
  \institution{ShanghaiTech University}
  \city{Shanghai}
  \country{China}
}

\author{Xufang Luo}
\email{xfluo@microsoft.com}
\affiliation{%
 \institution{Microsoft Research Asia}
  \city{Shanghai}
  \country{China}
}

\author{Weinan Zhang}
\email{wnzhang@sjtu.edu.cn}
\affiliation{%
  \institution{Shanghai Jiao Tong University}
  \city{Shanghai}
  \country{China}
}

\author{Dongsheng Li}
\email{dongsli@microsoft.com}
\affiliation{%
 \institution{Microsoft Research Asia}
  \city{Shanghai}
  \country{China}
}

\usepackage{natbib}
\usepackage{makecell}
\usepackage{mathtools}
\usepackage{amsthm}
\usepackage{pifont}
\usepackage{thmtools,thm-restate}
\usepackage{multirow}
\usepackage{multicol}
\usepackage{enumitem}
\usepackage{times}
\usepackage{bbm,bm}
\usepackage{soul}
\usepackage{url}
\usepackage{subcaption}
\usepackage{float}
\usepackage{tabularx}
\usepackage{pifont}
\usepackage{algorithm}
\usepackage{algorithmic}
\usepackage{xcolor}
\usepackage{colortbl}
\usepackage{tikz}

\newcommand{\operator}{\textsc{HR}}%

\definecolor{lightsunflower}{HTML}{FFEC67}
\definecolor{cloud}{HTML}{ecf0f1}

\usepackage{amsmath,amsfonts,bm}

\def\eqref#1{equation~\ref{#1}}

\def\1{\bm{1}}

\DeclareMathAlphabet{\mathsfit}{\encodingdefault}{\sfdefault}{m}{sl}
\SetMathAlphabet{\mathsfit}{bold}{\encodingdefault}{\sfdefault}{bx}{n}

\newcommand{\fig}[1]{Fig.~\ref{#1}}
\newcommand{\eq}[1]{Eq.~(\ref{#1})}
\newcommand{\tb}[1]{Tab.~\ref{#1}}
\newcommand{\se}[1]{Section~\ref{#1}}
\newcommand{\ap}[1]{Appendix~\ref{#1}}

\newcommand{\lm}[1]{Lemma~\ref{#1}}

\newcommand{\alg}[1]{Algo.~\ref{#1}}
\newcommand{\theo}[1]{Theorem~\ref{#1}}

\newcommand{\bbE}{\ensuremath{\mathbb{E}}} 
\newcommand{\bbR}{\ensuremath{\mathbb{R}}} 
\newcommand{\caA}{\ensuremath{\mathcal{A}}} 
\newcommand{\caS}{\ensuremath{\mathcal{S}}} 
\newcommand{\caD}{\ensuremath{\mathcal{D}}} 
\newcommand{\caL}{\ensuremath{\mathcal{L}}} 
\newcommand{\caP}{\ensuremath{\mathcal{P}}} 
\newcommand{\caT}{\ensuremath{\mathcal{T}}} 
 
\newcommand{\caZ}{\ensuremath{\mathcal{Z}}} 
\newcommand{\caH}{\ensuremath{\mathcal{H}}}

\theoremstyle{plain}
\newtheorem{theorem}{Theorem}[section]

\newtheorem{lemma}[theorem]{Lemma}

\theoremstyle{definition}

\theoremstyle{remark}

\newcommand{\minisection}[1]{\vspace{1pt}\noindent\textbf{#1}}

\begin{document}

\pagestyle{fancy}
\fancyhead{}

\begin{abstract}
Due to the nature of risk management in learning applicable policies, risk-sensitive reinforcement learning (RSRL) has been realized as an important direction. RSRL is usually achieved by learning risk-sensitive objectives characterized by various risk measures, under the framework of distributional reinforcement learning. However, it remains unclear if the distributional Bellman operator properly optimizes the RSRL objective in the sense of risk measures. In this paper, we prove that the existing RSRL methods do not achieve unbiased optimization and cannot guarantee optimality or even improvements regarding risk measures over accumulated return distributions. To remedy this issue, we further propose a novel algorithm, namely Trajectory Q-Learning (TQL), for RSRL problems with provable policy improvement towards the optimal policy. Based on our new learning architecture, we are free to introduce a general and practical implementation for different risk measures to learn disparate risk-sensitive policies. In the experiments, we verify the learnability of our algorithm and show how our method effectively achieves better performances toward risk-sensitive objectives.
\end{abstract}




\maketitle

\section{Introduction}\label{sec:intro}

Reinforcement learning (RL) has shown its success on various tasks \citep{mnih2015dqn,silver2017alphago,guan2022perfectdou}, which usually requires the agent to take enormous trial-and-error steps. However, most real-world applications are sensitive to failure and attach more importance to risk management, and thus need to turn to the help of risk-sensitive reinforcement learning (RSRL). 
Typically, RSRL can be implemented upon existing distributional RL approaches~\citep{bellemare2017c51, dabney2018iqn}. For instance, risk-sensitive actor-critic methods like \citet{urpi2021oraac,dabney2018iqn} first learn distributional critics as normal distributional RL methods, and then take various distortion risk measures representing different risk preferences as the objective for the actor (e.g., conditional value-at-risk (\texttt{CVaR})), which is computed based on critic output.

However, as we reveal in this paper, such solutions lead to biased optimization, fails to converge to an optimal solution in terms of risk-sensitive returns along the whole trajectory, and can sometimes lead to an arbitrarily bad policy. Therefore, an algorithm for unbiased optimization is desired in RSRL. Although some works~\citep{tamar2015policy,bauerle2022markov} define risk in a per-step manner and then derive valid solutions, it is still challenging to resolve RSRL problems where a policy is learned to optimize the risk measure over accumulated return distributions.

In this paper, we provide an in-depth analysis on the biased optimization issue of existing RSRL methods, conclude the reason, and present the intuition of the solution. 
Correspondingly, we propose Trajectory Q-Learning (TQL), a novel RSRL framework proven to learn the optimal policy w.r.t. various risk measures.
Specifically, TQL learns an historical value function that models the conditional distribution of accumulated returns along the whole trajectory given the trajectory history. 
We give an extensive theoretical analysis of the learning behavior of TQL, showing the convergence of policy evaluation, policy iteration, policy improvement, and (no) value iteration. Notably, we prove that the policy iteration of TQL can achieve unbiased optimization in RSRL.
To the best of our knowledge, TQL is the first algorithm that can converge to the optimal risk-sensitive policy for all kinds of distortion risk measures.
Experimentally, we verify our idea on both discrete mini-grid and continuous control tasks,
showing TQL can be practically effective for finding optimal risk-sensitive policies and outperforms existing RSRL algorithms.

\section{Preliminaries}
\subsection{Distributional RL}
We consider a Markov decision process (MDP), denoted as a tuple $\left(\mathcal{S},\mathcal{A},\caP,R,\gamma\right)$, where $\mathcal{S}$ and $\mathcal{A}$ represents the state and action space, $\caP\left(s^\prime|s,a\right)$ is the dynamics transition function and when it is deterministic we use $s'=M(s,a)$ to represent the transition. $R(s,a)$ is the (stochastic) reward function, and $\gamma$ denotes the discount factor. 
The return $Z^\pi(s,a)=\sum_{t=0}^{\infty}\gamma^t R\left(s_t,a_t\right)$ is a random variable representing the sum of the discounted rewards. 
The history $h_t=\{s_0,a_0,\cdots,s_t\}$ is state-action sequences sampled by agents in the environment, and its space is $\caH=\bigcup_t\left[\left(\prod_{i=0}^{t-1}(\caS\times\caA)\right)\times\caS\right]$.
The objective of reinforcement learning is to learn a policy to perform the action $a\sim\pi$ on a given state or history that maximizes the expected cumulative discounted reward $\bbE_\pi\left[Z^\pi\left(s,a\right)\right]$.

The optimization typically requires to compute the state-action value function $Q(s,a)=\bbE_{\pi}\left[Z^{\pi}(s,a)\right]$, which can be characterized by the Bellman operator $\mathcal{T}_B^\pi$:
\begin{equation*}
\mathcal{T}_B^\pi Q^\pi\left(s,a\right):=\mathbb{E}\left[R\left(s,a\right)\right]+\gamma\mathbb{E}_{s'\sim\caP,a'\sim\pi}\left[Q^\pi\left(s^\prime,a^\prime\right)\right]~.
\end{equation*}
The optimal policies can be obtained by learning the optimal value $Q^*=Q^{\pi^*}$ through the Bellman optimality operator $\mathcal{T}^*_B$:
\begin{equation*}
    \mathcal{T}^*_B Q\left(s,a\right)
    :=\mathbb{E}\left[R\left(s,a\right)\right]
    +\gamma\mathbb{E}_\caP[\max_{a^\prime}Q\left(s^\prime,a^\prime\right)]~.
\end{equation*}

Instead of utilizing a scalar value function $Q^{\pi}$, which can be seen as optimizing the expectation of the distribution over returns, distributional RL considers modeling the whole distribution~\cite{bellemare2017c51,dabney2018iqn}.
From a distributional perspective, we regard $Z^\pi\sim\caZ$ as a mapping from state-action pairs to distributions over returns, named the value distribution. 
Analogous to traditional RL, the goal is seeking a policy that maximizes the expected return over trajectories:
\begin{equation}\label{eqn:conventional-rl-obj}
\pi^*\in\mathop{\arg\max}_\pi~\bbE_{\pi}\left[Z^\pi\left(s,a\right)\right]~.
\end{equation}
Similarly, we can define a distributional Bellman operator $\mathcal{T^\pi}$ that estimates the return distribution $Z^\pi$ 
\begin{equation}\label{eq:dis-bellman}
\mathcal{T^\pi}Z\left(s,a\right)\mathop{:=}\limits^D R\left(s,a\right)
    +\gamma Z\left(s^\prime,a^\prime\right)~,
\end{equation}
where $A\mathop{:=}\limits^D B$ means that random variables $A$ and $B$ follows the same distribution, $s^\prime\sim\caP\left(\cdot|s,a\right)$ and $a^\prime\sim \pi\left(\cdot|s^\prime\right)$.
Correspondingly, the distributional Bellman optimality operator is
\begin{equation}\label{eq:dis-opt-bellman}
    \mathcal{T^*}Z\left(s,a\right)\mathop{:=}\limits^D R\left(s,a\right)
    +\gamma Z\left(s^\prime,\mathop{\arg\max}\limits_{a^\prime\in\mathcal{A}}\bbE\left[Z\left(s^\prime,a^\prime\right)\right]\right)~,
\end{equation}



\subsection{Distortion Risk Measure and RSRL}
\label{sec:rsrl}
As a type of risk measure, a \emph{distortion risk measure}~\citep{wang1996distorion} $\beta$ for a random variable $X$ with the cumulative distribution function (CDF) $F_X(x)$ is defined as $\beta[X]=\int_{-\infty}^\infty x~\frac{\partial}{\partial x}\left(h_\beta\circ F_X\right)(x)~{\rm d}x$,
where $h_\beta:[0,1]\rightarrow[0,1]$, called a distortion function, is a continuous non-decreasing function that transforms the CDF of $X$ into $\left(h_\beta\circ F_X\right)(x)$, and thus the probability density function (PDF) $f(x) = \frac{\rm d}{{\rm d}x}F_X(x)$ into $h_\beta^\prime(\tau)\big|_{\tau=F_X(x)}\cdot f(x)$.
Intuitively, a distortion function distorts the probability density of a random variable to assign more weight on either higher-risk or lower-risk events. 
For example, \texttt{mean} and \texttt{CVaR} are the most commonly used distortion risk measure, whose corresponding distortion functions are:
\begin{align}
    h_{\rm mean}(\tau) &= \tau~,\\
    h_{\eta{\rm -CVaR}}(\tau) &= \left\{
    \begin{array}{cl}
        \tau/\eta, & 0\le\tau\le\eta\\
        0, & \eta<\tau\le 1\\
    \end{array}
    \right.~.
\end{align}
For readers unfamiliar with distortion risk measures, we list some typical examples and their definitions in \ap{ap:exp-vs-distortion}.
Thereafter, risk-sensitive reinforcement learning is natural to combine various distortion risk measures with distributional RL to achieve a risk-sensitive behavior. In the sequel, a risk-sensitive optimal policy with distortion risk measure $\beta$ can be defined as a deterministic policy $\pi_\beta^*$ by the risk-sensitive return over random variable $s_0\sim\rho_0$ representing the initial state:
\begin{equation}\label{eqn:risk-sensitive-policy}
\pi_\beta^*\in\mathop{\arg\max}_\pi~\bbE_{s_0\sim\rho_0}\left[\beta\left[Z^\pi\left(s_0,a_0\right)\right]\right]~.
\end{equation}
We call \eq{eqn:risk-sensitive-policy} the RSRL objective, as it seeks a policy that maximizes the risk measure of accumulated return over whole trajectory given the initial state distribution. Such a formulation was initially implemented in \cite{dabney2018iqn}, by directly changing the objective to risk measures computed from the value distribution, which is shown to be problematic later in this paper. Some other works define and optimize risk in a per-step manner~\cite{tamar2015policy,bauerle2022markov}, but we only focuses on the RSRL objective, as it is the natural risk-sensitive extension of RL. For readers interested in per-step risk definition, we give a brief introduction in \ap{ap:dynamic-vs-static}.

\subsection{Metrics for Convergence}
In distributional RL, since the value function is modeled as a distribution, researchers utilize a maximal form of the Wasserstein metric to establish the convergence of the distributional Bellman operators~\cite{bellemare2017c51,dabney2018qrdqn}:
\begin{equation*}
    \bar{d}_p(Z_1, Z_2) := \sup_{x, a} d_p(Z_1(x,a), Z_2(x,a))~,
\end{equation*}
where $Z_1, Z_2 \in \caZ$ are two value distributions and $\caZ$ denotes the space of value distributions with bounded moments. 
The $p$-Wasserstein distance $d_p$ is the $L_p$ metric on inverse CDF, i.e., quantile functions \cite{muller1997integral}, which is defined as an optimal transport metric for random variables $U$ and $V$ with quantile functions $F_U^{-1}$ and $F_V^{-1}$ respectively:
\begin{equation*}
    d_p(U, V) = \left( \int_0^1 |F_U^{-1}(\omega) - F_V^{-1}(\omega)|^p d\omega \right)^{1/p}~.
\end{equation*}
This can be realized as the minimal cost of transporting mass to make the two distributions identical.

Requiring the distributional Bellman operators to converge in the metric of $\bar{d}_p$ indicates that we must match the value distribution. While in policy evaluation the distributional Bellman operator $\caT^{\pi}$ (\eq{eq:dis-bellman}) is shown to be a contraction in $p$-Wasserstein, in the control setting proving the distributional Bellman optimality
operator $\caT^{*}$ (\eq{eq:dis-opt-bellman}) is hard (see \citet{bellemare2017c51} for more details) and is not always necessary in practical cases. Instead, we may only need to achieve convergence in the sense of distributional statistics or measures. 
For example, we only require the learned value distribution to have the same \texttt{mean} of the optimal value distribution so that the policy learns to achieve the optimal return expectation, or we match a risk measure (like \texttt{CVaR}) of the optimal value distribution to learn a policy that achieves the optimal risk preference of the return distribution. 
As these measures upon value distributions are real functions w.r.t. states and actions, the convergence of distributional Bellman operators only need to lie in the infinity norm, a $L_\infty$ metric:
\begin{equation*}
    \|f_1 - f_2\|_\infty=\sup_{x, a} \|f_1(x,a) - f_2(x,a)\|~.
\end{equation*}
\section{Mismatch in RSRL Optimization}
Although the RSRL objective \eq{eqn:risk-sensitive-policy} seems reasonable, existing dynamic programming (DP) style algorithms does not optimize \eq{eqn:risk-sensitive-policy} properly, as we will reveal in this section.

\subsection{Dynamic Programming Fails in RSRL}

Recalling the RL objective \eq{eqn:conventional-rl-obj} or considering setting $\beta$ as \texttt{mean} in the RSRL objective \eq{eqn:risk-sensitive-policy}, we can optimize $\pi$ by Bellman equation in a dynamic programming style following the distributional Bellman optimality operator \eq{eq:dis-opt-bellman}, i.e., there is a deterministic policy that maximizes the return at every single step for a given return distribution $Z$:
\begin{equation}\label{eqn:conventional-mean-policy}
    \pi_\texttt{mean}(s)\in\mathop{\arg\max}_{a\in\mathcal{A}} \bbE\left[Z\left(s,a\right)\right]~.
\end{equation}
And the distributional Bellman optimality operator is equivalent to:
\begin{equation}\label{eq:dis-opt-bellman-2}
    \begin{small}\mathcal{T^*}Z\left(s,a\right)\mathop{:=}\limits^D R\left(s,a\right)
    +\gamma Z\left(s^\prime,\pi_{\texttt{mean}}(s^\prime)\right),~s'\sim\caP~.
    \end{small}
\end{equation}

Although \citet{bellemare2017c51} have shown that $\mathcal{T^*}$ itself is not a contraction in $\bar{d}_p$ such that it cannot be used for finding the optimal value distribution, we can realize $\mathcal{T^*}$ as a ``contraction" in $L_\infty$ from the perspective of \texttt{mean}, which induces a point-wise convergence. In other words, the \texttt{mean} of the value distribution $\bbE[Z]$ will converge to the \texttt{mean} of the value distribution $\bbE[Z^*]$.

\begin{lemma}[Value iteration theorem \citep{bellemare2017c51}]
\label{lemma:value-iteration}
Recursively applying the distributional Bellman optimality operator $Z_{k+1}=\caT^*Z_{k}$ on arbitrary value distribution $Z_0$ solves the objective \eq{eqn:risk-sensitive-policy} when $\beta$ is exactly \texttt{mean} where the optimal policy is obtained via \eq{eqn:conventional-mean-policy}, and for $Z_1, Z_2 \in \caZ$, we have:
\begin{equation}\label{eqn:bellman-optimality-contraction}
    \|\bbE[\caT^*Z_{1}]-\bbE[\caT^*Z_{2}]\|_\infty\le\gamma\|\bbE[Z_{1}]-\bbE[Z_{2}]\|_\infty~,
\end{equation}
and in particular $\bbE[Z_k]\rightarrow \bbE[Z^*]$ exponentially quickly.
\end{lemma}
The proof is just the proof of value iteration and Lemma 4 in \citep{bellemare2017c51}. For completeness, we include it in \ap{ap:value-iteration}. 
In the context of distributional RL, we can explain it as the \texttt{mean} of value will converge to the \texttt{mean} of optimal value.
Motivated by and simply resembling \eq{eqn:conventional-mean-policy}, previous implementation like \citet{dabney2018iqn} and \citet{urpi2021oraac} optimized a risk-sensitive policy:
\begin{equation}\label{eqn:conventional-risk-sensitive-policy}
	\pi_\beta(s)\in\mathop{\arg\max}_{a\in\mathcal{A}} \beta\left[Z\left(s,a\right)\right]~.
\end{equation}
From a practical perspective, this can be easily achieved by only a few modifications to distributional RL algorithms towards any given distortion risk measure $\beta$, which implies a dynamic programming style updating following a risk-sensitive Bellman optimality operator $\caT_{\beta}^*$ w.r.t. risk measure $\beta$:
\begin{equation}\label{eq:dis-opt-bellman-rs-sto}
    \caT_{\beta}^*Z(s,a)\mathop{:=}\limits^D R(s,a)+\gamma Z(s',a')~,
\end{equation}
where $s'\sim\caP(\cdot|s,a)$ and $a'\sim\pi_\beta(\cdot|s')$.
Note that the optimal risk-sensitive policy defined in \eq{eqn:conventional-risk-sensitive-policy} is completely different from \eq{eqn:risk-sensitive-policy}. The key difference is that \eq{eqn:conventional-risk-sensitive-policy} tends to maximize the risk measure everywhere inside an MDP, while \eq{eqn:risk-sensitive-policy} only requires finding a policy that can maximize the risk measure of trajectories started from the initial state $s_0$. 
Since \texttt{mean} has the linearly additive property that
\begin{equation*}
    \bbE[X]\ge\bbE[Y]\Longrightarrow\forall Z,~\bbE[X+Z]\ge\bbE[Y+Z]~,
\end{equation*}
given a state we have the same optimal future trajectories, which have the largest expected future return, for all different history trajectories. In this way, the optimal sub-structure for DP style algorithms exists and the two definitions of RSRL objective are equivalent.
However, as we will discuss next, other distortion risk measures have no linearly additive property, hence DP style algorithms have no optimal sub-structure, and \eq{eqn:conventional-risk-sensitive-policy} is not even a valid definition under distortion risk measures other than \texttt{mean}. Trying to optimize \eq{eqn:conventional-risk-sensitive-policy} will lead to the divergence in RSRL optimization and may result in an arbitrarily bad policy. Therefore, \eq{eqn:risk-sensitive-policy} is the only choice of RSRL objective which our solution is derived to optimize.

\subsection{$\mathcal{T}_{\beta}^*$ Leads to Biased Optimization}\label{subsec:biased-obj}

To show why $\caT_\beta^*$ leads to biased optimization, we first provide an analysis that optimizing towards the Bellman optimality operator $\caT_\beta^*$ w.r.t. risk measure $\beta$ does not converge at all, \textit{i.e.}, there is no contraction property for $\mathcal{T}_{\beta}^*$.
For simplicity and starting from the easiest case, in the rest of this paper, we assume deterministic dynamics, i.e., instead of $s'\sim \caP(\cdot|s,a)$, we simply consider $s'=M(s,a)$.

\paragraph{$\mathcal{T}_{\beta}^*$ is not contraction except $\beta$ is \texttt{mean}.} 
We already know that $\caT_{\beta}^*$ is not a contraction in $\bar{d}_p$,
but different from \lm{lemma:value-iteration}, even from the perspective of the risk measure $\beta$ if $\beta$ is not \texttt{mean}, it still cannot be realized as a ``contraction'' in $L_{\infty}$; in other words, the risk measure of the value distribution $\beta [Z]$ is not guaranteed to converge to $\beta$ of the optimal value distribution $\beta [Z^*]$. Thus, $\caT_{\beta}^*$ does not help to find an optimal solution to solve the RSRL objective \eq{eqn:risk-sensitive-policy}.
\begin{theorem}\label{theo:no-contraction-theorem}
Recursively applying risk-sensitive Bellman optimality operator $\caT_\beta^*$ w.r.t. risk measure $\beta$ does not solve the RSRL objective \eq{eqn:risk-sensitive-policy} and $\beta [Z_k]$ is not guaranteed to converge to $\beta [Z^*]$ if $\beta$ is not \texttt{mean}.
\end{theorem}
The formal proof can be referred to \ap{ap:no-contraction-theorem}. The above theorem of no contraction indicates that optimizing towards $\caT_\beta$ may lead to arbitrarily worse solutions than the optimal solution of the RSRL objective \eq{eqn:risk-sensitive-policy}. 

\begin{figure}
\begin{center}
\begin{minipage}{0.44\columnwidth}
\hspace{-7em} 
\includegraphics[width=0.7\textwidth]{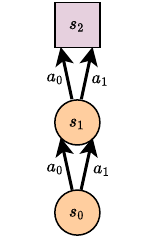}
\end{minipage}
\hspace{-11em} 
\begin{minipage}{0.55\columnwidth}
\resizebox{1.4\textwidth}{!}{
$
\begin{array}[c]{l|cccc}
\multicolumn{1}{c}{\beta=\texttt{CVaR}(\eta=0.1)} & \multicolumn{1}{c}{s_0,a_0} & \multicolumn{1}{c}{s_0,a_1} & \multicolumn{1}{c}{s_1, a_0} & \multicolumn{1}{c}{s_1, a_1} \\
\midrule
R & \makecell[c]{\left\{\begin{matrix}
 & 100, p=0.9 \\
 & -10, p=0.1 \\
\end{matrix}\right.} & -5 & \makecell[c]{\left\{\begin{matrix}
 & 100, p=0.9 \\
 & -10, p=0.1 \\
\end{matrix}\right.} & -5 \\
\midrule 
Z^* & \makecell[c]{\left\{\begin{matrix}
 & 200, p=0.81 \\
 & 90, p=0.18 \\
 & -20, p=0.01 \\
\end{matrix}\right.} & \makecell[c]{\left\{\begin{matrix}
 & 95, p=0.9 \\
 & -15, p=0.1 \\
\end{matrix}\right.} & \makecell[c]{\left\{\begin{matrix}
 & 100, p=0.9 \\
 & -10, p=0.1 \\
\end{matrix}\right.} & -5 \\
\midrule
\beta(Z^*) & 7.9 & -15 & -10 & -5 \\
\midrule
\midrule
Z & 0 & 0 & \makecell[c]{\left\{\begin{matrix}
 & 100, p=0.9 \\
 & -10, p=0.1 \\
\end{matrix}\right.} & -5 \\
\midrule
\beta(Z) & 0 & 0 & -10 & -5 \\
\midrule
\caT_{\beta}^* Z & \cellcolor{lightsunflower} \makecell[c]{\left\{\begin{matrix}
 & 95, p=0.9 \\
 & -15, p=0.1 \\
\end{matrix}\right.} & \cellcolor{lightsunflower} -10 & \makecell[c]{\left\{\begin{matrix}
 & 100, p=0.9 \\
 & -10, p=0.1 \\
\end{matrix}\right.} & -5 \\
\midrule
\beta(\caT_{\beta}^* Z) &\cellcolor{lightsunflower} -15 & \cellcolor{lightsunflower} -10 & \cellcolor{lightsunflower} -10 & -5 \\
\bottomrule
\end{array}
$}
\end{minipage}
\end{center}
\vspace{-10pt}
\caption{Undiscounted 3-state MDP for which the optimality operator $\caT_\beta^*$ does not converge and obtain non-optimal result. 
We highlight the entries that are incorrectly updated.}
\vspace{-12pt}
\label{fig:3state-mdp}
\end{figure}

To better understand how the problem occurs, we consider a naive contradictive example on a 3-state MDP (\fig{fig:3state-mdp}, left), 
where the agents have a constant reward of -5 for conducting $a_1$ and a binomial reward for $a_0$, as shown in the first row of \fig{fig:3state-mdp} (right). 
In such context, we optimize towards $\beta=\texttt{CVaR}(\eta=0.1)$. 
Now consider the initial value estimation $Z$ to be accurate at $s_1$ (\fig{fig:3state-mdp}, right). We list the value of $Z$ and its corresponding risk measure $\beta[Z]$; the results $\caT_{\beta}^* Z$ when updating $Z$ on $s_0$ using $\caT_\beta^*$ and its corresponding risk measure.
In this case, when updating $Z(s_0)$, $Z(s_1)$ will always indicate to use $a_1$, although this can lead to a worse risk measure evaluated along the whole trajectory starting from $s_0$, and prevent the agent from finding the optimality, \textit{i.e.}, applying $a_0$ at both states.

\paragraph{History return distribution matters.}
Now we dive deeper into this example. Different from the optimal trajectory, if we have chosen $a_1$ at $s_0$, then $a_1$ will be the optimal action at $s_1$ in terms of \eq{eqn:risk-sensitive-policy}. However, this does not holds when we have chosen $a_0$ at $s_0$, hence we know that the optimal action at the same state can be different given different history trajectories. That is to say, \emph{Markovian (dynamic programming) algorithms are never capable of handling an RSRL objective as \eq{eqn:risk-sensitive-policy}, and some history information must be included in policy and value function if one wants an optimal solution for such an objective.} Back to algorithmic level, we may easily notice that the reason for different optimal actions given different history trajectory is the different history return distributions, which is not taken into account in the (Markovian) risk-sensitive Bellman optimality operator $\caT_\beta^*$ as in \eq{eq:dis-opt-bellman-rs-sto}. Therefore, $\caT_\beta^*$ diverges and leads to sub-optimal policies, and we should take into account the return distribution along the past trajectory starting from $s_0$ at every state to get the optimal policy in terms of \eq{eqn:risk-sensitive-policy}.

Note that very few distortion risk measures (e.g. \texttt{CVaR}) can rely on sufficient statistics to achieve Markovian policy optimization \cite{bauerle2011markov,hau2023on}, while it is impossible to figure out such statistics for infinite distortion risk measures. Therefore, we will next present a method to model history return distribution directly.
\section{Solving RSRL}
\label{sec:tql}
To remedy the biased optimization issue of bellman-style update, we propose a novel algorithm that lies in a non-Markovian formulation without dynamic programming style optimization.

As we pointed out before, the key problem that leads the risk-sensitive optimal Bellman operator $\caT_{\beta}^{*}$ into biased optimization is that \textit{the risk measure over future return distributions cannot be maximized everywhere inside an MDP}.
Thereafter, \textit{the dynamic programming style optimization that only utilizes the information forward, i.e., in the future, does not help to find the policy that maximizes the risk measures along the whole trajectories as defined in \eq{eqn:risk-sensitive-policy}}.
Thus,
when we compute the value distribution at certain states, we must include information backward, \textit{i.e.}, in the past, to help with modeling the risk measure along the whole trajectory.
This motivates us to model the history-action value distribution $Z^\pi(h_t,a_t)\sim\caZ$, called \emph{historical return distribution}, instead of the state-action value distribution $Z^\pi(s_t,a_t)$, along with a history-based (non-Markovian) policy $A\sim\pi(\cdot|h)$:
\begin{equation}\label{eqn:new-z}
\begin{aligned}
    &Z^\pi(h_t,a)\triangleq\sum_{i=0}^{t-1} \gamma^{i} R(s_{i},a_{i})+\gamma^{t} Z^\pi(\{s_t\},a)\\
    =&\sum_{i=0}^{t} \gamma^{i} R(s_{i},a_{i})+\gamma^{t+1} Z^\pi\left(\{s_{t+1}\},A_{t+1})\right)~,
\end{aligned}
\end{equation}
where $A_{t+1}\sim\pi(\cdot|h_{t+1}), s_{t+1}=M(s_t, a_t), h_t=\{s_0,a_0,\cdots,s_t\}\in\caH$ denotes the history sequence that happened before reaching (including) state $s_t$. 
Therefore, the history-action value $Z^\pi(h_t,a)$ just records the discounted return of the whole trajectory given history $h_t$ backward and moves forward following policy $\pi$. Note that the policy is now Markovian under the history-based MDP, \textit{i.e.}, the policy gives action only based on the current history.

\subsection{Policy Evaluation}
Similar to Bellman operators, we now define a new type of operator, named the history-relied (\operator) operator, that defines the principle of updating the history-action value. 
\begin{equation}\label{eq:hr-operator}
\caT^\pi_h Z(h_t,a)\mathop{:=}\limits^D R_{0:t}+\gamma^{t+1} Z(\{s_{t+1}\},A_{t+1})~,
\end{equation}
where $A_{t+1}\sim\pi(\cdot|h_{t+1}), s_{t+1}=M(s_t, a_t), R_{0:t}=\sum_{i=0}^{t}\gamma^i r_i$ is the discounted return accumulated before the timestep $t$.
To continue, we show our first theoretical result, that the policy evaluation with \operator~operator converges in the metric of $\bar{d}_p$.

\begin{theorem}[Policy Evaluation for $\caT^\pi_h$]\label{theo:pe-contraction}
$\caT^\pi_h:\caZ\rightarrow\caZ$ is a $\gamma$-contraction in the metric of the maximum form of $p$-Wasserstein distance $\bar{d}_p$.
\end{theorem}
The proof of Theorem \ref{theo:pe-contraction} can be referred to \ap{ap:pe-contraction-theorem}. Using Theorem \ref{theo:pe-contraction} and combining Banach's fixed point theorem, we can conclude that $\caT^\pi_{h}$ has a unique fixed point. By inspection, this fixed point must be $Z^\pi$ as defined in \eq{eqn:new-z} since $\caT^{\pi}_h Z^\pi = Z^\pi$.

\subsection{Policy Improvement and (No) Value Iteration}
So far, we have considered the value distribution of a fixed policy $\pi$ and the convergence of policy evaluation. Now let's turn to the control setting and find out the optimal value distribution and its corresponding policy under the risk-sensitive context.

In the form above, we want to find the optimal risk-sensitive policy that maximizes the risk measure over the whole trajectory given the initial state distribution as defined in \eq{eqn:risk-sensitive-policy}, which is equivalent,
\begin{equation}\label{eqn:risk-sensitive-policy-form2}
\pi^*(h)\in\mathop{\arg\max}_{a\in\mathcal{A}}~\bbE_{h\sim\caH,a\sim\pi}\left[\beta\left[Z^\pi(h,a)\right]\right]~.
\end{equation}
Suppose $\caH$ and $\caA$ are both finite, the solution of \eq{eqn:risk-sensitive-policy-form2} will always exist (but may not be unique). 
Denoting the optimal risk-sensitive policy set is $\Pi^*$, where $\forall \pi_1^*,\pi_2^* \in \Pi^*$, we have their return distribution $\beta[Z_1^*]=\beta[Z_2^*]$ and they must satisfy the risk-sensitive \operator~optimality equation:
\begin{align}
\beta[Z^*_1(h_t,a)] &= \beta\left[R_{0:t}+\gamma^{t+1} Z^*_2(\{s_{t+1}\},a_{t+1}^*)\right]\label{eq:hr-optimal-equation}\\
a_{t+1}^* &\in \mathop{\arg\max}_{a\in\caA}~\beta\left[Z^*_2(h_{t+1},a)\right]\label{eq:hr-optimal-policy} ~.
\end{align}
We can prove \eq{eq:hr-optimal-equation} is also sufficient for \eq{eqn:risk-sensitive-policy-form2}, see \ap{ap:optimal-operator}. Hereby, we define the risk-sensitive \operator~optimality operator $\caT^{*}_{h,\beta}$:
\begin{equation}\label{eq:hr-optimal-operator}
\begin{aligned}
    &\caT^{*}_{h,\beta} Z(h_t,a) \leftarrow R_{0:t}+\gamma^{t+1} Z(\{s_{t+1}\},a_{t+1})\\
    &a_{t+1}=\pi^\prime(h_{t+1})=\mathop{\arg\max}_{a\in\caA}~\beta\left[Z(h_{t+1},a)\right]~,
\end{aligned}
\end{equation}
where the policy is obtained by deterministically maximizing the history-action value under risk measure $\beta$. And \eq{eq:hr-optimal-equation} implies some ``fixed" points for \eq{eqn:risk-sensitive-policy} or \eq{eqn:risk-sensitive-policy-form2} from the perspective of risk measure $\beta$ for $\caT^{*}_{h,\beta}$.

Correspondingly, we can present our second theoretical result, that the policy improvement under \operator~optimality operator is also guaranteed to converge into the risk-sensitive optimal policy.
\begin{theorem}[Policy Improvement for $\caT^*_{h,\beta}$]\label{theo:risk-policy-improvement}
For two deterministic policies $\pi$ and $\pi^\prime$, if $\pi^\prime$ is obtained by $\caT^{*}_{h,\beta}$:
\begin{equation} 
\pi^\prime(h_{t})\in\mathop{\arg\max}_{a\in\mathcal{A}}~\beta\left[Z^\pi(h_{t},a)\right]\nonumber~,
\end{equation}
then the following inequality holds
\begin{equation}
\beta\left[Z^\pi(h_{t},\pi(h_{t}))\right]\le
\beta\left[Z^{\pi^\prime}(h_{t},\pi^\prime(h_{t}))\right]\nonumber~.
\end{equation}
\end{theorem}
The formal proof can be referred to \ap{ap:risk-policy-improvement-theorem}.
When the new greedy policy $\pi^\prime$, is as good as, but not better than, the old policy $\pi$ in the sense of risk measures, we have that:
\begin{equation}
\begin{aligned}
\beta[Z^{\pi}(h_t,a_t)] &= \beta\left[\caT^*_{h,\beta} Z^{\pi}(h_t,a_t)\right] \nonumber~,
\end{aligned}
\end{equation}
Unfolding the right side, we get:
\begin{equation}
\begin{aligned}
\beta[Z^{\pi}(h_t,a_t)] &= \beta\left[R_{0:t}+\gamma^{t+1}Z^{\pi}(\{s_{t+1}\},a_{t+1})\right]\\
a_{t+1}=\pi^\prime(h_{t+1})&\in\mathop{\arg\max}_{a\in\caA}~\beta\left[Z(h_{t+1},a)\right]~,
\end{aligned}
\end{equation}
which is exactly the risk-sensitive HR optimality equation~\eq{eq:hr-optimal-equation}. Therefore, we conclude that utilizing $\caT_{h,\beta}^*$ for policy improvement will give us a strictly better policy except when the original policy is already optimal.

In the sequel, we understand that if the optimal solution of \eq{eqn:risk-sensitive-policy-form2} exists, there exists at least a sequence of distributional value function $\{Z_0, Z_1, \cdots, Z_n, Z_1^*, \cdots, Z^*_k\}$ induced by the sequence of policy $\{\pi_0, \pi_1, \cdots, \pi_n, \pi^*_1, \cdots, \pi^*_k\}$ such that $\beta[Z_1]\le\beta[Z_2]\le\cdots\le\beta[Z_n]\le\beta[Z^*_1]=\cdots=\beta[Z^*_k]$. However, starting from an arbitrary $Z$ (which may not correspond to any policy), it is non-trivial to prove $\caT^{*}_{h,\beta}$ converges to $\beta[Z^*_i]$.
\begin{theorem}
\label{theo:risk-value-iteration}
For $Z_1, Z_2 \in \caZ$,
\operator~optimality operator $\caT^{*}_{h,\beta}$ has the following property:
\begin{equation}\label{eqn:beta-nonexpensive}
    \|\beta[\caT^{*}_{h,\beta}Z_{1}]-\beta[\caT^{*}_{h,\beta}Z_{2}]\|_\infty\le\|\beta[Z_{1}]-\beta[Z_{2}]\|_\infty~,
\end{equation}
\end{theorem}

The proof is in \ap{ap:risk-value-iteration}.
Theorem \ref{theo:risk-value-iteration} told us that the value iteration for $\caT^{*}_{h,\beta}$ may not converge. Specifically, our proposed \operator~operator can be realized as a ``nonexpensive mapping" from the perspective of risk measure $\beta$ in $L_\infty$. 
For our cases of limited spaces, we might expect there exists some ``fixed" point $Z^*$, and the best we can hope is a pointwise convergence such that $\beta Z$ converges to $\beta Z^*$ after recursively applying \operator~optimality operator $\caT^{*}_{h,\beta}$ w.r.t. risk measure $\beta$. However, from \theo{theo:risk-value-iteration}, we know that $\beta Z_n$ is not assured to be converged to $\beta Z^*$ at any speed, hence the starting from arbitrary value distribution $Z_0$, $\caT^{*}_{h,\beta}$ does not necessarily solve the RSRL objective \eq{eqn:risk-sensitive-policy}. As a result, $\beta Z_n$ may possibly fall on a sphere around $Z^*$.

\subsection{Trajectory Q-Learning}
As discussed above, by estimating the historical return distribution and improving the policy accordingly, we can now derive our practical RSRL algorithm, namely Trajectory Q-Learning (TQL).  Representing the policy $\pi$, the historical value function $Q$ as neural networks parameterized by $\phi$ and $\theta$ respectively, and denoting the historical return distribution approximated by critics as
\begin{equation}\label{eqn:z}
    Z_\theta\left(h,a\right)=\frac{1}{N}\sum_{j=0}^{N-1}~{\rm Dirac}\left[
    Q_\theta\left(h,a;\tau_j\right)\right]~,
\end{equation}
where $\tau_j\in[0,1]$ refer to the quantile, we optimize the following loss functions:
\begin{align}
    J_\pi\left(\phi\right)=&~\beta\left[Z_\theta\left(h,a\right)\right]~,\label{eqn:loss-pi}\\
    J_Q\left(\theta\right)=&~
    \bbE_{a^\prime\sim\pi(h_{t+1});~\tau,\tau^\prime\sim U([0,1])}\Big[\rho_\tau^\kappa\Big(R_{0:t}+\gamma^{t+1} \nonumber\\
    &\bar{Q}_{\theta^\prime}(\{s_{t+1}\},a^\prime;\tau^\prime)
    -Q_\theta\left(h_t,a_t;\tau\right)\Big)\Big]~,\label{eqn:loss-Q-before}
\end{align}
where $\rho_\tau^\kappa$ denotes the quantile Huber loss (see \ap{ap:qr-loss} for details). In \eq{eqn:loss-Q-before}, $\bar{Q}_{\theta^\prime}(\{s_{t+1}\},a^\prime;\tau^\prime)$ resembles the target $Q$ in mean-based Markovian distributional RL \citet{bellemare2017c51}. Note that $s_{t+1}=M(s_t,a_t)$ can lie entirely outside the initial state distribution $\rho_0$, and thus the estimation of this target $Q$ can be arbitrarily inaccurate. Therefore, observing that $Z(\{s^\prime\},\cdot)$ is just a normal state-based value function \citep{dabney2018iqn}, we model $\bar{Q}_{\theta^\prime}(\{s_{t+1}\},a^\prime;\tau^\prime)$ with an extra Markovian value function as $Q_{\psi}(s^\prime,a^\prime;\tau)$, and update $Q_\psi$ by
\begin{equation}
\begin{aligned}
    J_Q\left(\psi\right)&=\bbE_{a^\prime\sim\pi;~\tau_i,\tau_j^\prime\sim U([0,1])}\Big[\rho_\tau^\kappa\Big(r\left(s,a\right)\\
    &\quad+\gamma \bar{Q}_{\psi^\prime}\left(s^\prime,a^\prime;\tau_j^\prime\right)
    -Q_\psi\left(s,a;\tau_i\right)\Big)\Big]~.\label{eqn:loss-Q-markovian}
\end{aligned}
\end{equation}
This promotes more accurate estimation of target $\bar{Q}$ and helps with the overall performance of TQL. Correspondingly, \eq{eqn:loss-Q-before} should be modified as
\begin{align}
    J_Q\left(\theta\right)=&~
    \bbE_{a^\prime\sim\pi(h_{t+1});~\tau,\tau^\prime\sim U([0,1])}\Big[\rho_\tau^\kappa\Big(R_{0:t}+\gamma^{t+1} \nonumber\\
    &\bar{Q}_{\psi^\prime}(s_{t+1},a^\prime;\tau^\prime)
    -Q_\theta\left(h_t,a_t;\tau\right)\Big)\Big]~.\label{eqn:loss-Q}
\end{align}

To help readers have a clear understanding of the actor and critics, we summarize their roles, update methods, and connections in \tb{tab:functions}.
\begin{table}[htbp]
    \centering
    \vspace{-12pt}
    \caption{A brief summarization of $\pi_\phi$, $Z_\theta$, and $Z_\psi$}
    \vspace{-6pt}
    \resizebox{\columnwidth}{!}{
    \begin{tabular}{cp{1.75cm}p{2.5cm}p{3cm}}
        \Xhline{1.5pt}
         & \multicolumn{1}{c}{$\pi_\phi$} & \multicolumn{1}{c}{$Z_\theta$} & \multicolumn{1}{c}{$Z_\psi$} \\
        \hline
        \textbf{Role} & The actor. & The history-action value function. & The state-action value function (as target). \\
        \textbf{Update} & Maximize $Z_\theta$. & Minimize \eq{eqn:loss-Q} with $Z_\psi$ as target. & Minimize \eq{eqn:loss-Q-markovian} with itself as target. \\
        \Xhline{1.5pt}
    \end{tabular}
    }
    \vspace{-6pt}
    \label{tab:functions}
\end{table}

In total, the algorithm learns a policy $\pi_\phi$, a history-based value function $Z_\theta$, and a Markovian value function $Z_\psi$. At each timestep, $Z_\psi$ and $Z_\theta$ are updated according to \eq{eqn:loss-Q-markovian} and \eq{eqn:loss-Q}, and the policy $\pi_\phi$ is optimized with \eq{eqn:loss-pi}. For discrete control, we can omit $\phi$ and implement $\pi$ by taking argmax from $\beta[Z(h,a)]$. We list the step-by-step algorithm in \alg{alg:tql-disc} (discrete) and \alg{alg:tql-cont} (continuous).
\section{Related Work}
\subsection{Distributional Reinforcement Learning}
Distributional RL considers the uncertainty by modeling the return distribution, enabling risk-sensitive policy learning. \citet{bellemare2017c51} first studied the distributional perspective on RL and proposed C51, which approximates the return distribution with a categorical over fixed intervals.
\citet{dabney2018qrdqn} proposed QR-DQN, turning to learning the critic as quantile functions and using quantile regression to minimize the Wasserstein distance between the predicted and the target distribution. 
\citet{dabney2018iqn} further proposed IQN, improving QR-DQN by quantile sampling and other techniques, which further investigate risk-sensitive learning upon various distortion risk measures.
\citet{rowland2019edrl} found the abuse of statistics and samples in \citet{dabney2018qrdqn}, and thus derived an imputation strategy to recover the return distribution from its statistics.

\subsection{Risk in Reinforcement Learning}
Risk management in RL towards real-world applications can be roughly divided into two categories, i.e., safe and constrained RL and distributional risk-sensitive RL.
Safe and constrained RL formulates the risk as some kind of constraint to the policy optimization problem. For instance, \citet{achiam2017cpo} proposed a Lagrangian method which provides a theoretical bound on cost function while optimizing the policy; \citet{dalal2018safelayer} built a safe layer to revise the action given by an unconstrained policy; \citet{chow2018lyapunov} used the Lyapunov approach to systematically transform dynamic programming and RL algorithms into their safe counterparts.

When the form of risks is either too complex or the constraints are hard to be explicitly defined, safe RL algorithms can be challenging to learn. In that case, distributional RL provides a way to utilize risk measures upon the return distributions for risk-sensitive learning. Among them, \citet{tang2019wcpg} modeled the return distribution via its mean and variance and then learned an actor optimizing the \texttt{CVaR} of the return distribution; \citet{keramati2020cvar} proposed a novel optimistic version of the distributional Bellman operator that moves probability mass from the lower to the upper tail of the return distribution for sample-efficient learning of optimal policies in terms of \texttt{CVaR}; \citet{ma2020dsac} modified SAC~\citep{haarnoja2018sac} with distributional critics and discussed its application to risk-sensitive learning; \citet{urpi2021oraac} proposed their offline risk-averse learning scheme based on IQN~\citep{dabney2018iqn} and BCQ~\citep{fujimoto2019bcq}; \citet{ma2021codac} proposed CODAC, which adapts distributional RL to the offline setting by penalizing the predicted quantiles of the return for out-of-distribution actions; recently, \citep{lim2022distributional} proposed a solution to resolve a similar issue specifically in optimizing policies towards \texttt{CVaR}, and we note that specifically for \texttt{VaR} and \texttt{CVaR} there are non-distributional solutions with history-dependent statistics \citep{bauerle2011markov,chow2014cvar,chow2015risk} which can guarantee the optimality. However, our TQL is different in that it is designed not for a specific risk measure but generally optimal for all kinds of risk measures. A more detailed comparison between our work and the most relevant distributional algorithm \citep{lim2022distributional} can be referred to \ap{ap:compare}.

Several works \citep{tamar2015policy,bauerle2022markov} utilize dynamic risk measures as their objective, which considers per-step risk instead of the static (trajectory-wise) risk in this paper. Dynamic risk has the advantage of time-consistency, but can be hard to estimate practically and short-sighted due to per-step optimization. Entropic risk measures \citep{fei2020entropic,hau2023entropic} can also be optimized efficiently, but they are less commonly applied in practice. For completeness, we present a more detailed discussion on dynamic and static risk in \ap{ap:dynamic-vs-static}.

\begin{figure}[tbp]
    \centering
    \begin{subfigure}[b]{0.14\textwidth}
        \includegraphics[width=\textwidth]{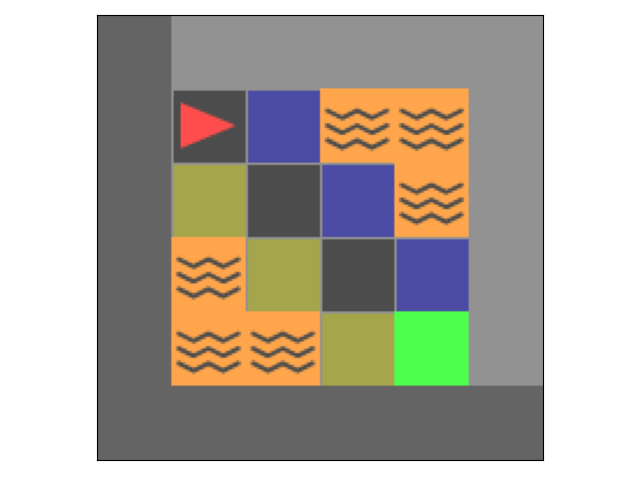}
        \vspace{-14pt}
        \caption{Mini-Grid Task.}
        \label{fig:grid-env}
    \end{subfigure}
    \begin{subfigure}[b]{0.16\textwidth}
        \includegraphics[width=\textwidth]{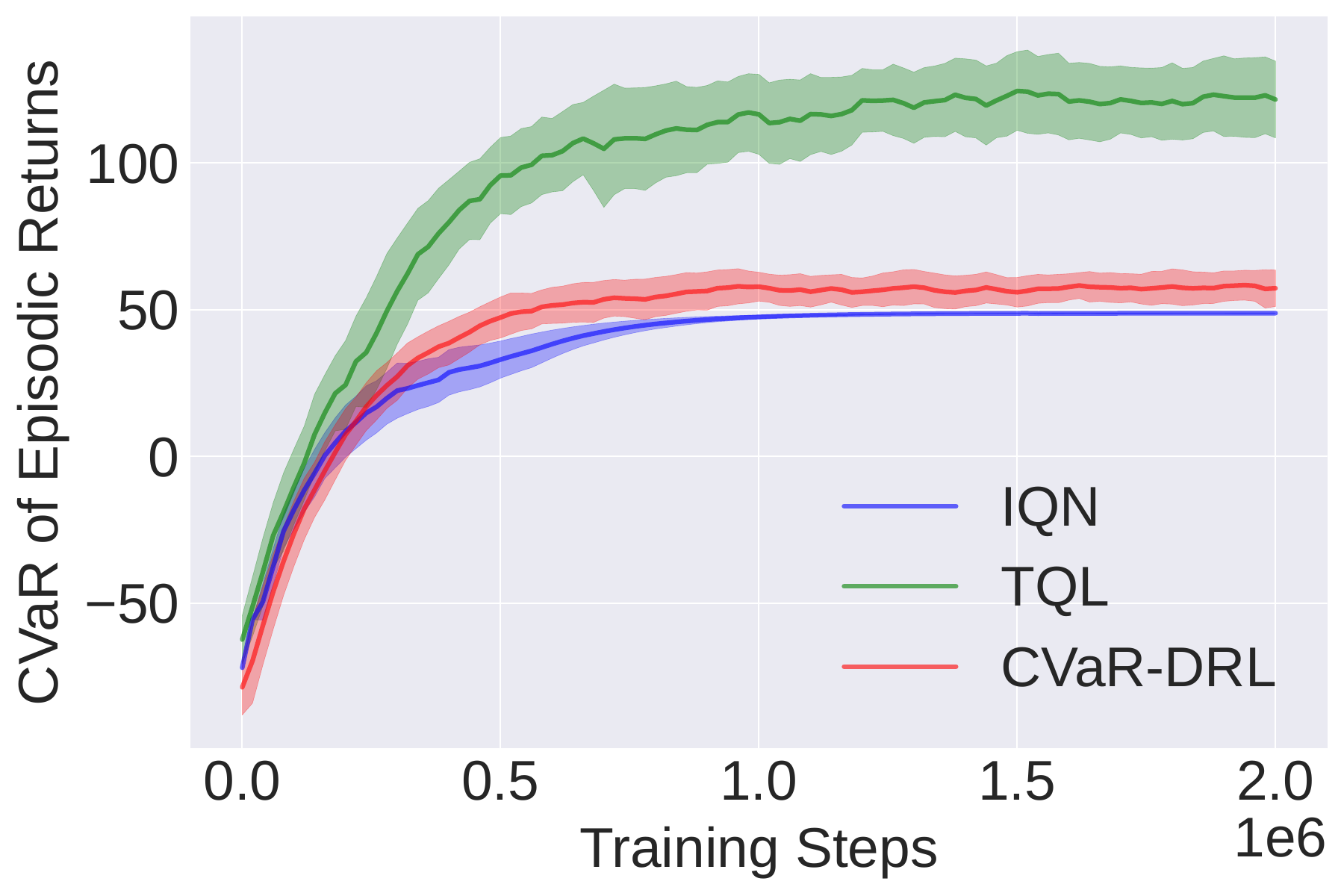}
        \vspace{-14pt}
        \caption{Learning Curves.}
        \label{fig:grid-learning}
    \end{subfigure}
    \begin{subfigure}[b]{0.16\textwidth}
        \includegraphics[width=\textwidth]{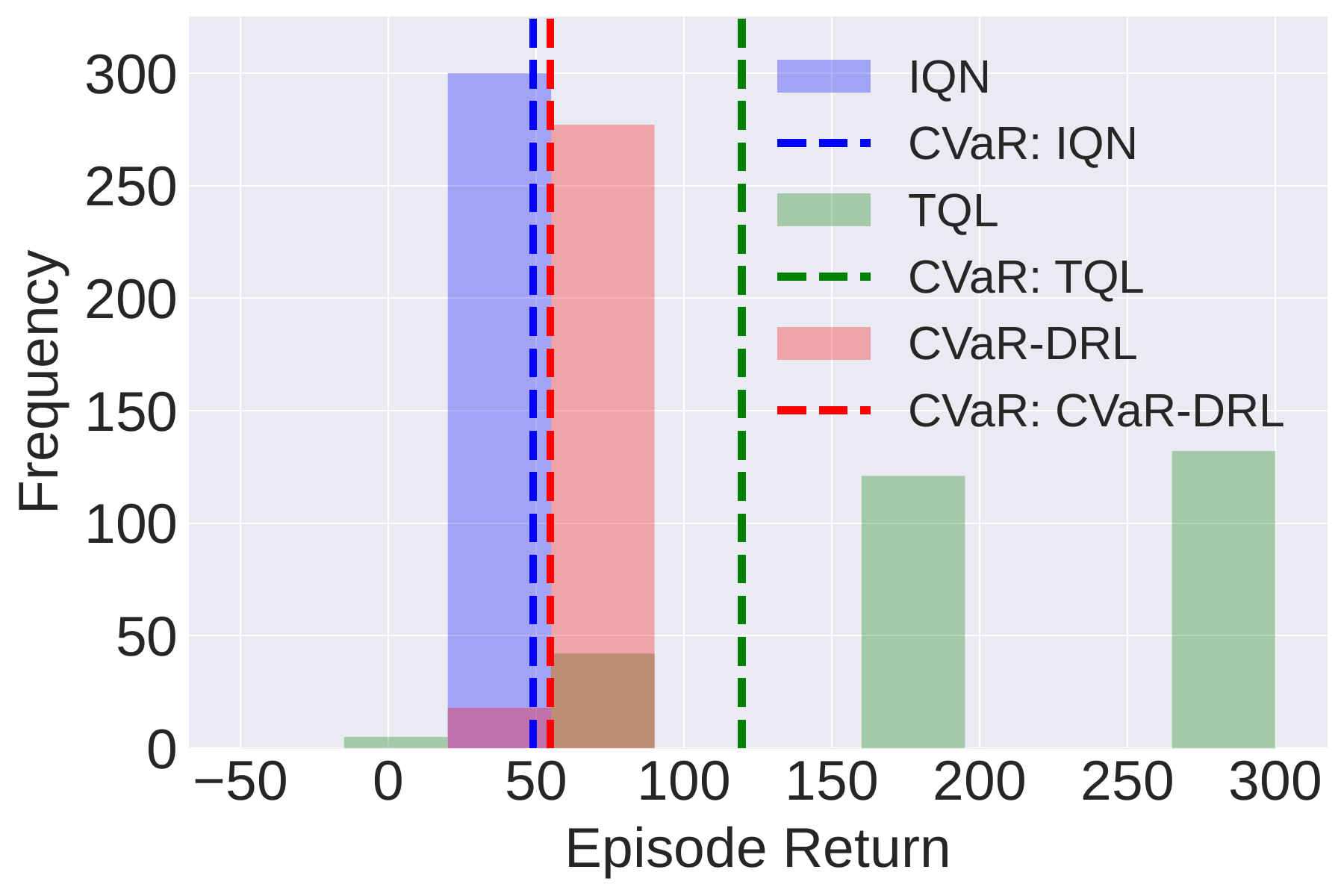}
        \vspace{-14pt}
        \caption{Final Returns.}
        \label{fig:grid-final-return}
    \end{subfigure}
\vspace{-8pt}
\caption{Mini-grid experiments designed for learning \texttt{CVaR} objective. (a) Illustration of risky mini-grid environment. The agent starts at the upper left corner of the grid (red triangle), and reaches the bottom right green grid to end the episode. At each timestep, the agent receives a constant penalty of $-2$. The yellow grids give a $+100$ bonus with the probability of $p=0.75$ and $0$ with the probability of $p=0.25$, while the blue grids always give a reward of $+20$. Each yellow or blue grid can give its reward only once. The orange grids have a heavy penalty of $-100$ to avoid the agent from going there. (b-c) Experiment results on the task: (b) Vanilla IQN quickly converges to a sub-optimal solution; CVaR-DRL discovers a slightly better policy; TQL finds the optimal policy. (c) The return distributions of vanilla IQN and CVaR-DRL are more conservative, while that of TQL results in a higher \texttt{CVaR}.}
\vspace{-14pt}
\end{figure}
\section{Experiments}
In this section, we design a series of experiments aimed to seek out: \textbf{RQ1}: Can our proposed TQL fit the ground-truth risk measures? \textbf{RQ2}: Can TQL find the optimal risk-sensitive policy or achieve better overall performance? 

\begin{figure*}[tbp]
    \centering
    \begin{subfigure}[b]{0.19\textwidth}
        \includegraphics[width=\textwidth]{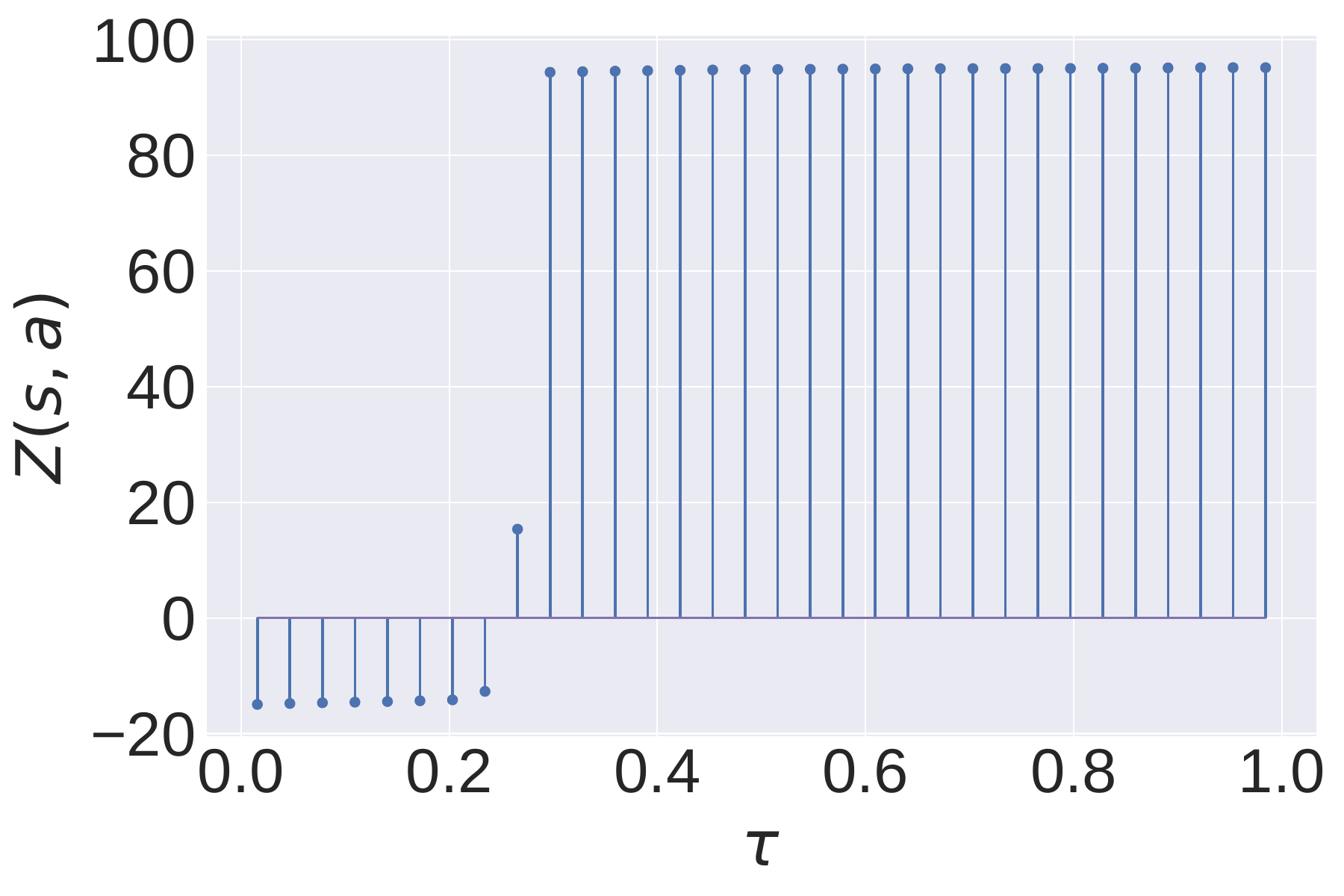}
        \vspace{-14pt}
        \caption*{(a) $s=0,~a=0$}
        \label{fig:simple-z-0-0}
    \end{subfigure}
    \begin{subfigure}[b]{0.19\textwidth}
        \includegraphics[width=\textwidth]{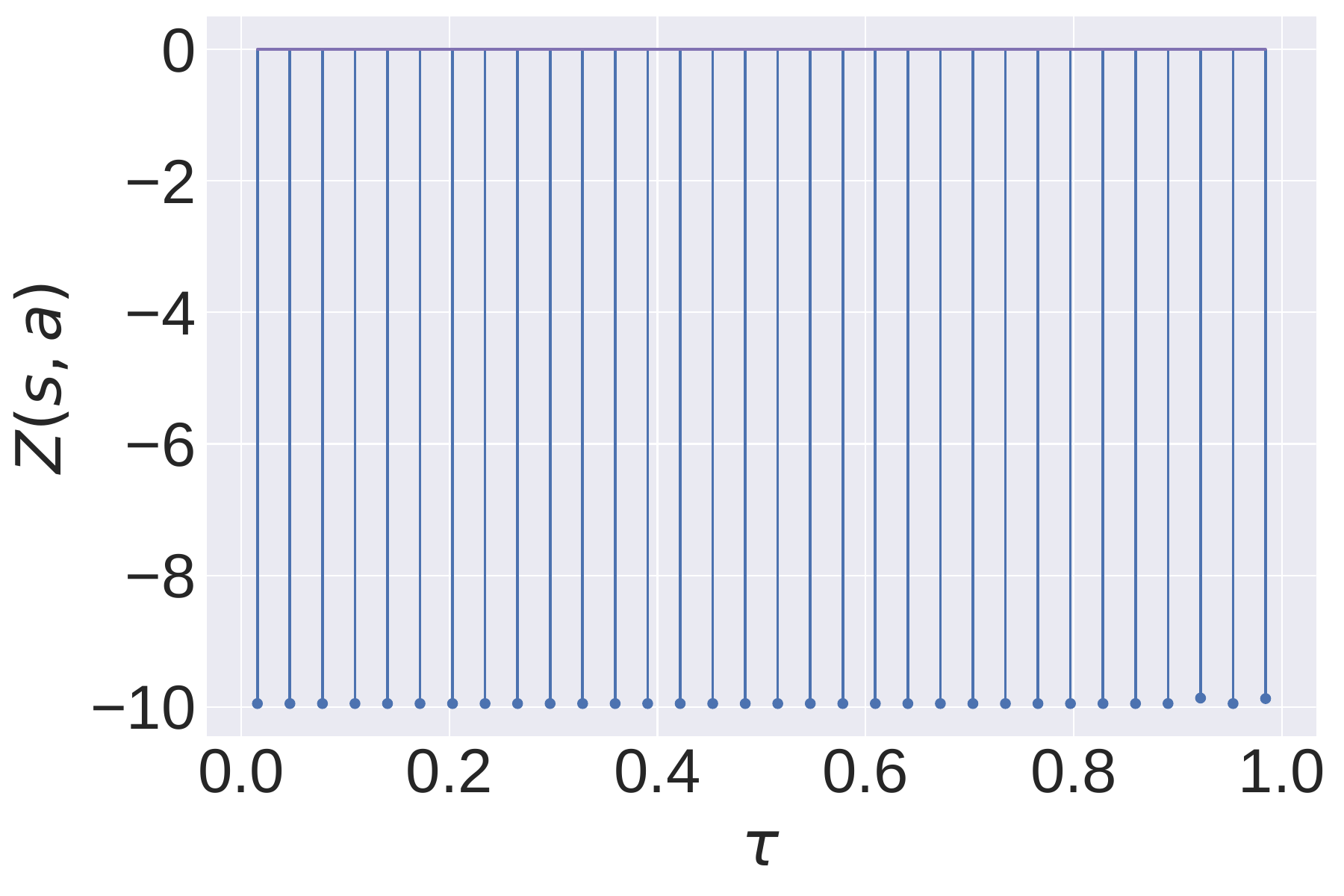}
        \vspace{-14pt}
        \caption*{(b) $s=0,~a=1$}
        \label{fig:simple-z-0-1}
    \end{subfigure}
    ~\tikz{\draw[-,black, densely dashed, thick](0,-1.80) -- (0,0.90);}~
    \begin{subfigure}[b]{0.19\textwidth}
        \includegraphics[width=\textwidth]{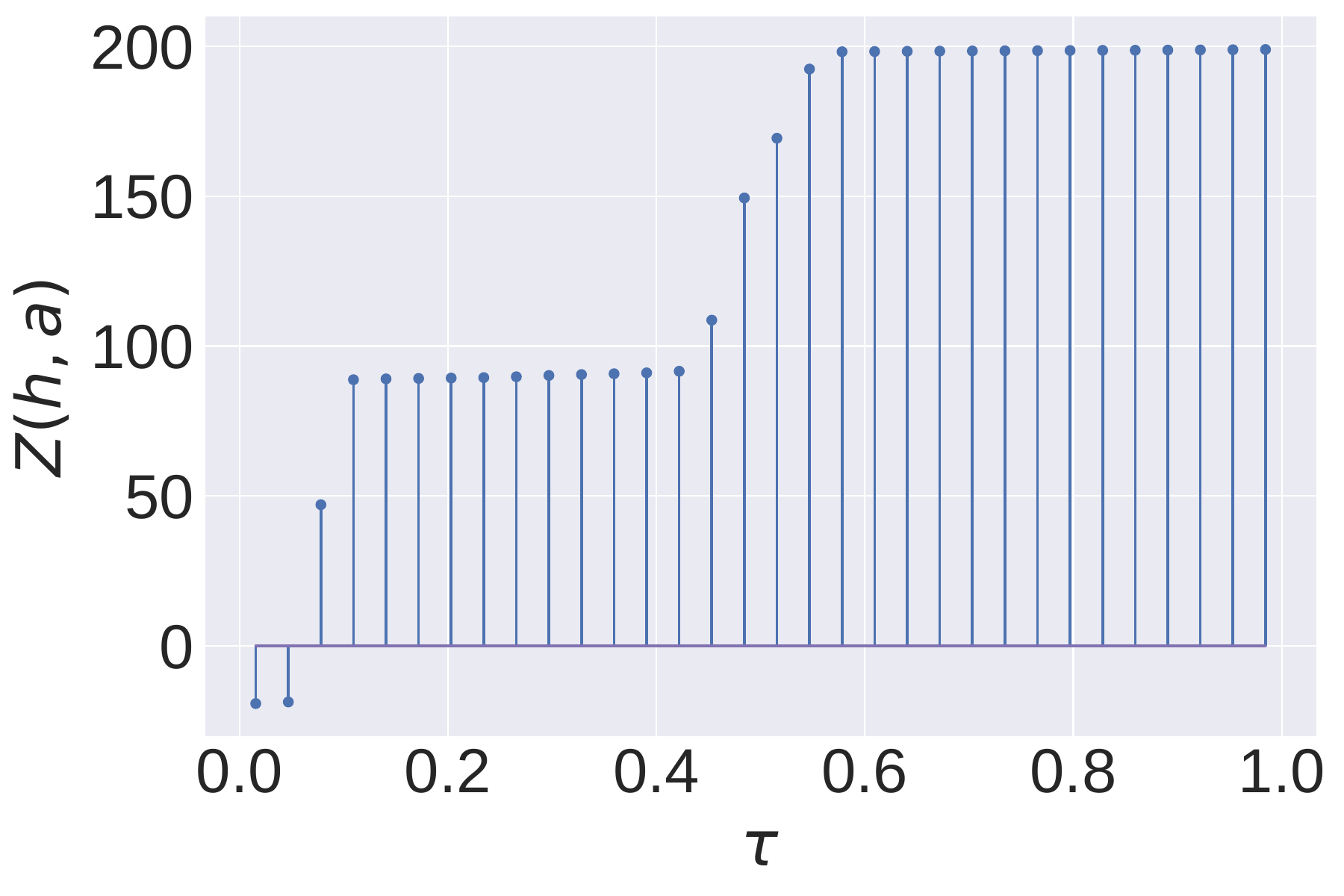}
        \vspace{-14pt}
        \caption*{(e) $h=\{0\},~a=0$}
        \label{fig:simple-z-tilde-0}
    \end{subfigure}
    \begin{subfigure}[b]{0.19\textwidth}
        \includegraphics[width=\textwidth]{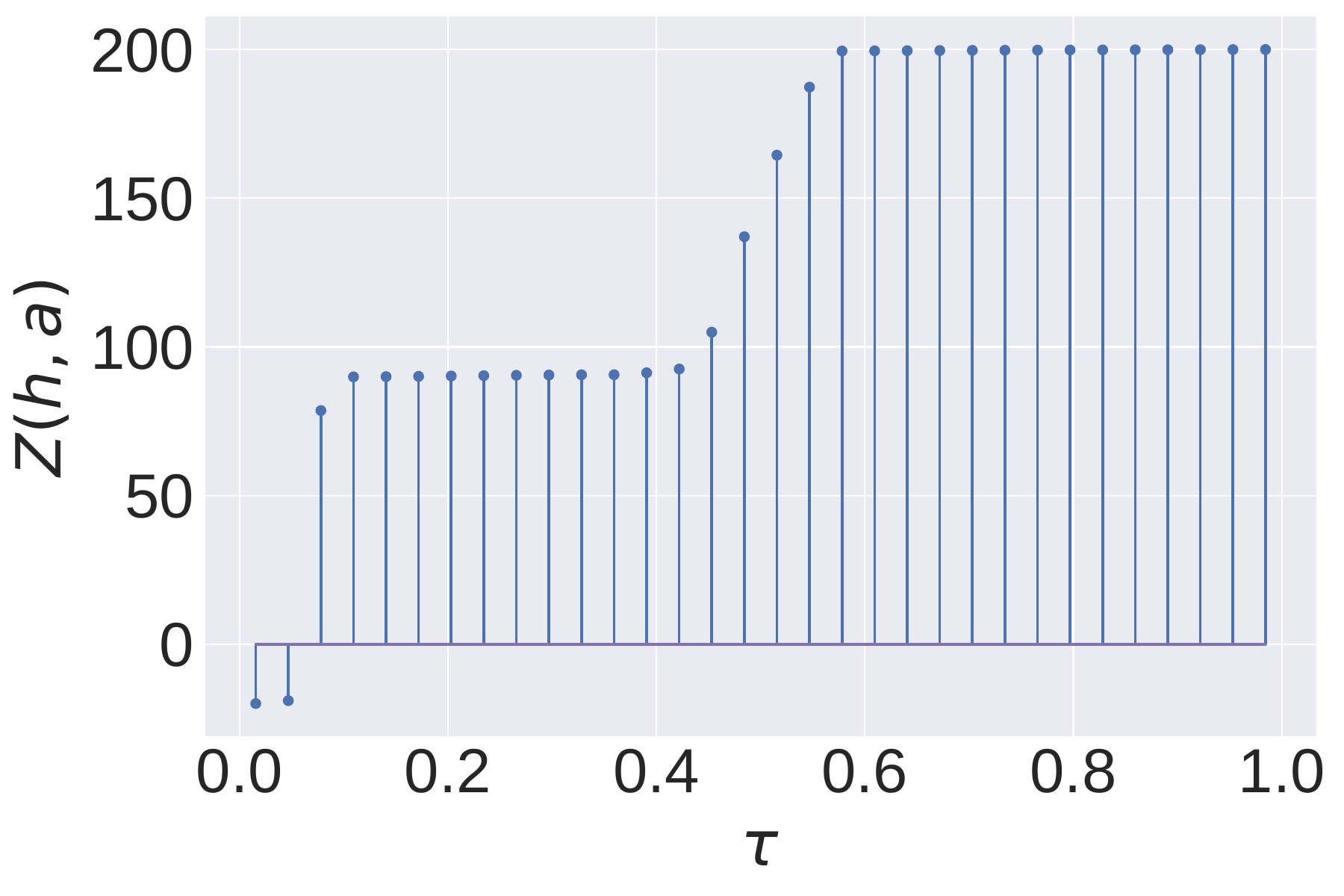}
        \vspace{-14pt}
        \caption*{(f) $h=\{0,0,1\},~a=0$}
        \label{fig:simple-z-tilde-0-0}
    \end{subfigure}
    \begin{subfigure}[b]{0.19\textwidth}
        \includegraphics[width=\textwidth]{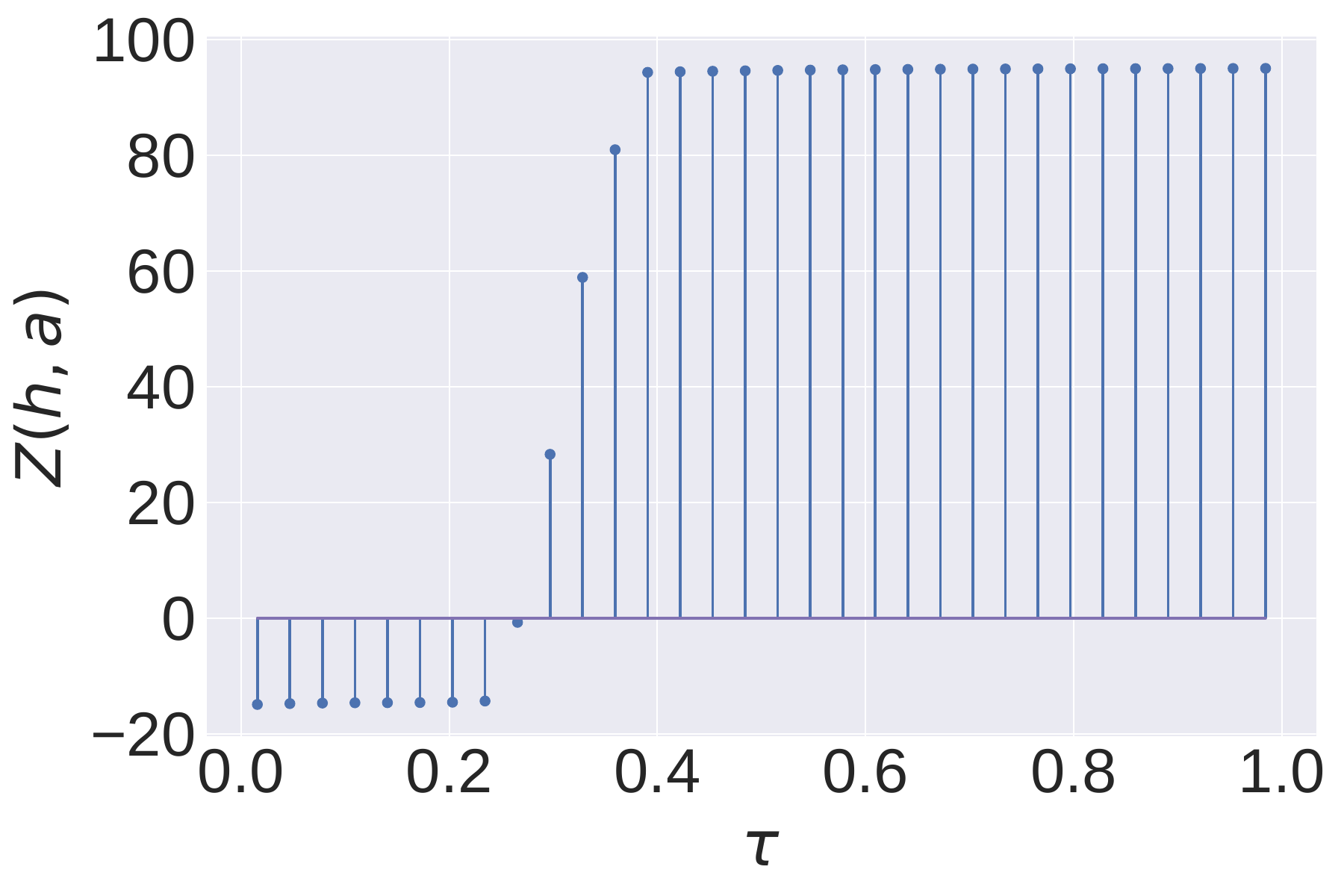}
        \vspace{-14pt}
        \caption*{(g) $h=\{0,0,1\},~a=1$}
        \label{fig:simple-z-tilde-0-1}
    \end{subfigure}\\
    \begin{subfigure}[b]{0.19\textwidth}
        \includegraphics[width=\textwidth]{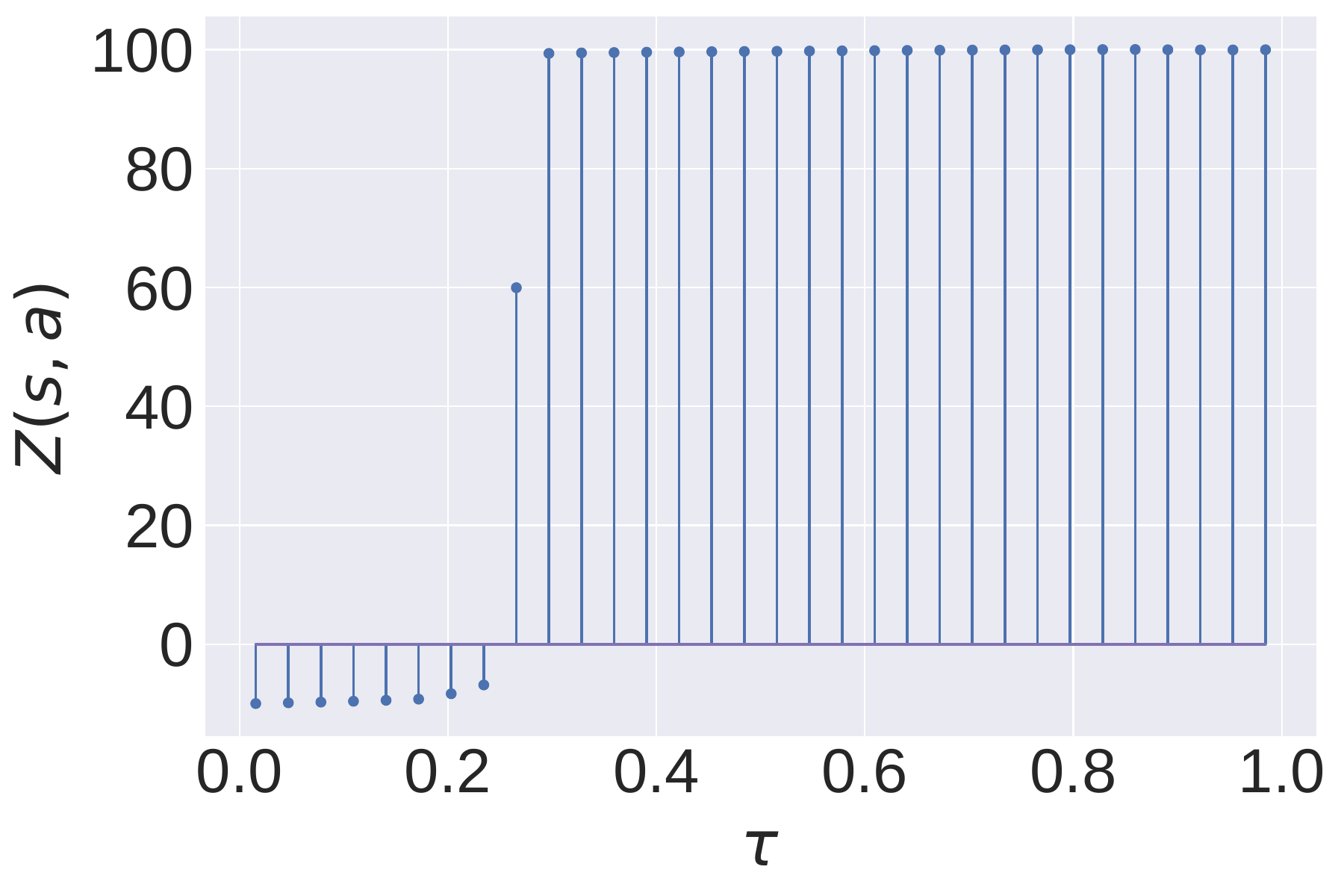}
        \vspace{-14pt}
        \caption*{(c) $s=1,~a=0$}
        \label{fig:simple-z-1-0}
    \end{subfigure}
    \begin{subfigure}[b]{0.19\textwidth}
        \includegraphics[width=\textwidth]{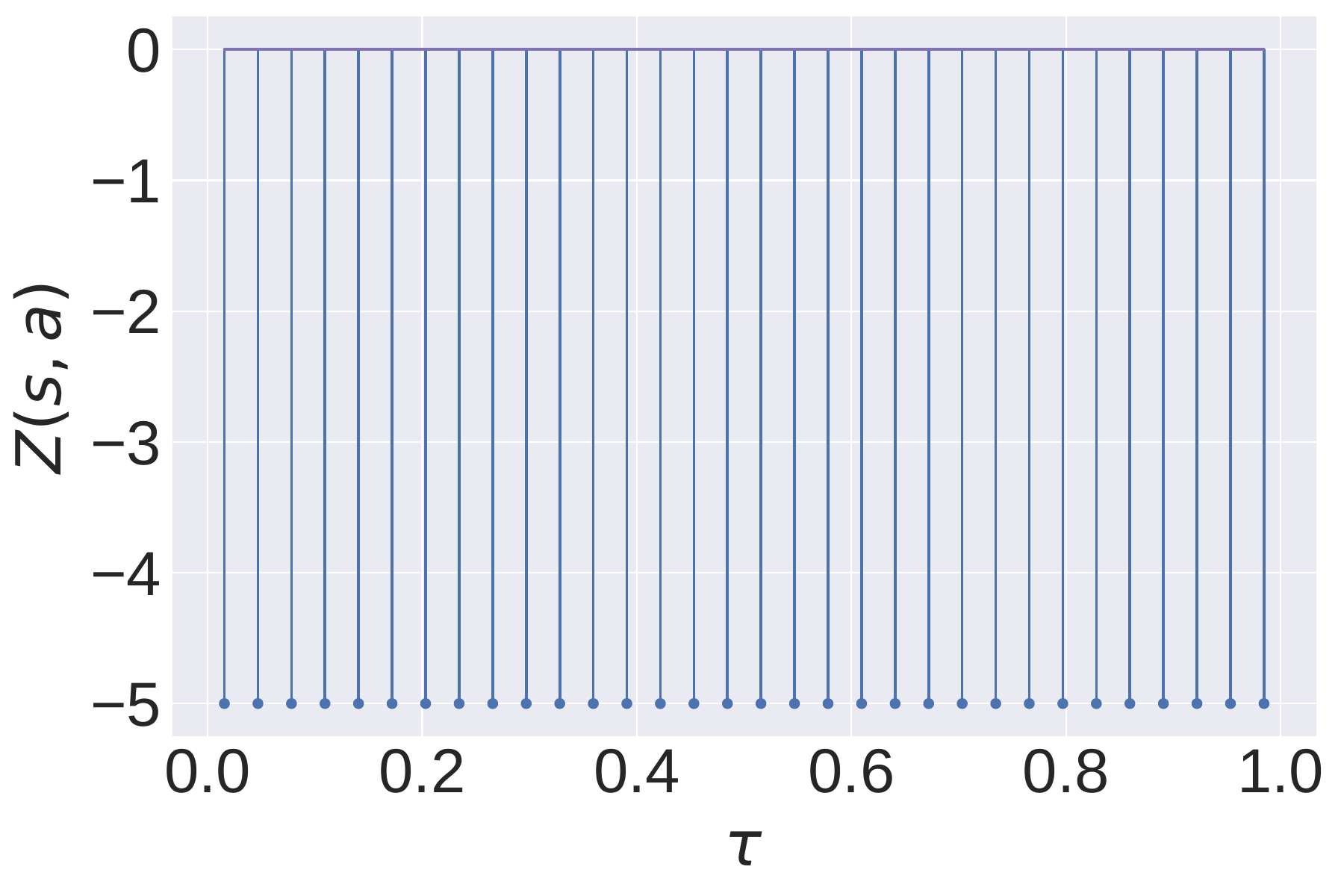}
        \vspace{-14pt}
        \caption*{(d) $s=1,~a=1$}
        \label{fig:simple-z-1-1}
    \end{subfigure}
    ~~\tikz{\draw[-,black, densely dashed, thick](0,-1.80) -- (0,1.00);}~
    \begin{subfigure}[b]{0.19\textwidth}
        \includegraphics[width=\textwidth]{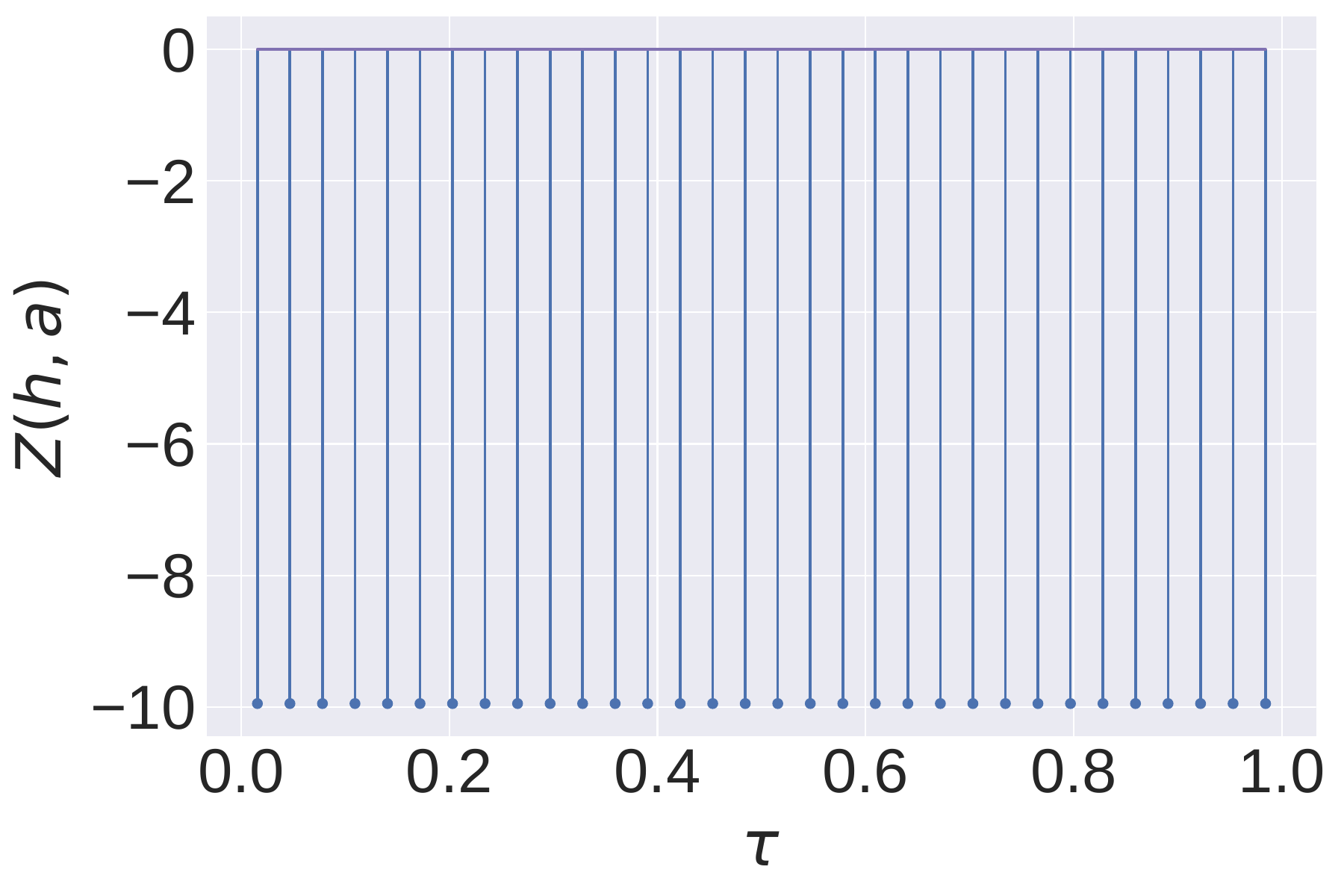}
        \vspace{-14pt}
        \caption*{(h) $h=\{0\},~a=1$}
        \label{fig:simple-z-tilde-1}
    \end{subfigure}
    \begin{subfigure}[b]{0.19\textwidth}
        \includegraphics[width=\textwidth]{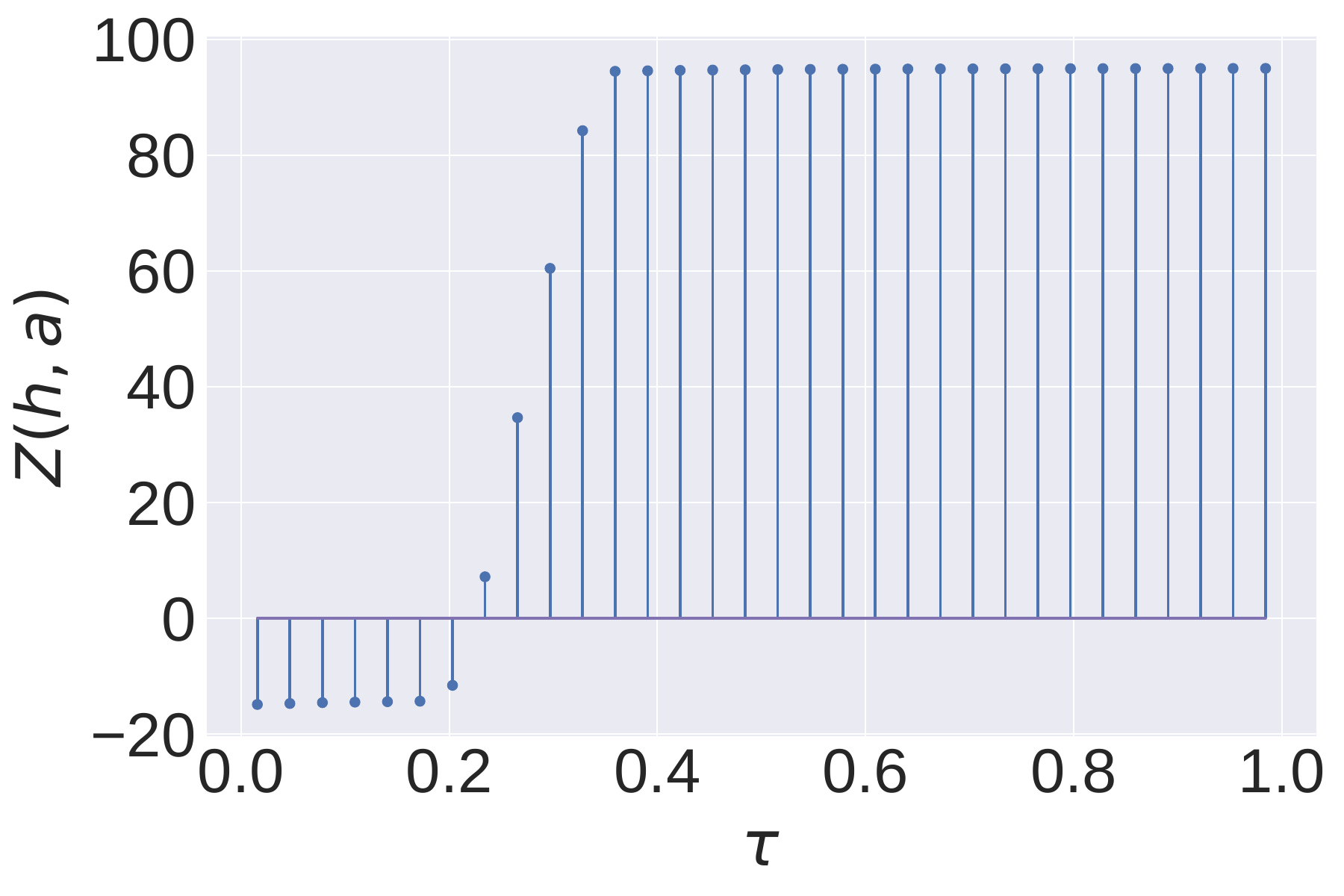}
        \vspace{-14pt}
        \caption*{(i) $h=\{0,1,1\},~a=0$}
        \label{fig:simple-z-tilde-1-0}
    \end{subfigure}
    \begin{subfigure}[b]{0.19\textwidth}
        \includegraphics[width=\textwidth]{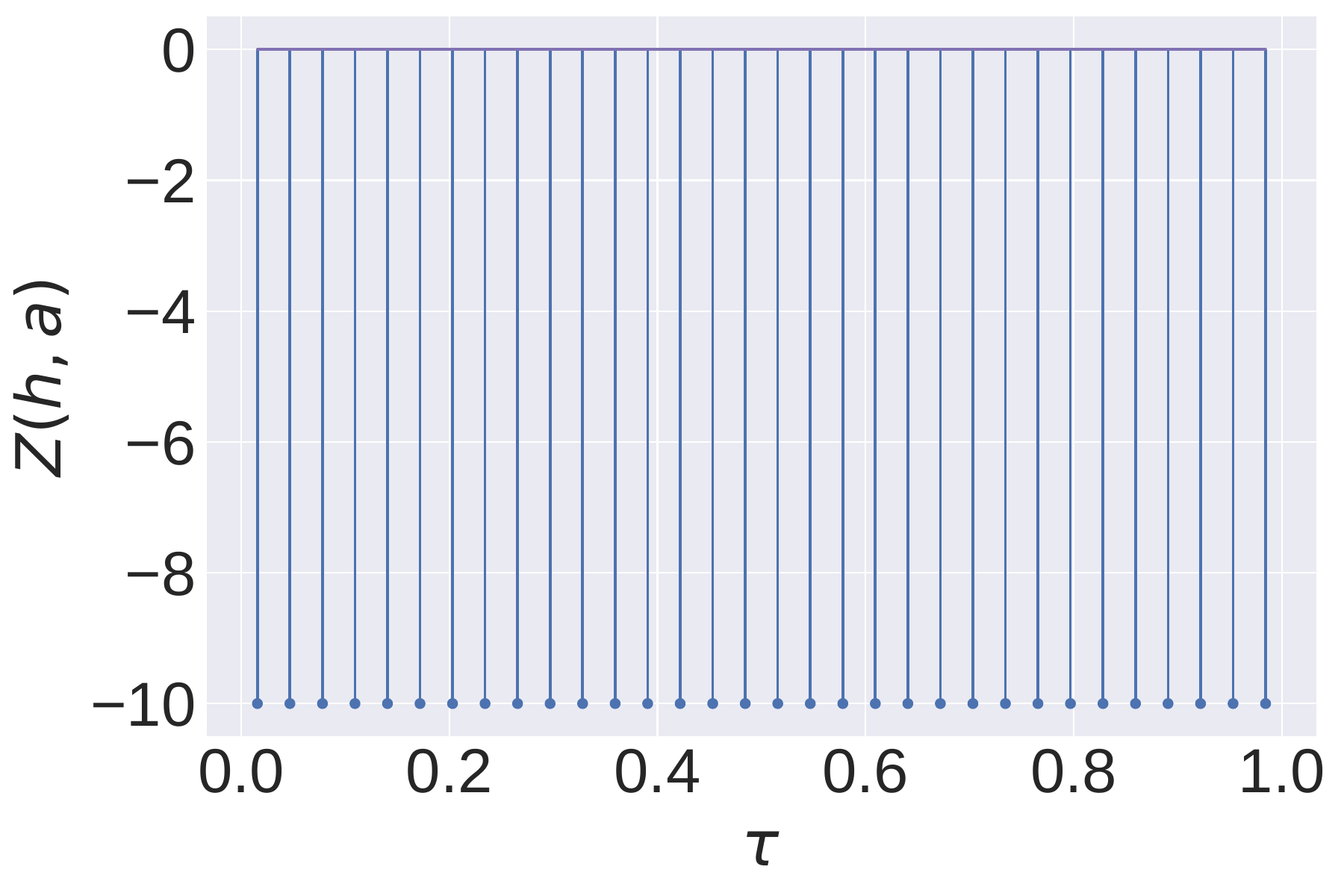}
        \vspace{-14pt}
        \caption*{(j) $h=\{0,1,1\},~a=1$}
        \label{fig:simple-z-tilde-1-1}
    \end{subfigure}
    \vspace{-8pt}
    \caption{Predicted return distribution on different $s$ or $h$ and $a$ input. The left 4 figures correspond to IQN: IQN first learns $Z(s_1,\cdot)$, see (c-d). It finds $a_1=1$ better and keeps this strategy when learning $Z(s_0,\cdot)$, leading to (a) and (b); the right 6 figures correspond to our proposed method TQL: (e) matches (f) as taking $a_1=0$ has better \texttt{CVaR} after taking $a_0=0$; (h) matches (j) as taking $a_1=1$ has better \texttt{CVaR} after taking $a_0=1$. Overall, the policy corresponds to (e) and (f), which achieve global optimum.}
    \label{fig:simple-results}
    \vspace{-4pt}
\end{figure*}

\subsection{Experiment Settings}
\minisection{Environments.} In order to examine the ability to optimize risk-sensitive policy, we design two specified environments for discrete and continuous control respectively. 
For discrete control, we design a risky mini-grid \citep{gym_minigrid} task shown in Fig. 2a. The environment includes $4\times 4$ grids, in which the agent (red triangle) starts at the upper left corner of the grid and aims to reach the bottom right green grid to end the episode. At each timestep, the agent can only choose to move to the adjacent cell either to the right or below its current position. When the agent steps onto a blue or yellow grid for the first time within an episode, it will receive a stochastic reward as in \tb{tab:grid-reward}.

\begin{table}[htbp]
    \centering
    \vspace{-6pt}
    \caption{Stochastic reward of colored grids in mini-grid task.}
    \vspace{-6pt}
    \begin{tabular}{c|c}
        \Xhline{1.5pt}
        Color & Reward Distribution\\
        \hline
        \textbf{Yellow} & $\left\{\begin{array}{cc}
           +100 & p=0.75 \\
           0 & p=0.25
        \end{array}\right.$ \\
        \textbf{Blue} & $+20$ \\
        \Xhline{1.5pt}
    \end{tabular}
    \vspace{-6pt}
    \label{tab:grid-reward}
\end{table}
To avoid the agent from exceeding the region bounded by yellow and blue grids, the outside orange grids will always give the agent a heavy penalty of $-100$.

For continuous control, we augment the original reward of the continuous Mountain-Car environment \citep{Moore90efficientmemory-based} with an extra risky penalty term while entirely preserving the environment dynamics. The penalty is defined as
\begin{equation*}
    R_{\text{risky}}(s,a)=\left\{
    \begin{array}{cl}
        -c\cdot (2-|a|), & p=\frac{1}{4-3|a|} \\
        0, & p=1-\frac{1}{4-3|a|}
    \end{array}
    \right.~.
\end{equation*}
where the risk coefficient $c\in[0,1]$ is a scaling factor that controls the degree of risk related to the scale of actions. At each timestep, we augment the original reward with the risky penalty $R_{\text{risky}}$.
Generally, actions close to $0$ will result in higher expected accumulated rewards. However, to complete the task fast, the agent should choose larger actions that are close to 1, leading to more risky penalties.

\minisection{Implementation, baselines, and metric.}
For discrete action space, we implement TQL based on IQN~\cite{dabney2018iqn} to obtain the value distribution, and compare TQL with vanilla IQN and CVaR-DRL, a specific solution for learning a risk-sensitive policy towards better \texttt{CVaR}, proposed by~\citep{lim2022distributional}. 
For continuous control problems, we combine TD3~\citep{fujimoto2018td3} with IQN~\citep{dabney2018iqn}, named IQTD3, by replacing the critics in TD3 with distributional critics. We further build TQL upon IQTD3 and take IQTD3 as the baseline algorithm.
For comparison, each algorithm is optimized towards various risk-sensitive objectives that are represented by different risk measures, including \texttt{mean}, \texttt{CVaR}, \texttt{POW}, and \texttt{Wang}, whose detailed description is in \ap{ap:exp-vs-distortion}; and the evaluation metrics are also those risk measures.

\subsection{Results and Analysis}\label{subsec:exp-results}

\minisection{Value distribution analysis on 3-state MDP.}
In \se{subsec:biased-obj}, we have illustrated in \fig{fig:3state-mdp} that vanilla distributional RL is not able to reveal the global optimal risk-sensitive policy and its value. To validate our theoretical results in practice, we learn the return distribution with a tabular version of vanilla IQN and TQL respectively, and visualize the learned return distribution in \fig{fig:simple-results}. The results show that vanilla distributional RL tends to learn $Z(s_1,\cdot)$ first as it is irrelevant to $a_0$, and thus $a_1=1$ will chosen at $s_1$. However, when learning $Z(s_0,\cdot)$, Bellman update will use $Z(s_1,a_1)$ in target, ignoring all trajectories where $a_1=0$ and leading to a sub-optimal policy $a_0=a_1=1$. On the contrary, TQL learns historical value distribution, which enforces the agent to consider all possible trajectories and thus reveal the optimal solution where $a_0=a_1=0$.

\minisection{Discrete control evaluations.}
We first show the result of the discrete mini-grid task, which is designed for learning \texttt{CVaR} objective.
We present the learning curves of TQL and vanilla IQN in Fig. 2b, which indicates that IQN consistently converges to the sub-optimal solution of visiting blue grids, similar to its behavior in the aforementioned 3-state MDP. CVaR-DRL does improve the \texttt{CVaR} of historical return distribution to some extent, while it still produces sub-optimal policies (see \ap{ap:compare} for a more detailed analysis). In contrast, TQL is able to discover a better policy that achieves significantly higher \texttt{CVaR} than the vanilla IQN baseline. 
Furthermore, in Fig. 2c, we visualize the final policy's return distribution. The blue, red, and green bars indicate the frequency of episode returns for vanilla IQN, CVaR-DRL, and TQL respectively, and the corresponding dashed lines show the \texttt{CVaR} of return distribution for two policies. TQL is very likely to obtain high positive returns with little risk of negative returns, while vanilla IQN's return is always negative due to its Markovian policy and CVaR-DRL learns an intermediate policy.

To better understand the difference in the optimization process, we further illustrate how the policy evolves during the training process in \tb{tab:grid-evolve}. In particular, we observe that vanilla IQN converges from the end of the episode to the beginning due to its updating mechanism of dynamic programming, and its property of Markovian prevents it from finding the global optimum; 
moreover, CVaR-DRL fails due to its approximation in \texttt{CVaR} estimation but leads to a slightly-better policy.
However, TQL is always doing a global search and thus finally reveals the optimal policy.
\begin{table}[tbp]
    \centering
    \vspace{-4pt}
    \caption{The action sequences of IQN and TQL policies learnt at different training steps. IQN converges from back to front; CVaR-DRL leads to a slightly-better policy; TQL finds out the global optimal policy.}
    \vspace{-6pt}
    \resizebox{\columnwidth}{!}{
    \begin{tabular}{c|ccc}
        \hline
        \# Train steps & IQN & CVaR-DRL & TQL \\
        \hline
        $2\times 10^4$ & [$\downarrow$, $\rightarrow$, $\downarrow$, $\rightarrow$, $\rightarrow$, $\downarrow$] & [$\rightarrow$, $\downarrow$, $\downarrow$, $\rightarrow$, $\rightarrow$, $\downarrow$] & [$\rightarrow$, $\downarrow$, $\downarrow$, $\rightarrow$, $\rightarrow$, $\downarrow$] \\
        $1\times 10^5$ & [$\downarrow$, $\rightarrow$, $\rightarrow$, $\downarrow$, $\rightarrow$, $\downarrow$] & [$\rightarrow$, $\downarrow$, $\downarrow$, $\rightarrow$, $\rightarrow$, $\downarrow$] & [$\downarrow$, $\rightarrow$, $\downarrow$, $\rightarrow$, $\rightarrow$, $\downarrow$] \\
        $2\times 10^5$ & [$\rightarrow$, $\downarrow$, $\rightarrow$, $\downarrow$, $\rightarrow$, $\downarrow$] & [$\downarrow$, $\rightarrow$, $\rightarrow$, $\downarrow$, $\rightarrow$, $\downarrow$] & [$\downarrow$, $\rightarrow$, $\downarrow$, $\rightarrow$, $\downarrow$, $\rightarrow$] \\
        \hline
    \end{tabular}
    }
    \vspace{-10pt}
    \label{tab:grid-evolve}
\end{table}

\begin{figure*}[t!]
    \centering
    \rotatebox{90}{\scriptsize{$c=0.0$~~}}~~~~~~
    \begin{minipage}{0.95\textwidth}
    \begin{subfigure}[b]{0.18\textwidth}
        \includegraphics[width=\textwidth]{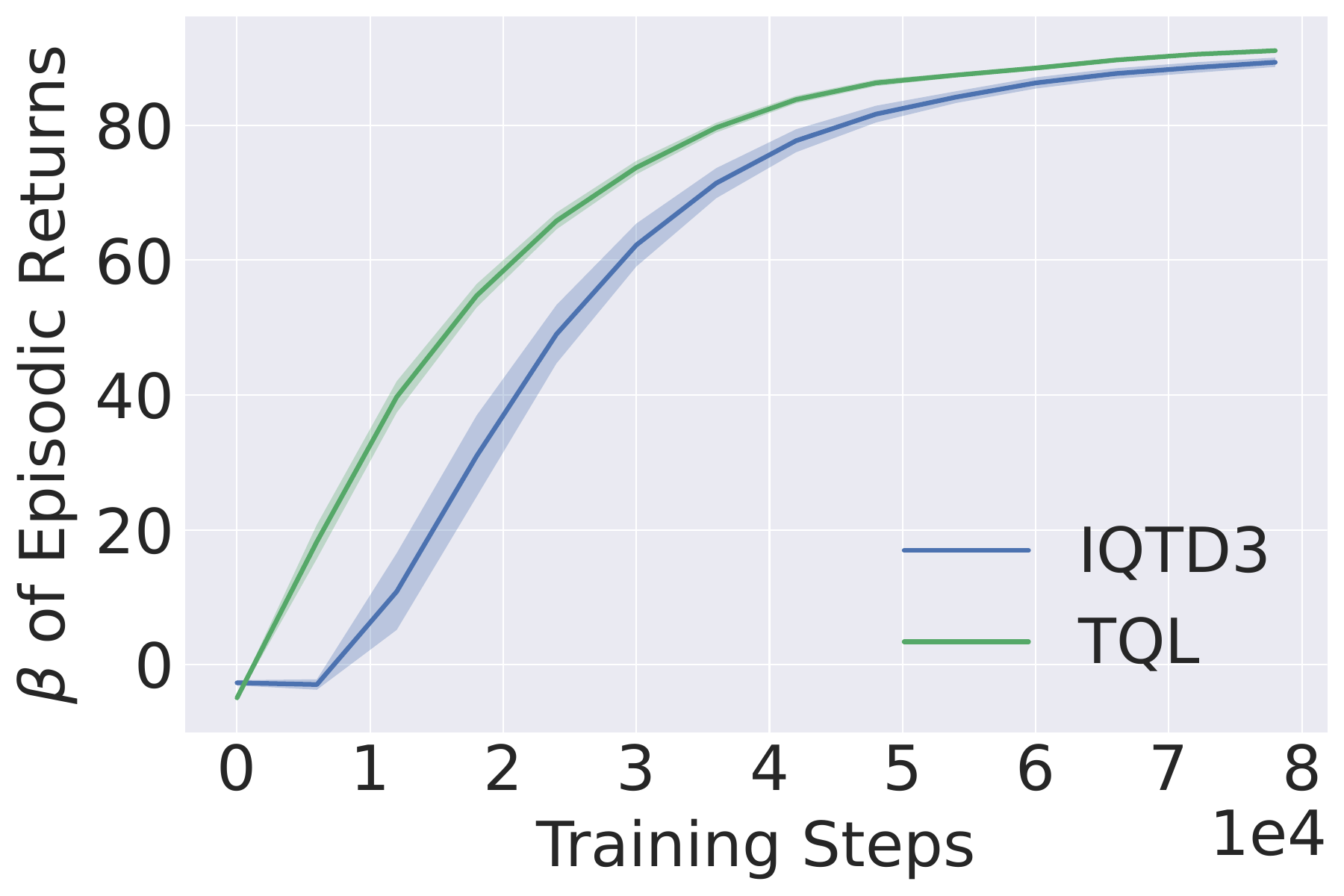}
        \label{fig:mcc-0.0-cvar}
        \vspace{-10pt}
    \end{subfigure}~
    \begin{subfigure}[b]{0.18\textwidth}
        \includegraphics[width=\textwidth]{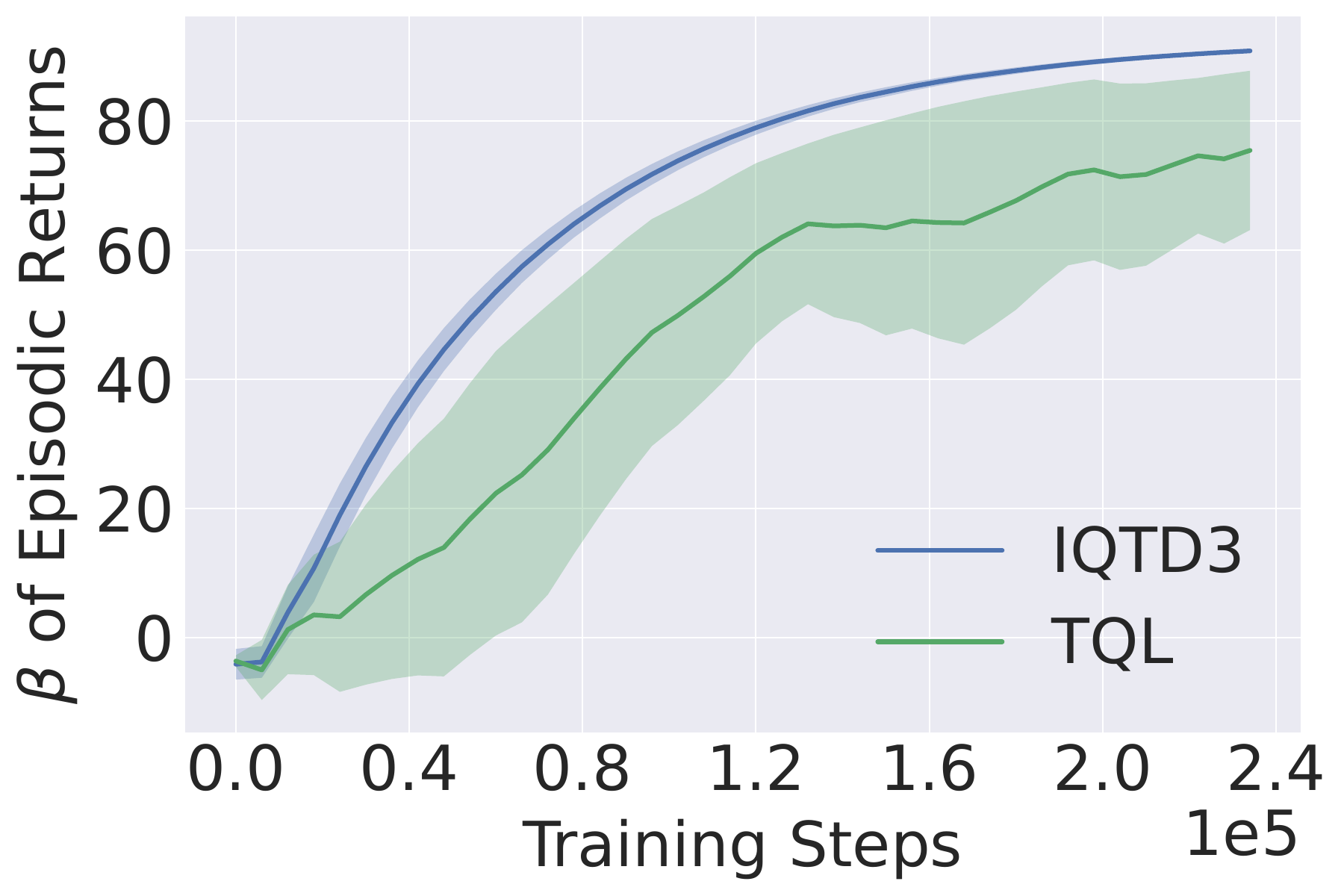}
        \label{fig:mcc-0.0-cpw}
        \vspace{-10pt}
    \end{subfigure}~
    \begin{subfigure}[b]{0.18\textwidth}
        \includegraphics[width=\textwidth]{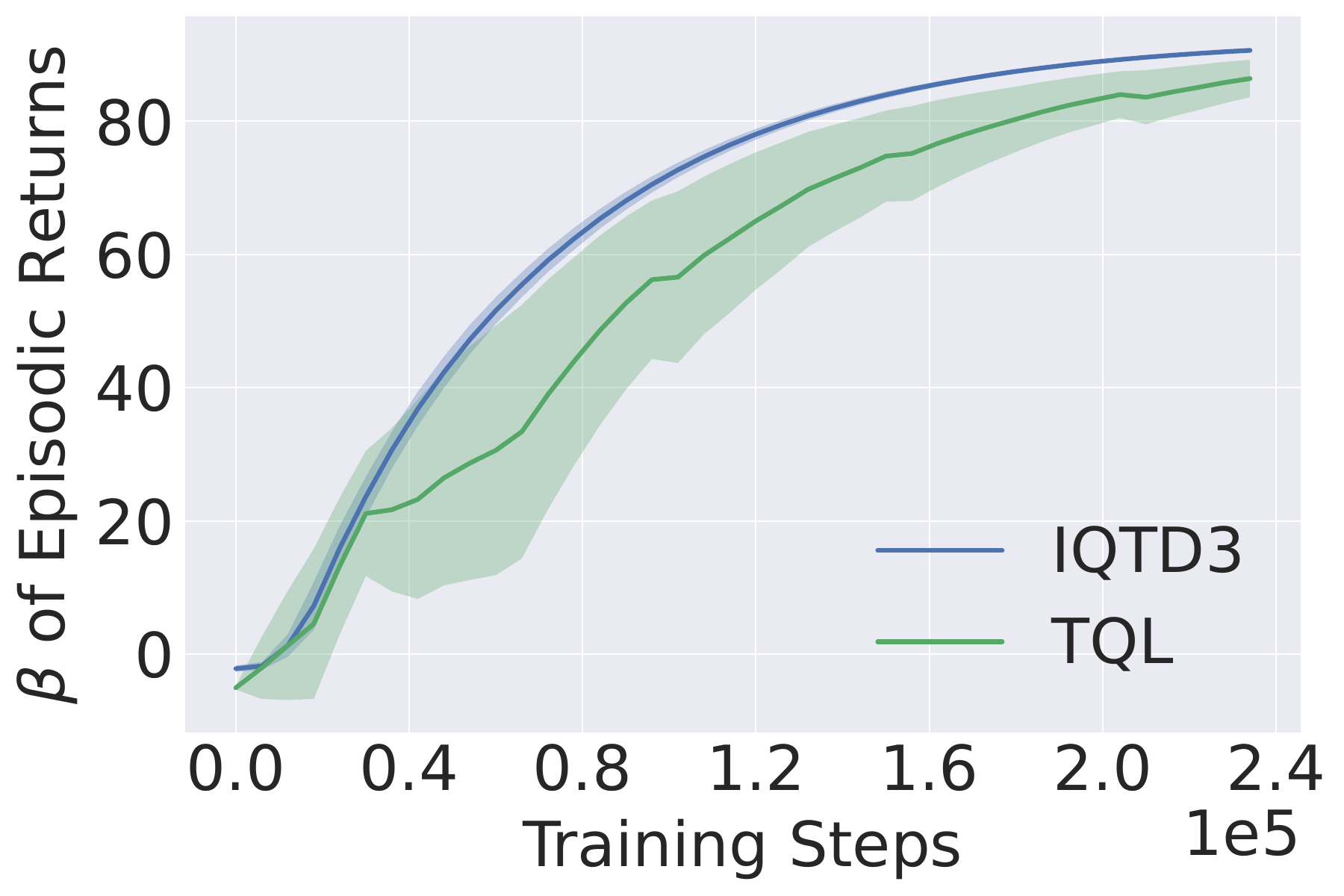}
        \label{fig:mcc-0.0-pow}
        \vspace{-10pt}
    \end{subfigure}~
    \begin{subfigure}[b]{0.18\textwidth}
        \includegraphics[width=\textwidth]{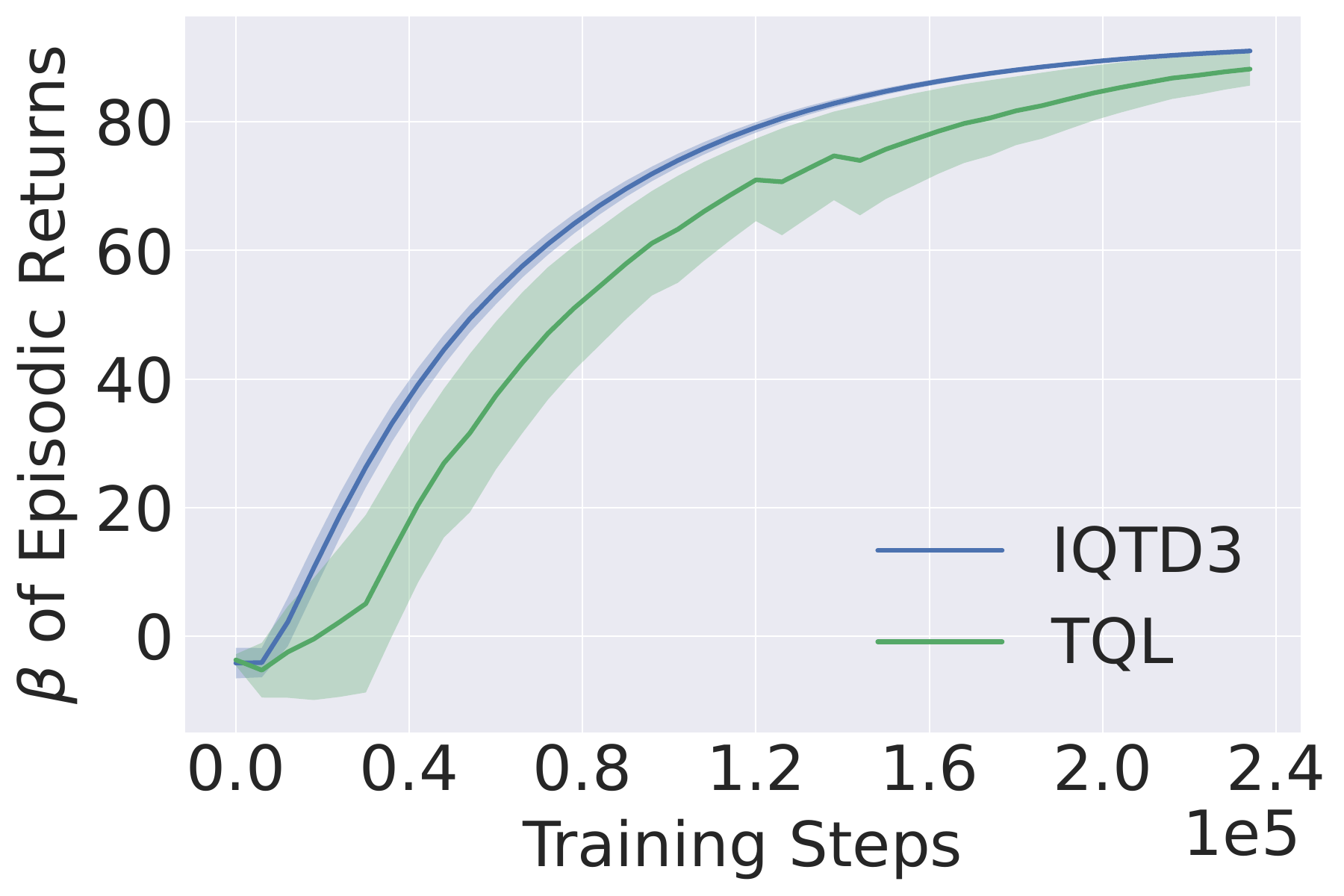}
        \label{fig:mcc-0.0-wang-pos}
        \vspace{-10pt}
    \end{subfigure}~
    \begin{subfigure}[b]{0.18\textwidth}
        \includegraphics[width=\textwidth]{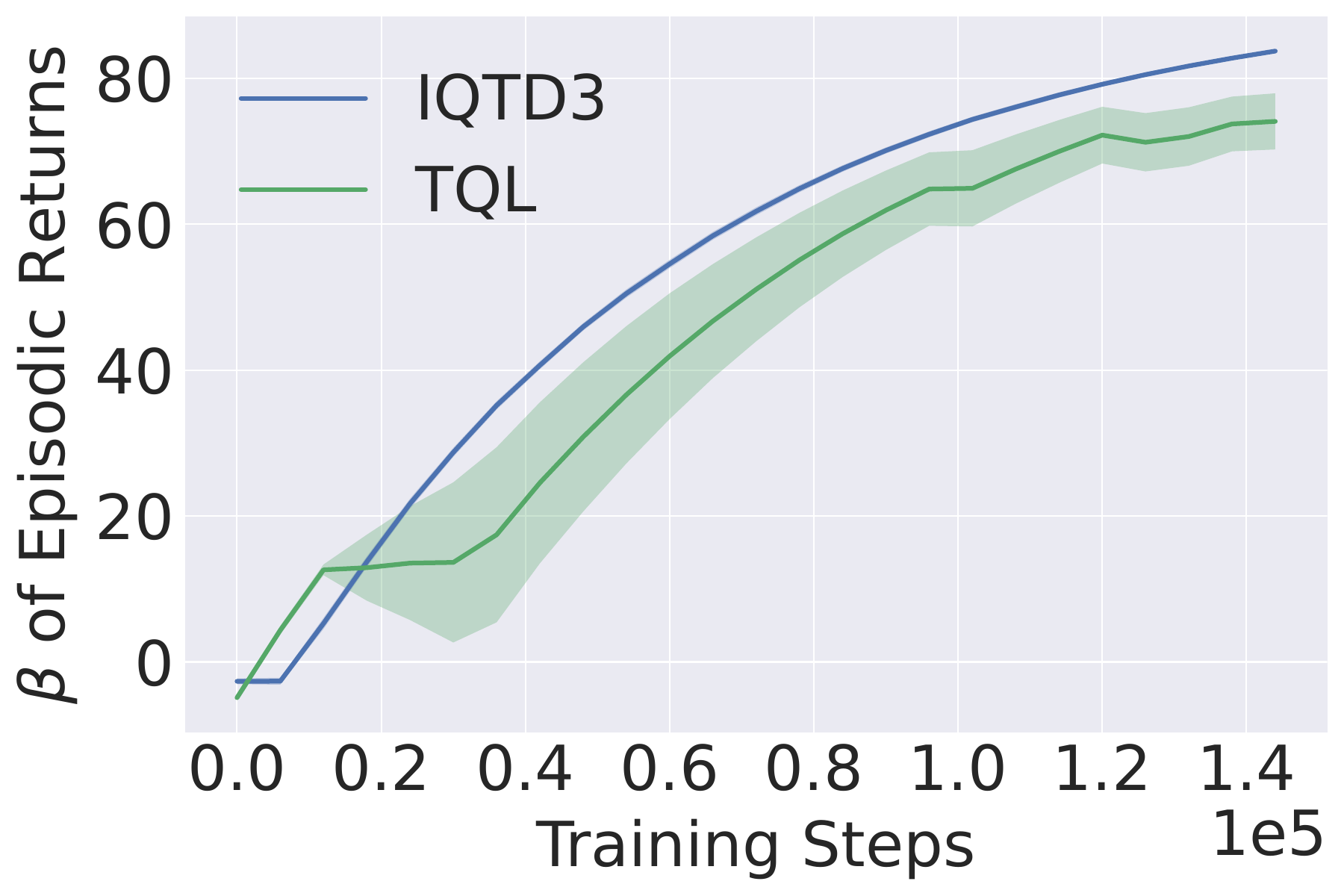}
        \label{fig:mcc-0.0-wang-neg}
        \vspace{-10pt}
    \end{subfigure}
    \end{minipage}\\
    \rotatebox{90}{\scriptsize{$c=0.1$~~}}~~~~~~
    \begin{minipage}{0.95\textwidth}
    \begin{subfigure}[b]{0.18\textwidth}
        \includegraphics[width=\textwidth]{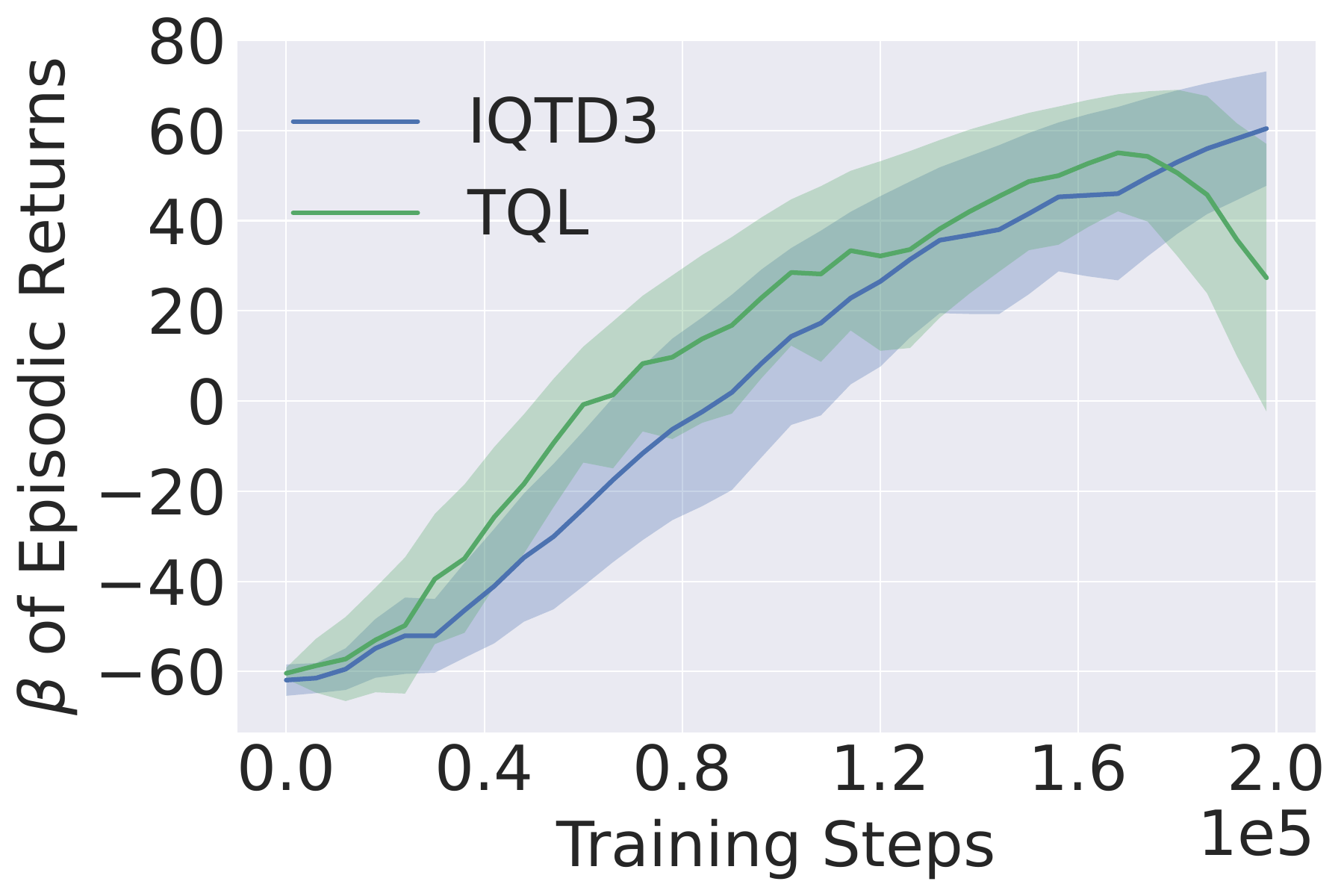}
        \label{fig:mcc-0.1-cvar}
        \vspace{-10pt}
    \end{subfigure}~
    \begin{subfigure}[b]{0.18\textwidth}
        \includegraphics[width=\textwidth]{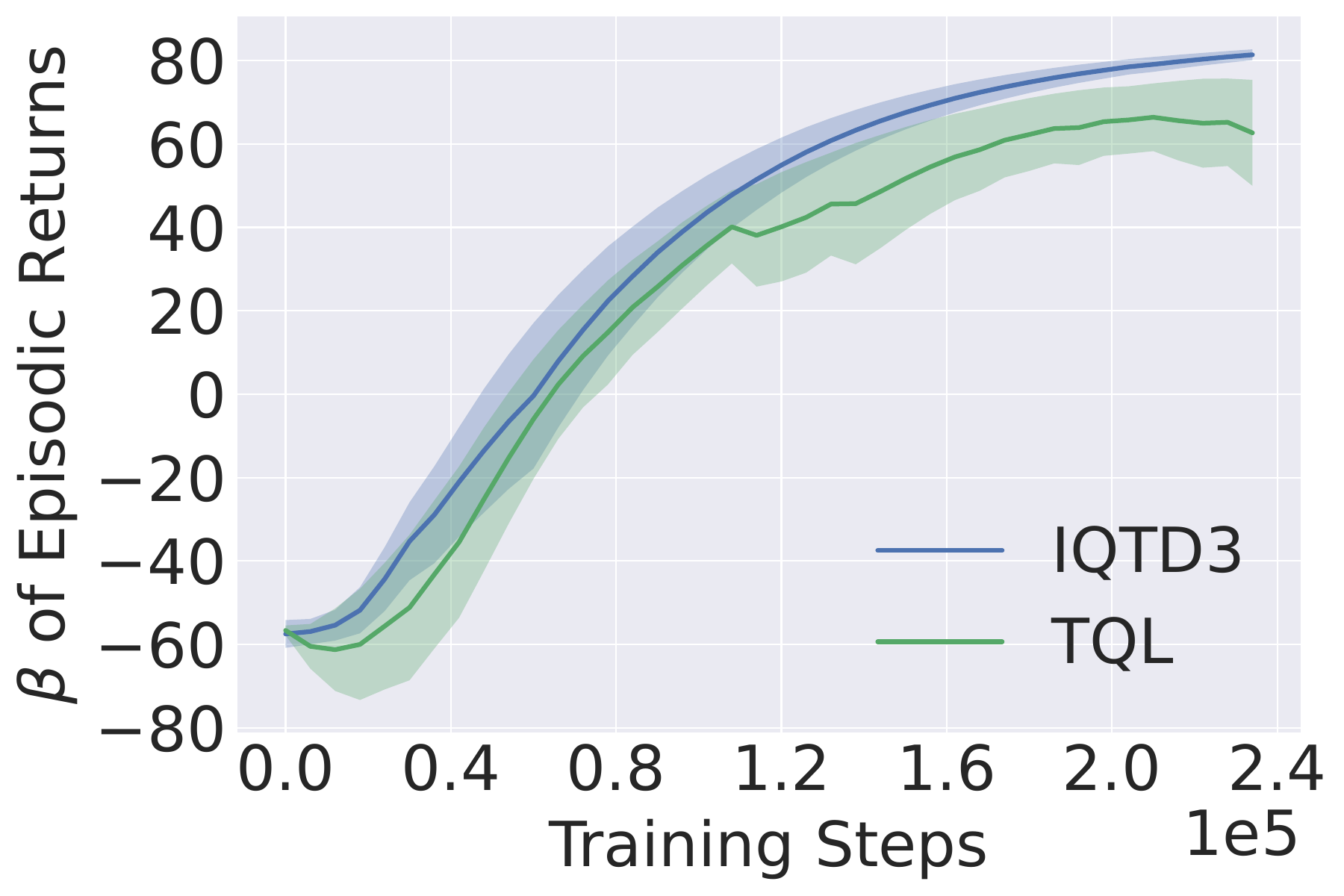}
        \label{fig:mcc-0.1-cpw}
        \vspace{-10pt}
    \end{subfigure}~
    \begin{subfigure}[b]{0.18\textwidth}
        \includegraphics[width=\textwidth]{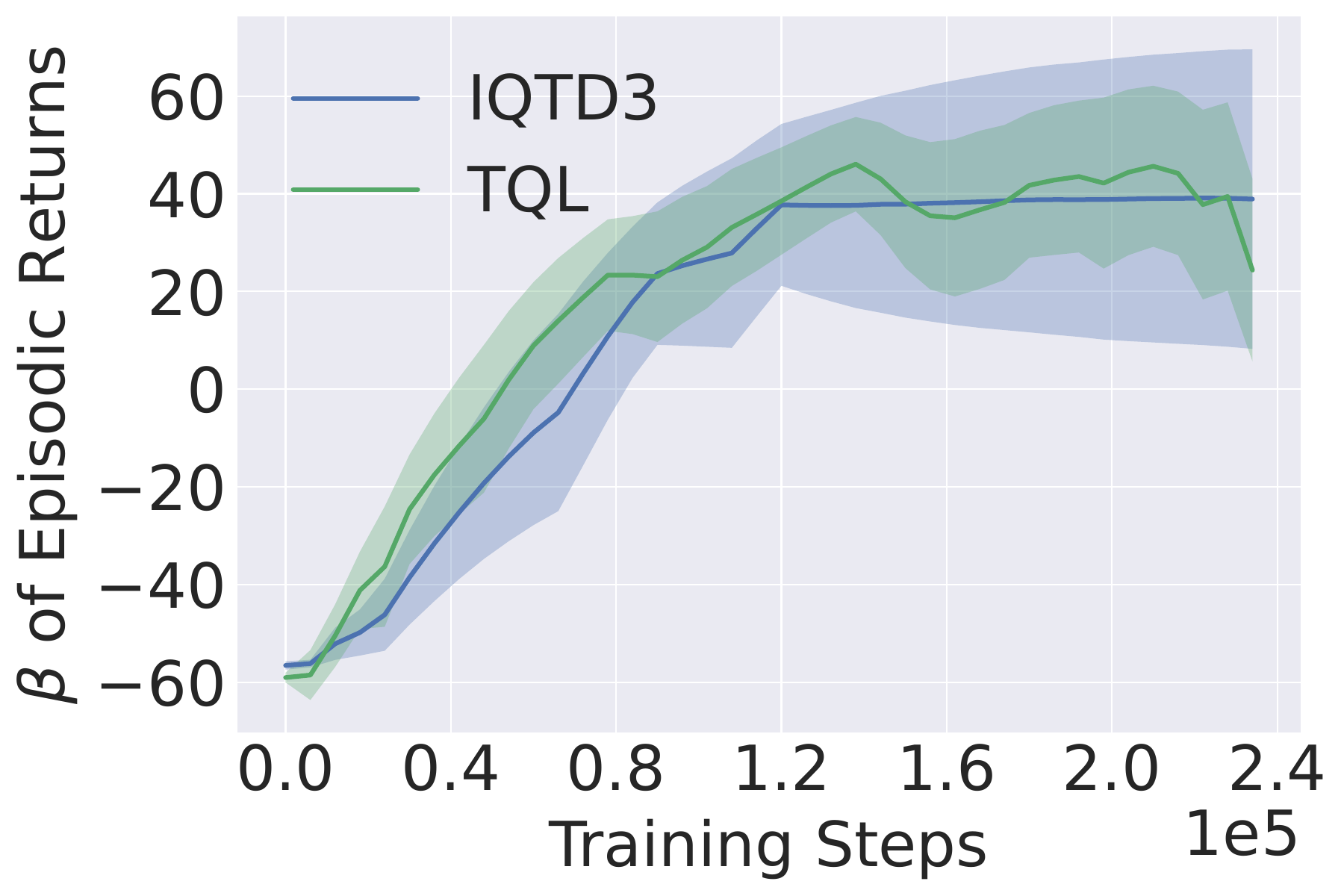}
        \label{fig:mcc-0.1-pow}
        \vspace{-10pt}
    \end{subfigure}~
    \begin{subfigure}[b]{0.18\textwidth}
        \includegraphics[width=\textwidth]{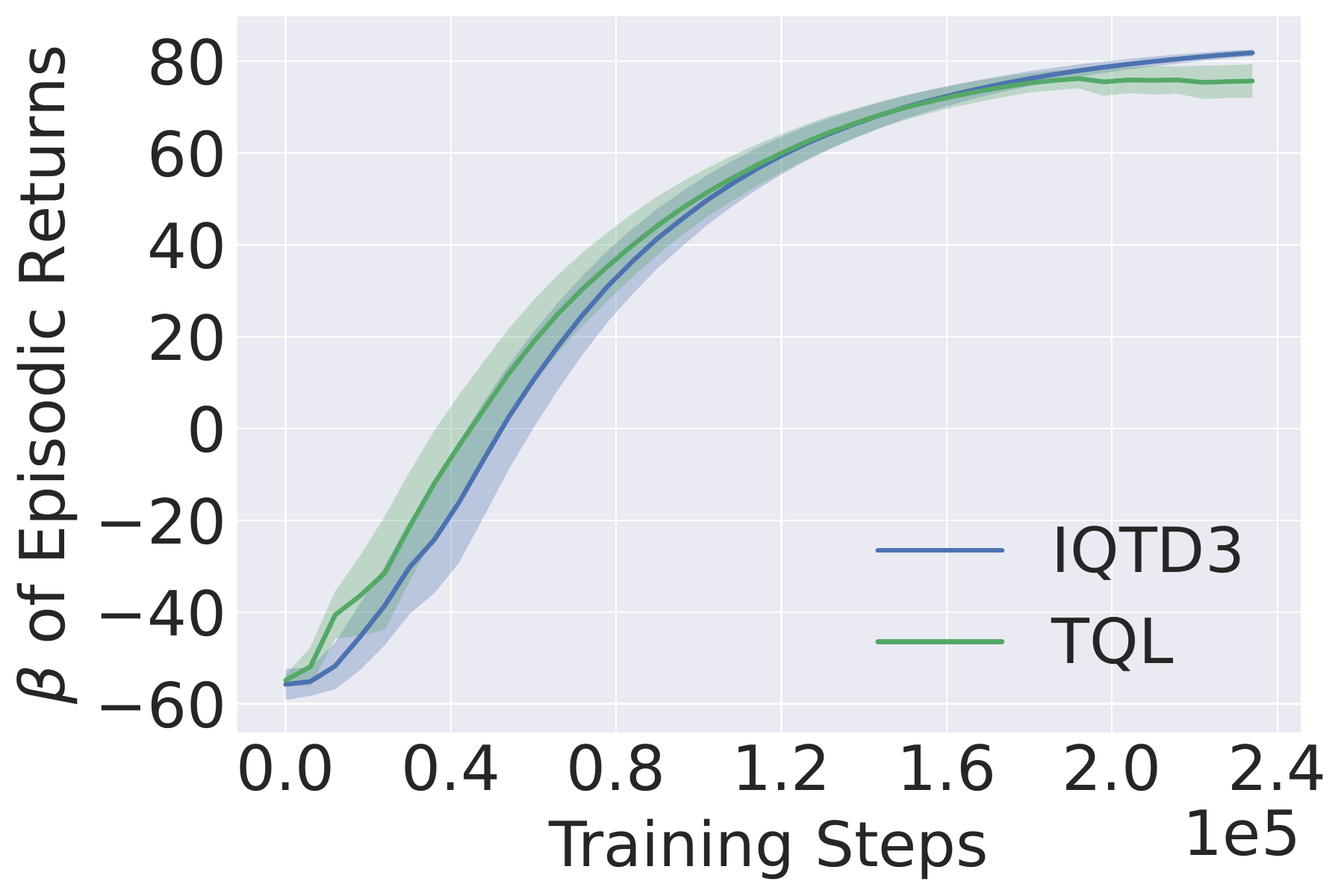}
        \label{fig:mcc-0.1-wang-pos}
        \vspace{-10pt}
    \end{subfigure}~
    \begin{subfigure}[b]{0.18\textwidth}
        \includegraphics[width=\textwidth]{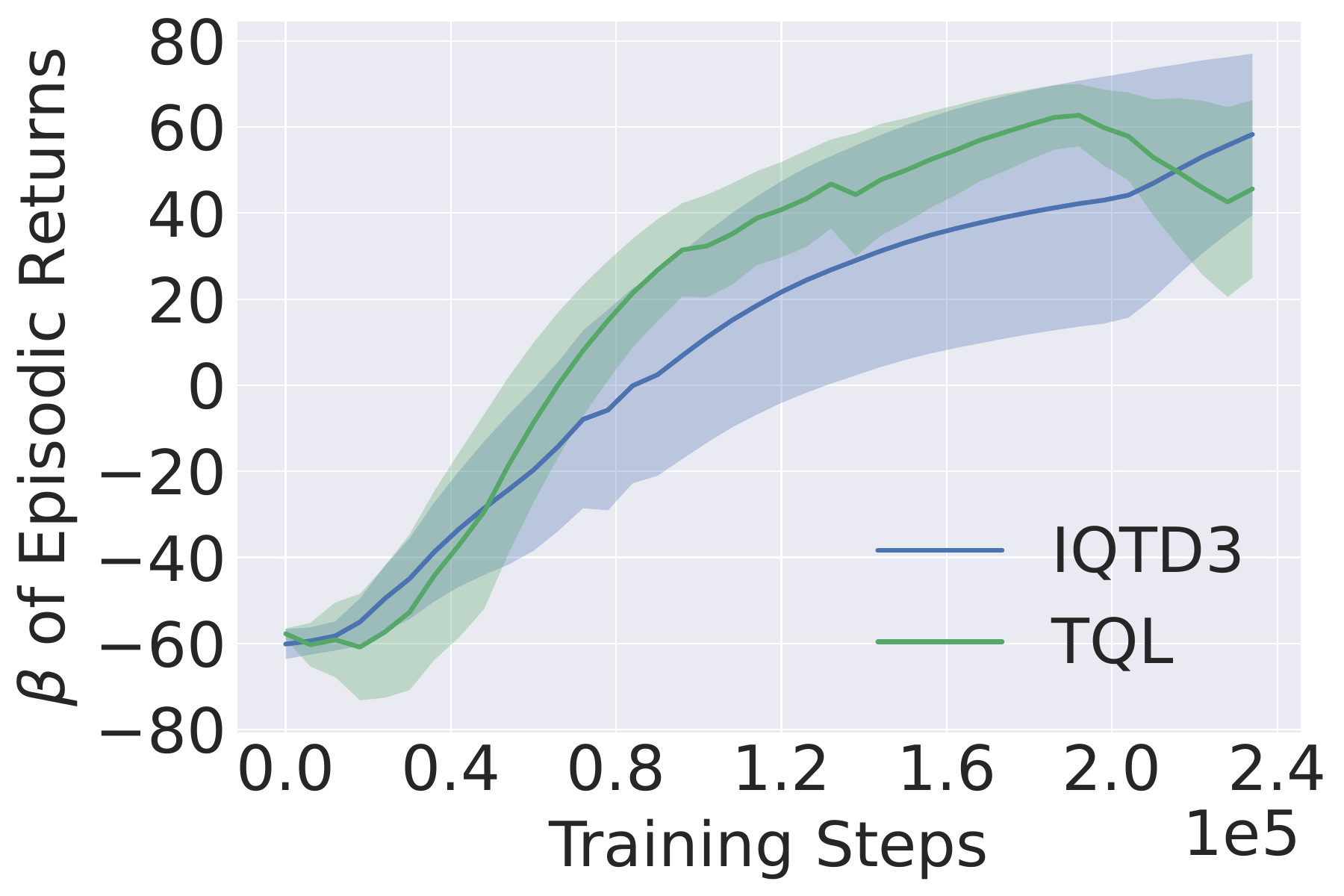}
        \label{fig:mcc-0.1-wang-neg}
        \vspace{-10pt}
    \end{subfigure}
    \end{minipage}\\
    \rotatebox{90}{\scriptsize{$c=0.25$~~}}~~~~~~
    \begin{minipage}{0.95\textwidth}
    \begin{subfigure}[b]{0.18\textwidth}
        \includegraphics[width=\textwidth]{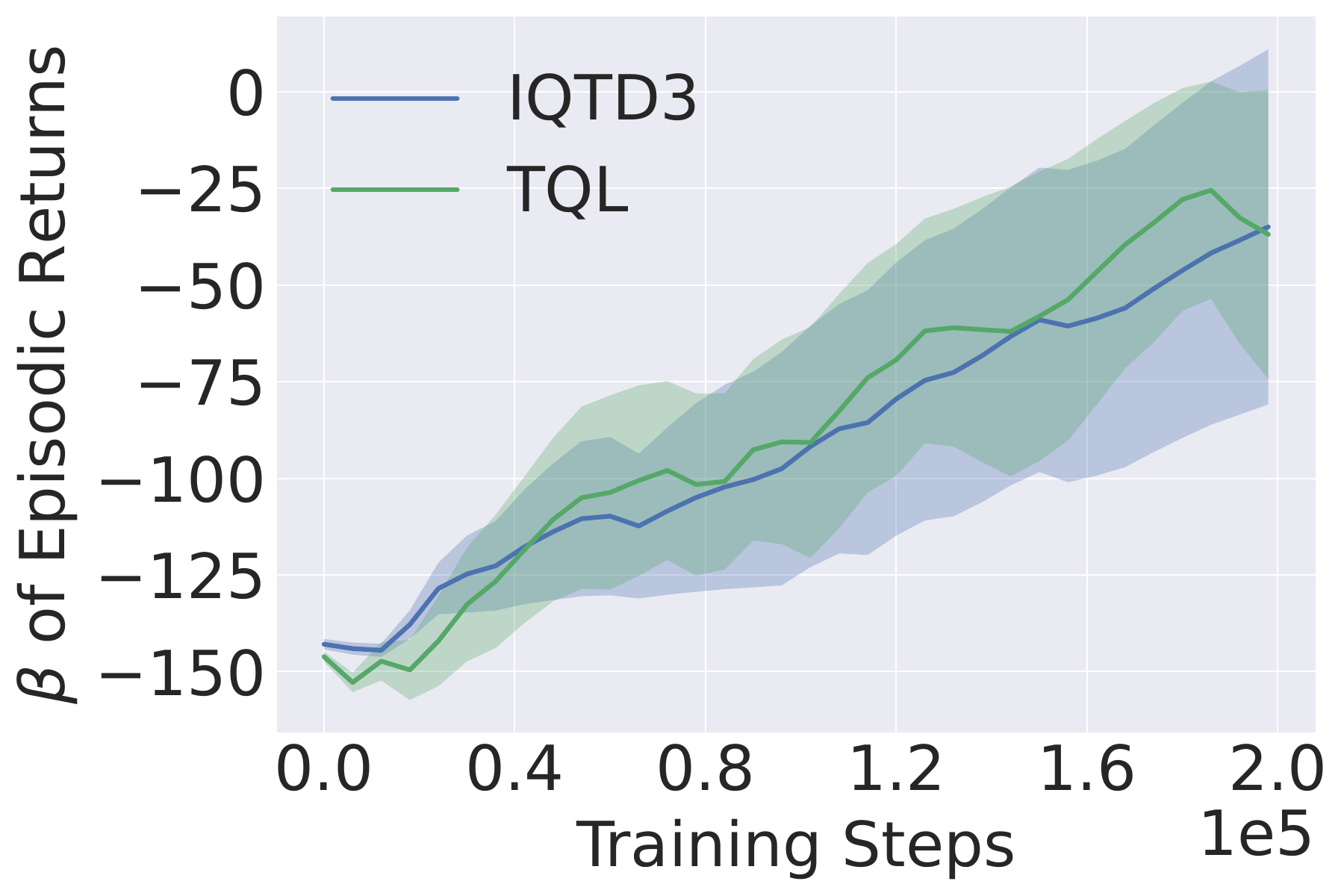}
        \label{fig:mcc-0.25-cvar}
        \vspace{-10pt}
    \end{subfigure}~
    \begin{subfigure}[b]{0.18\textwidth}
        \includegraphics[width=\textwidth]{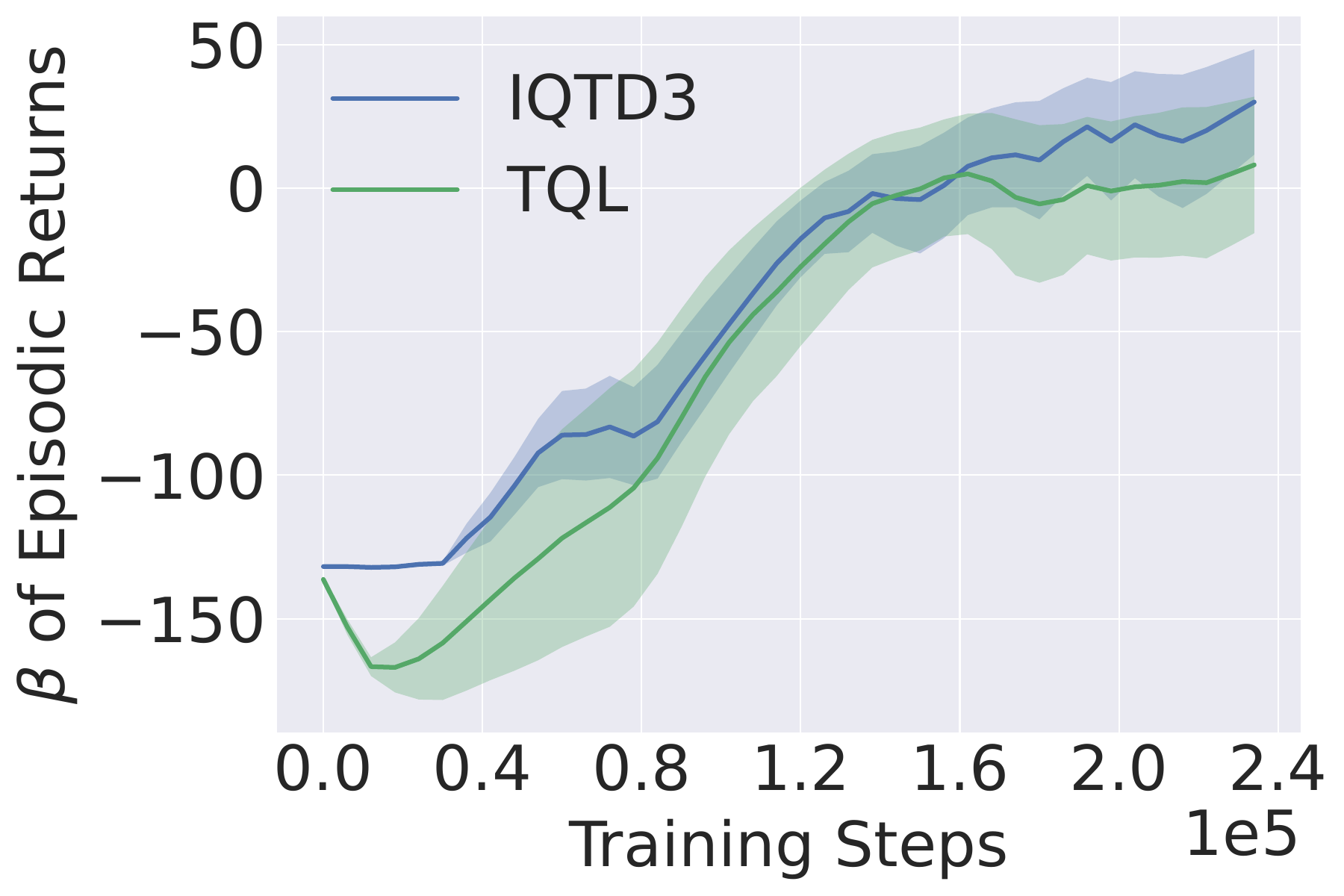}
        \label{fig:mcc-0.25-cpw}
        \vspace{-10pt}
    \end{subfigure}~
    \begin{subfigure}[b]{0.18\textwidth}
        \includegraphics[width=\textwidth]{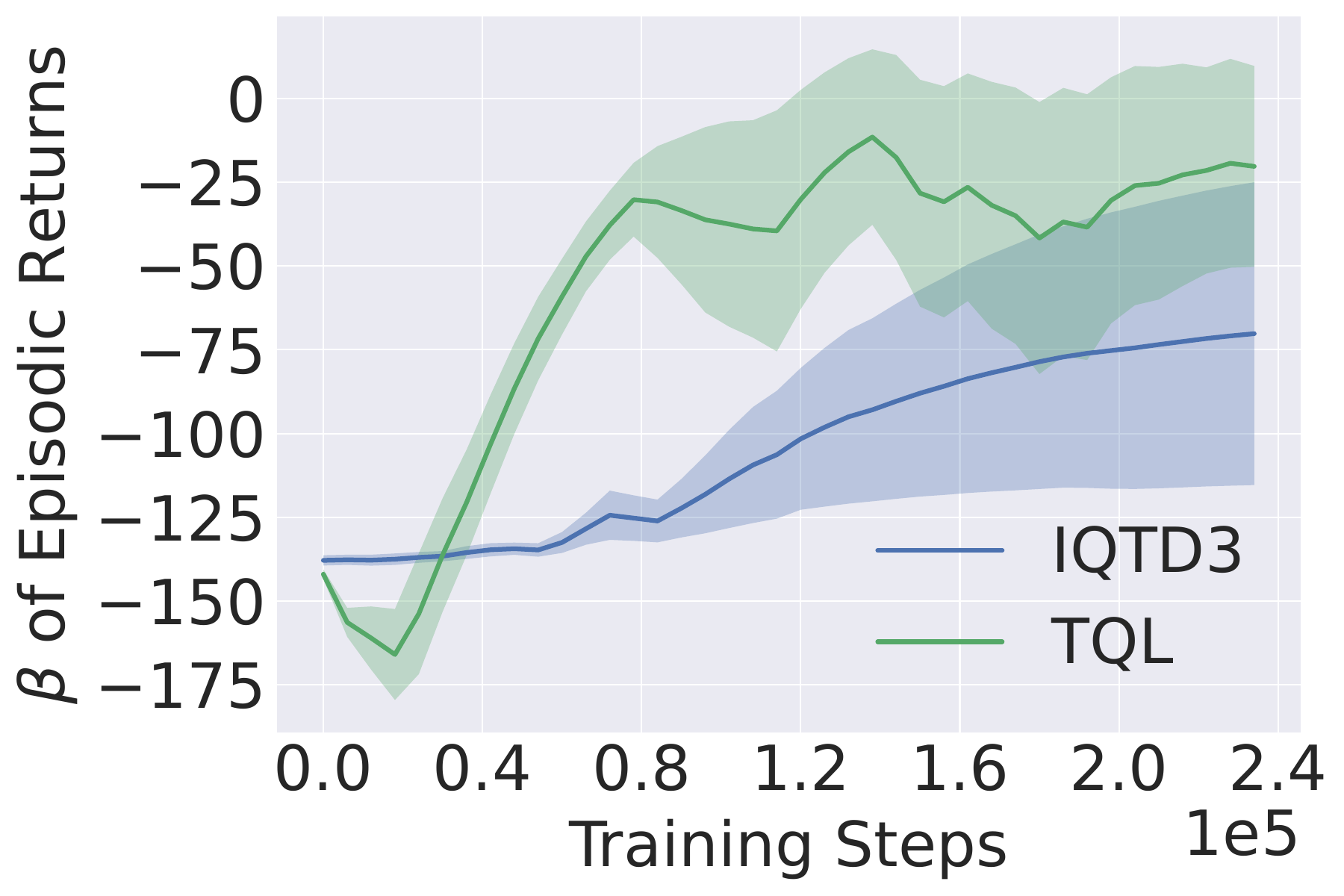}
        \label{fig:mcc-0.25-pow}
        \vspace{-10pt}
    \end{subfigure}~
    \begin{subfigure}[b]{0.18\textwidth}
        \includegraphics[width=\textwidth]{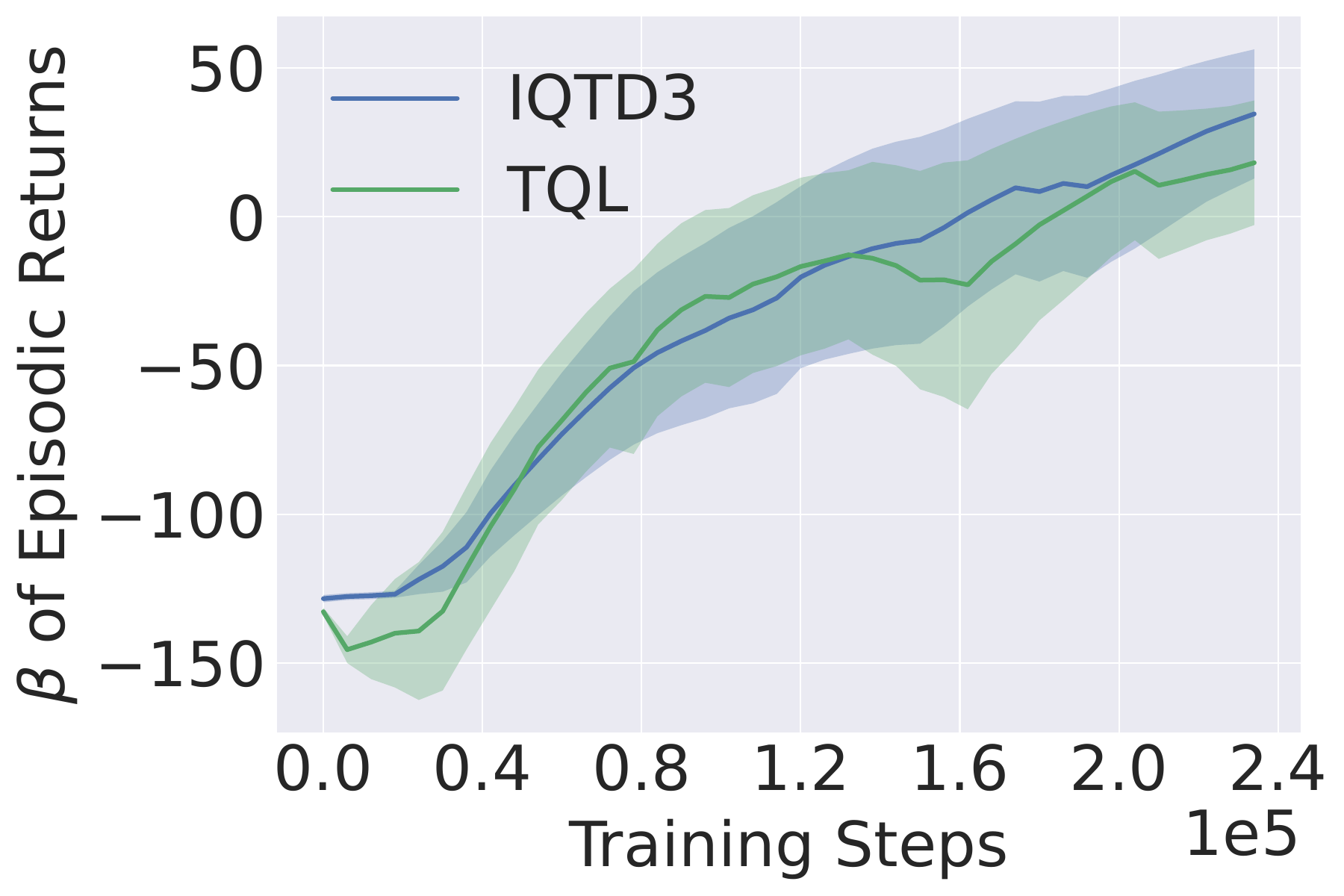}
        \label{fig:mcc-0.25-wang-pos}
        \vspace{-10pt}
    \end{subfigure}~
    \begin{subfigure}[b]{0.18\textwidth}
        \includegraphics[width=\textwidth]{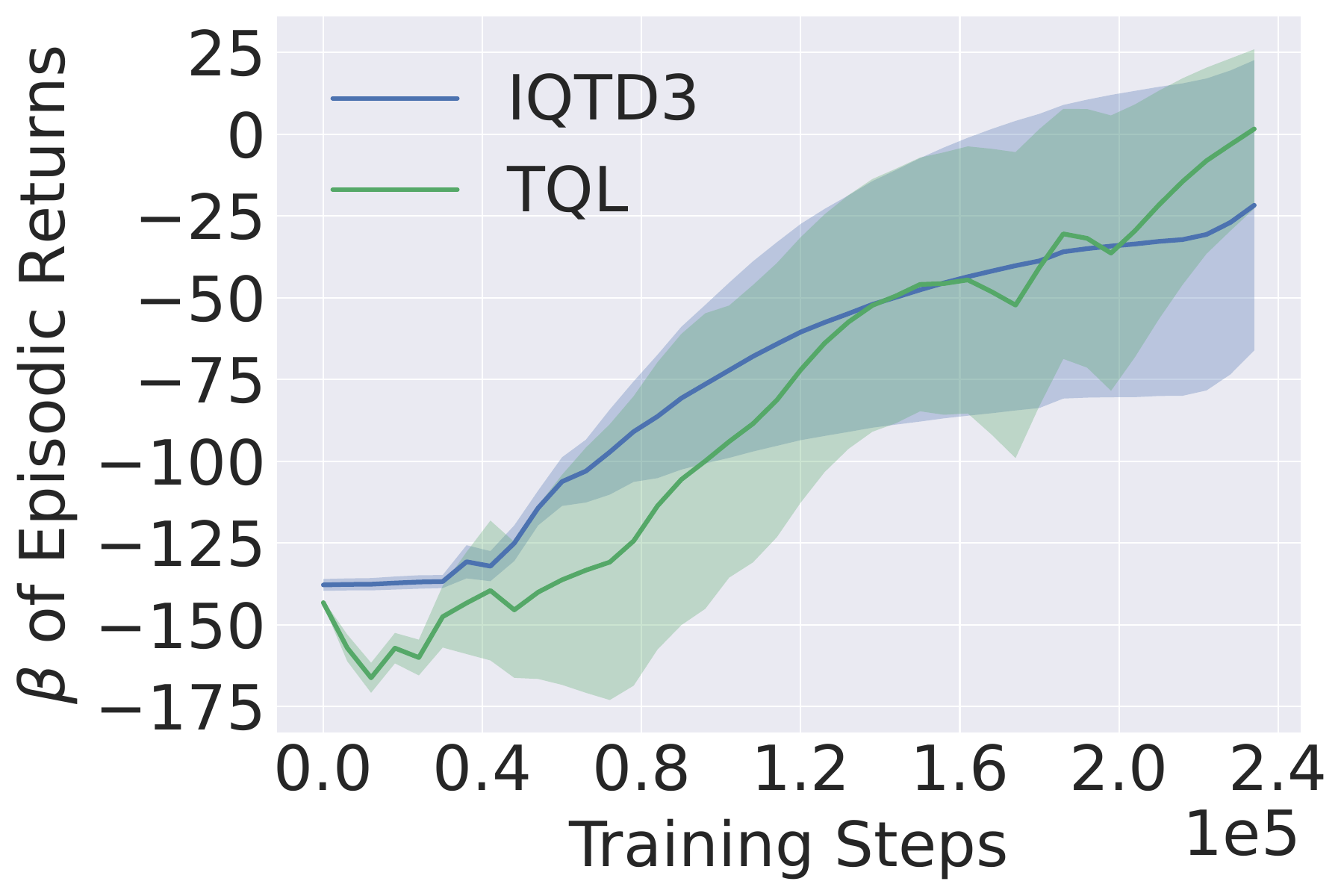}
        \label{fig:mcc-0.25-wang-neg}
        \vspace{-10pt}
    \end{subfigure}
    \end{minipage}\\
    \rotatebox{90}{\scriptsize{$c=0.5$~~}}~~~~~~
    \begin{minipage}{0.95\textwidth}
    \begin{subfigure}[b]{0.18\textwidth}
        \includegraphics[width=\textwidth]{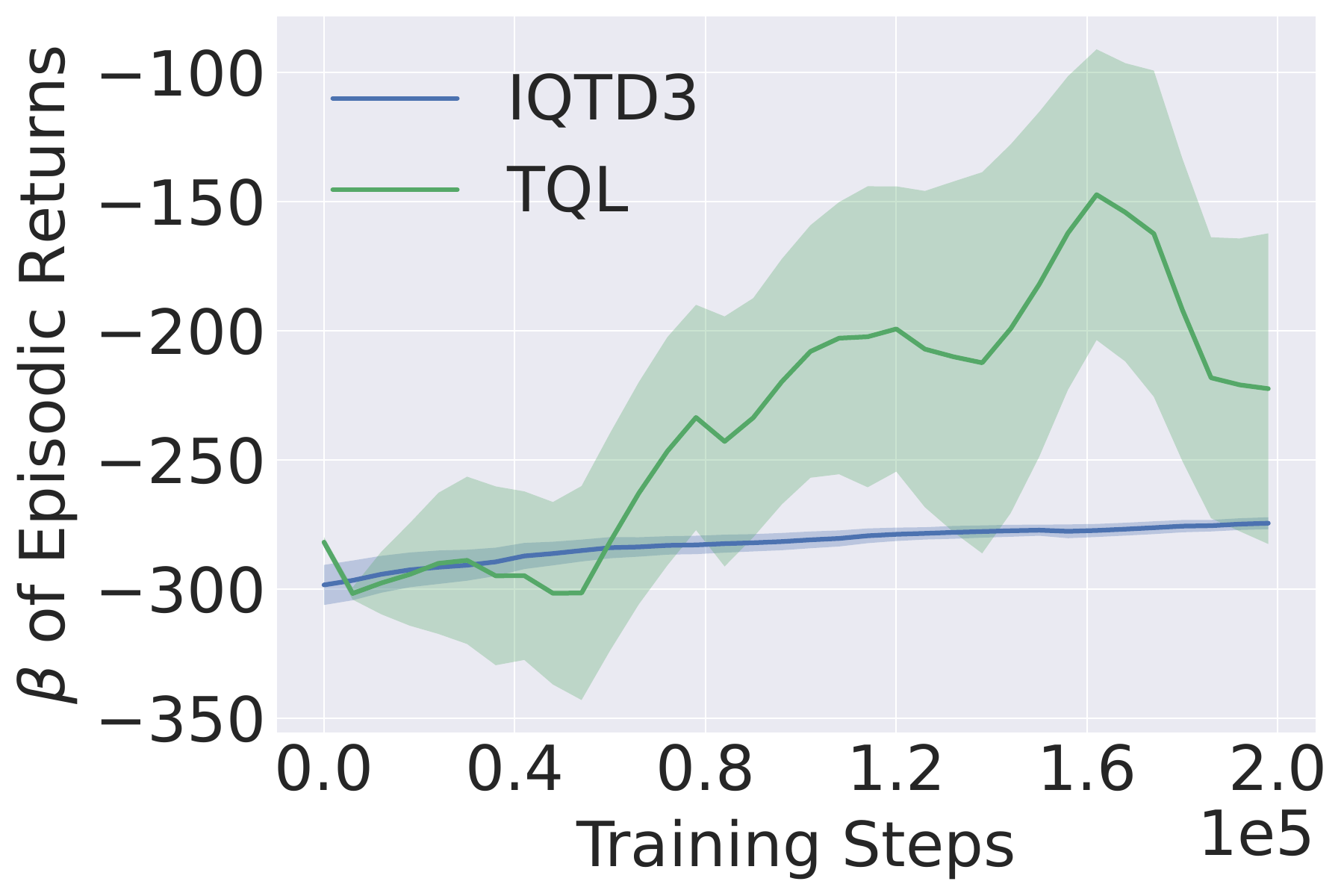}
        \label{fig:mcc-0.5-cvar}
        \vspace{-10pt}
    \end{subfigure}~
    \begin{subfigure}[b]{0.18\textwidth}
        \includegraphics[width=\textwidth]{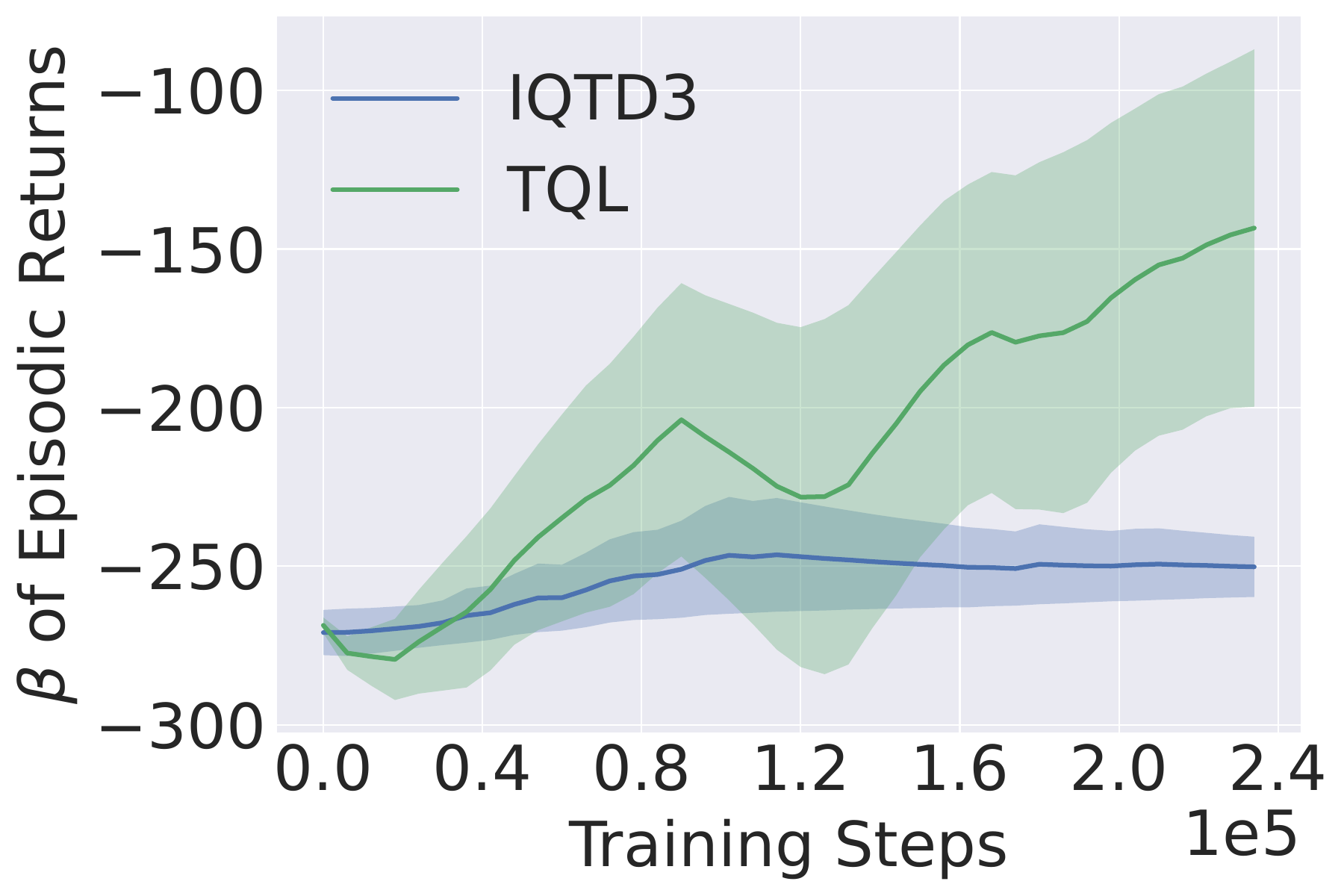}
        \label{fig:mcc-0.5-cpw}
        \vspace{-10pt}
    \end{subfigure}~
    \begin{subfigure}[b]{0.18\textwidth}
        \includegraphics[width=\textwidth]{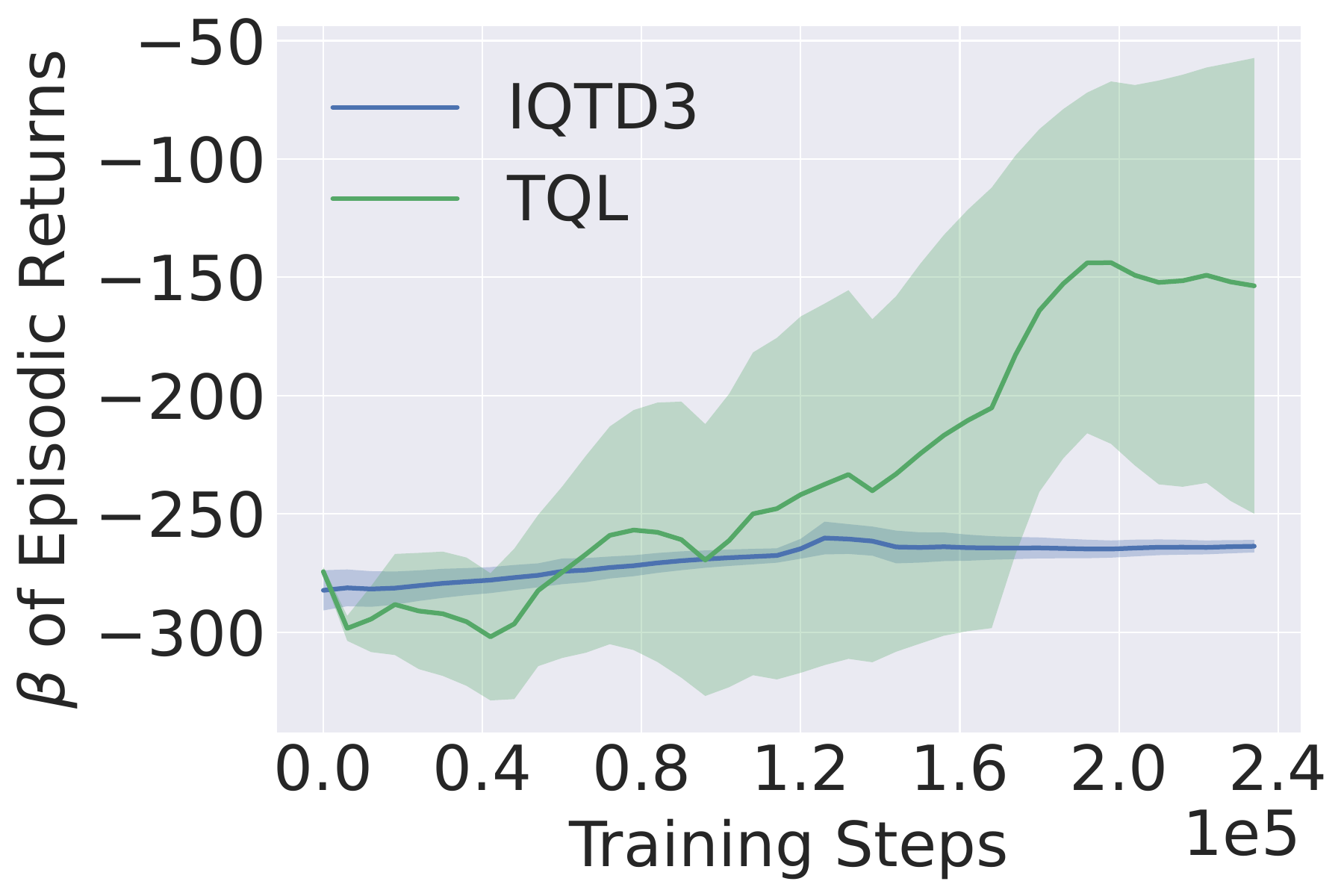}
        \label{fig:mcc-0.5-pow}
        \vspace{-10pt}
    \end{subfigure}~
    \begin{subfigure}[b]{0.18\textwidth}
        \includegraphics[width=\textwidth]{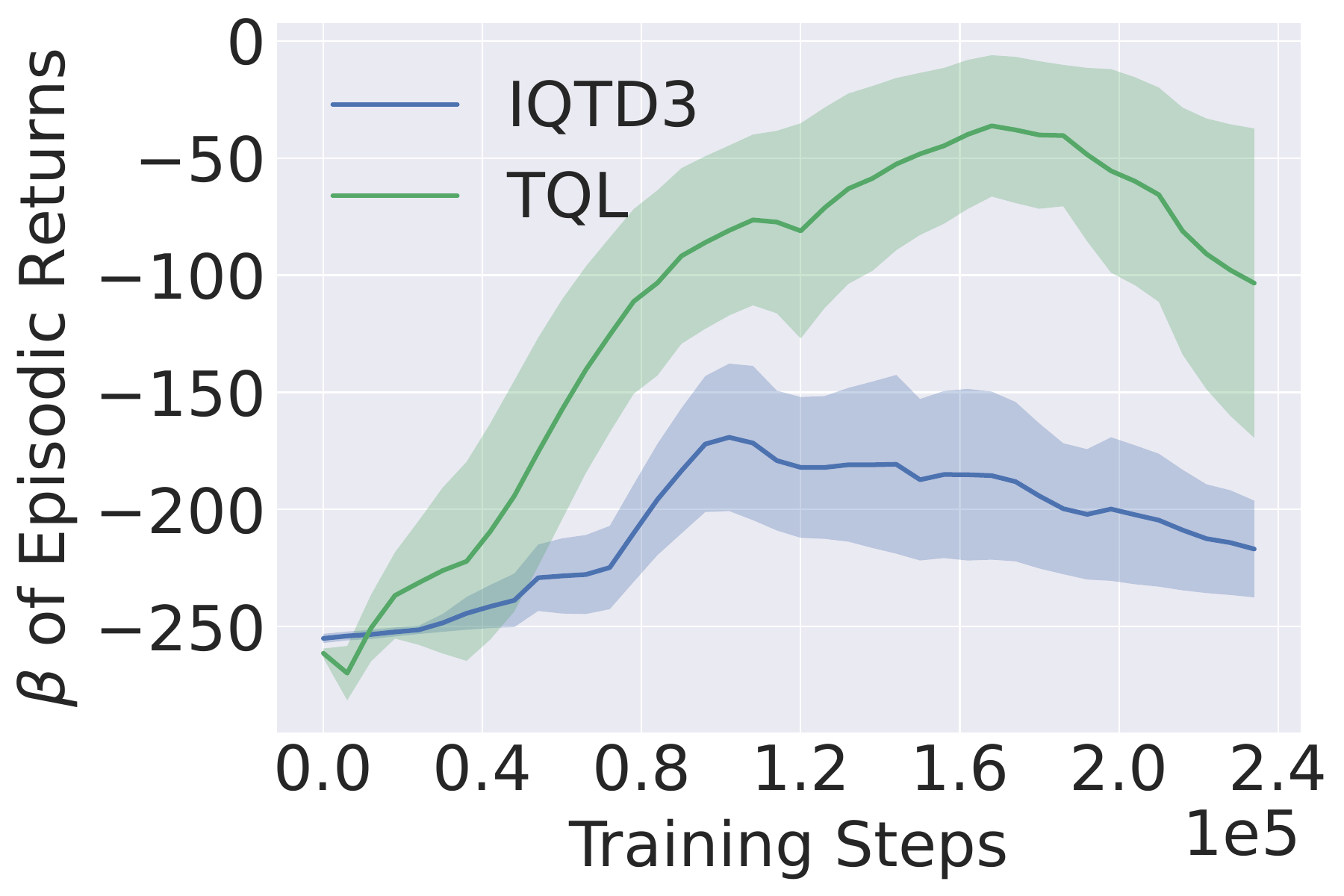}
        \label{fig:mcc-0.5-wang-pos}
        \vspace{-10pt}
    \end{subfigure}~
    \begin{subfigure}[b]{0.18\textwidth}
        \includegraphics[width=\textwidth]{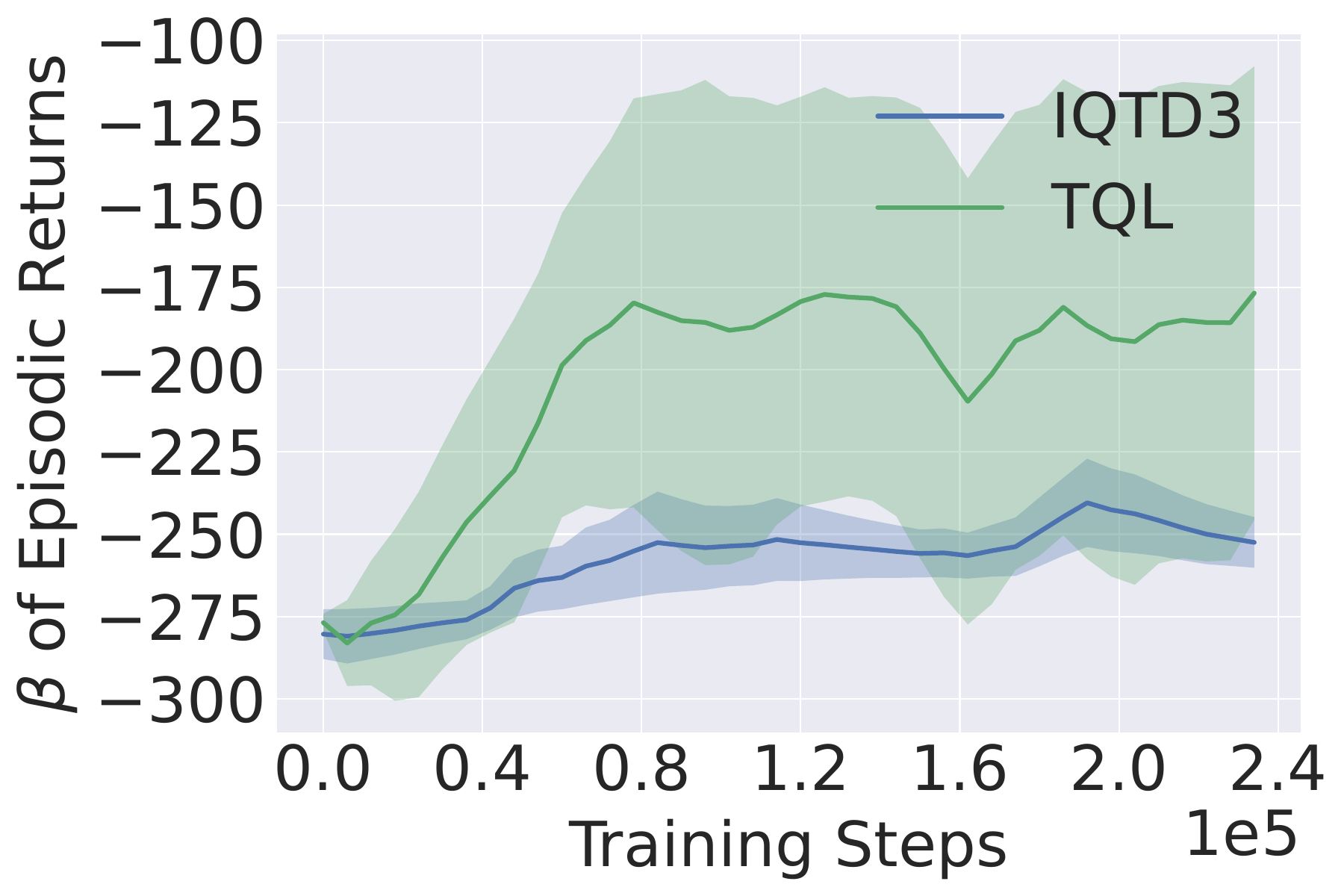}
        \label{fig:mcc-0.5-wang-neg}
        \vspace{-10pt}
    \end{subfigure}
    \end{minipage}\\
    \rotatebox{90}{\scriptsize{$c=0.75$~~}}~~~~~~
    \begin{minipage}{0.95\textwidth}
    \begin{subfigure}[b]{0.18\textwidth}
        \includegraphics[width=\textwidth]{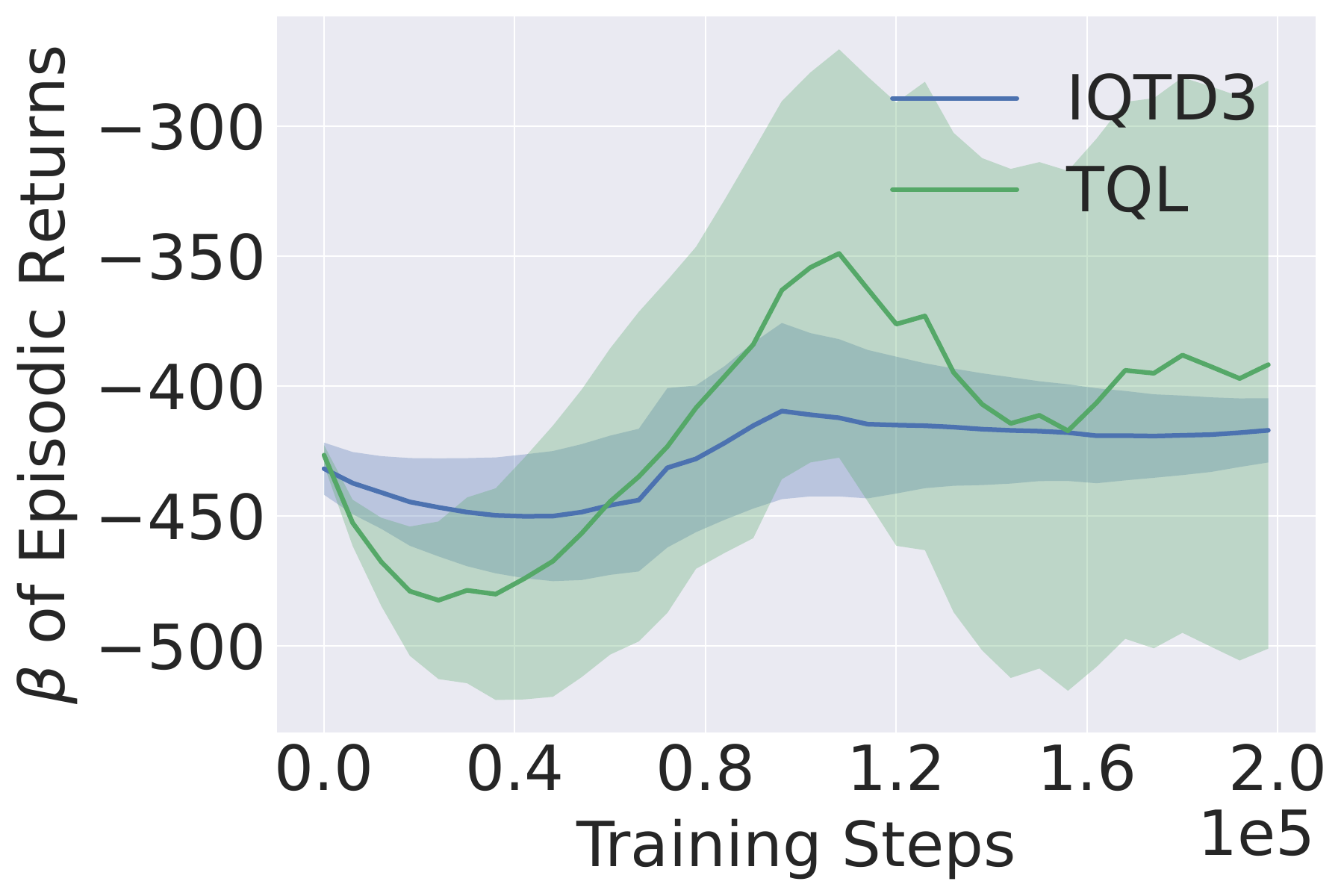}
        \label{fig:mcc-0.75-cvar}
        \vspace{-10pt}
    \end{subfigure}~
    \begin{subfigure}[b]{0.18\textwidth}
        \includegraphics[width=\textwidth]{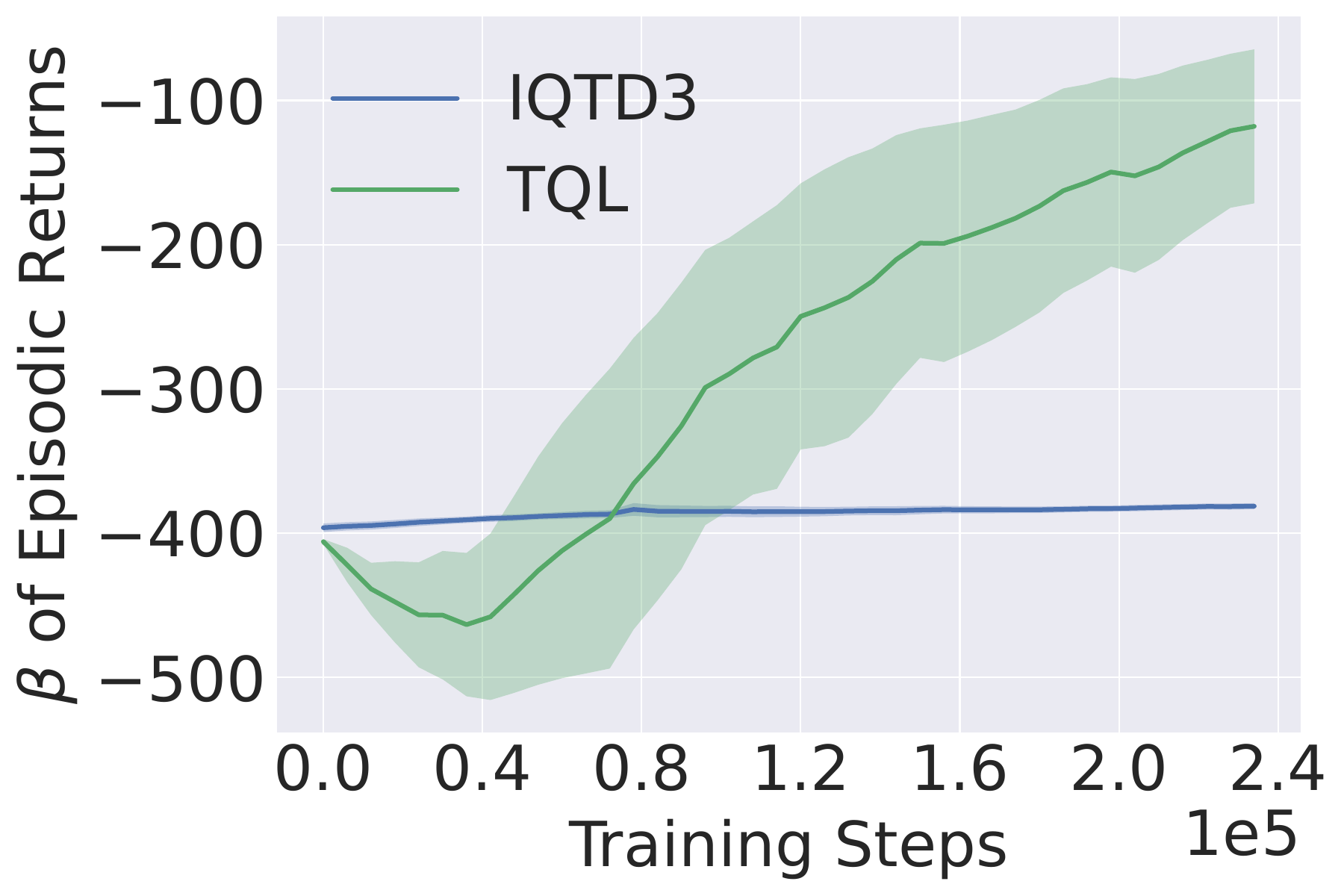}
        \label{fig:mcc-0.75-cpw}
        \vspace{-10pt}
    \end{subfigure}~
    \begin{subfigure}[b]{0.18\textwidth}
        \includegraphics[width=\textwidth]{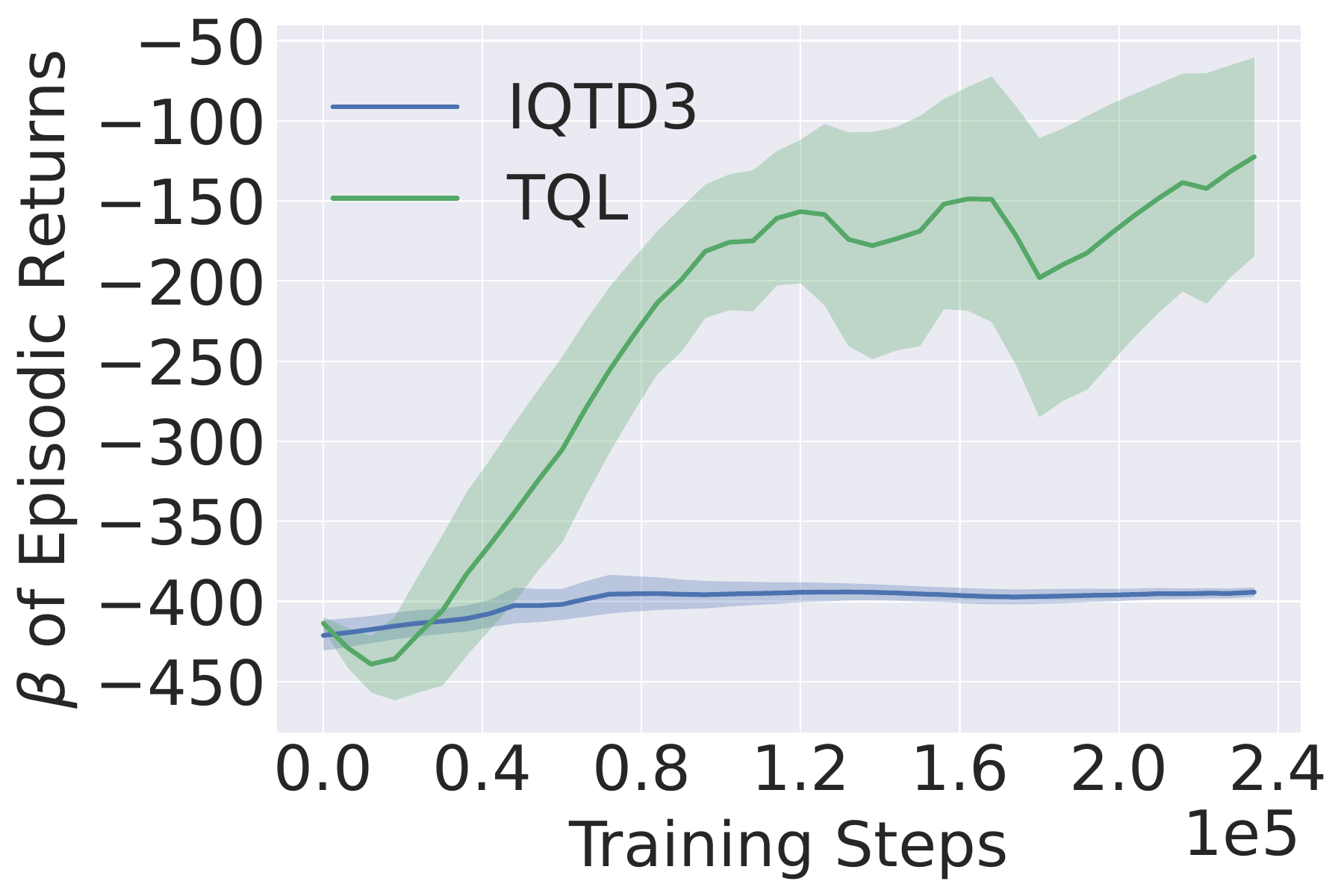}
        \label{fig:mcc-0.75-pow}
        \vspace{-10pt}
    \end{subfigure}~
    \begin{subfigure}[b]{0.18\textwidth}
        \includegraphics[width=\textwidth]{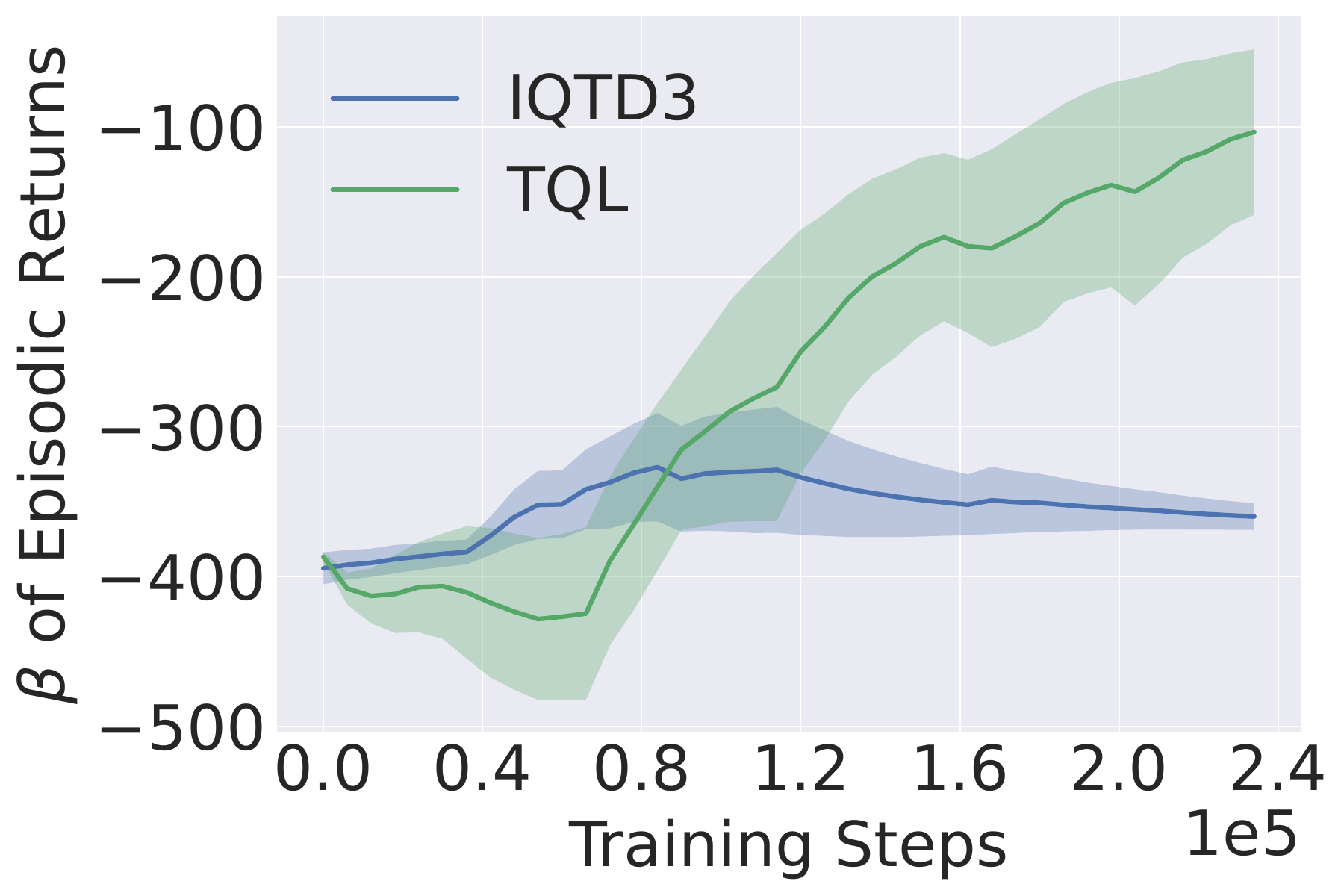}
        \label{fig:mcc-0.75-wang-pos}
        \vspace{-10pt}
    \end{subfigure}~
    \begin{subfigure}[b]{0.18\textwidth}
        \includegraphics[width=\textwidth]{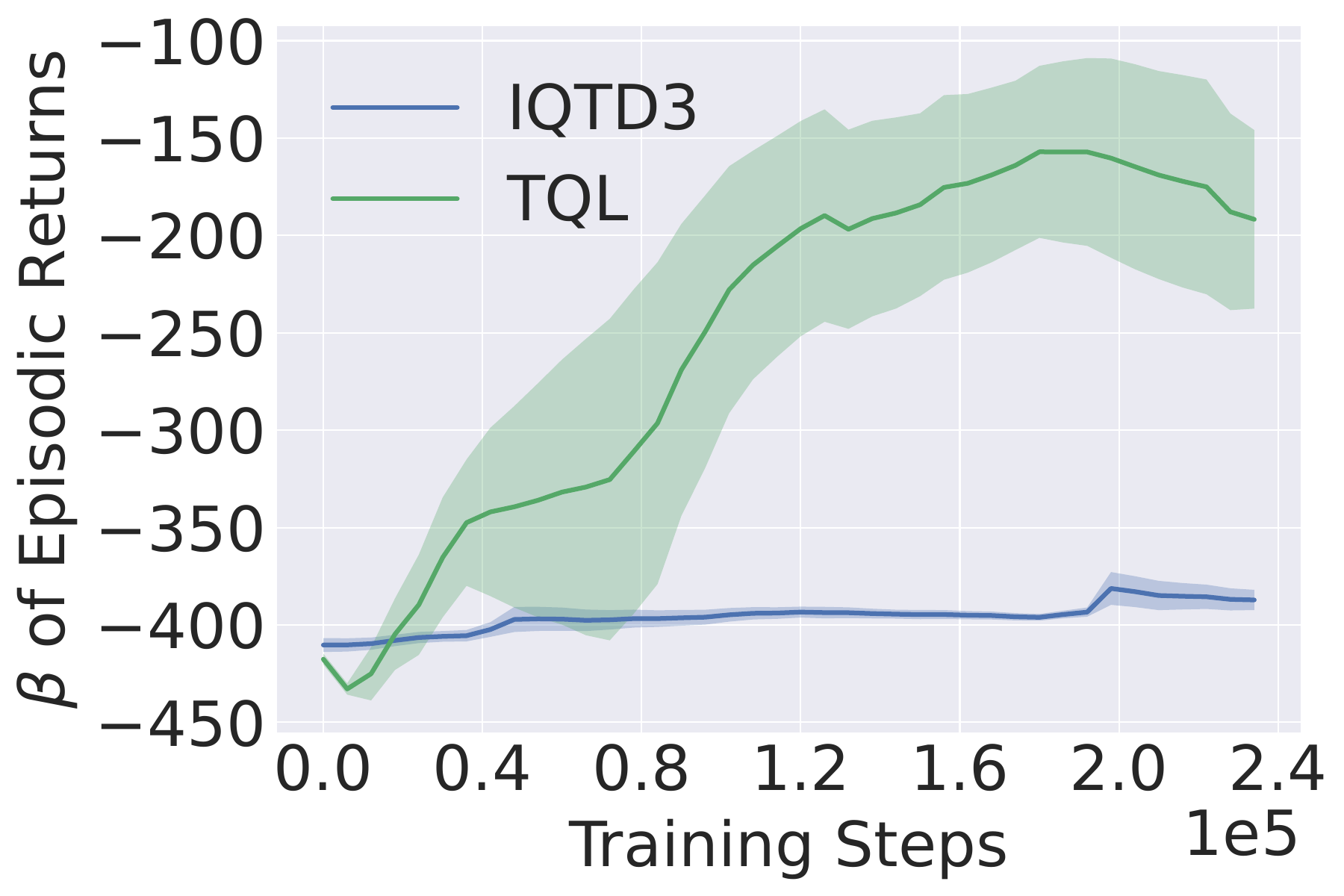}
        \label{fig:mcc-0.75-wang-neg}
        \vspace{-10pt}
    \end{subfigure}
    \end{minipage}\\
    \rotatebox{90}{\scriptsize{$c=1.0$~~}}~~~~~~
    \begin{minipage}{0.95\textwidth}
    \begin{subfigure}[b]{0.18\textwidth}
        \includegraphics[width=\textwidth]{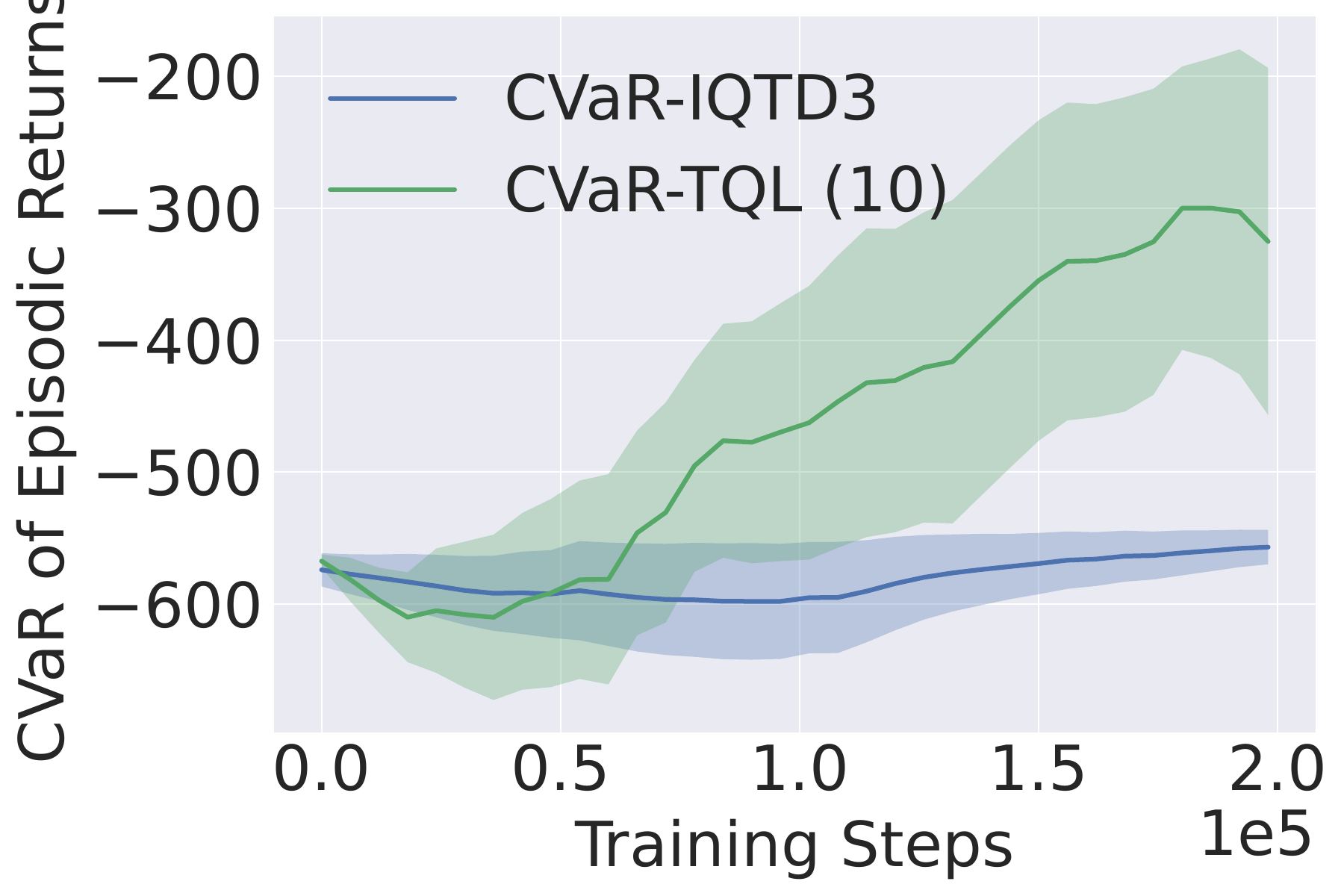}
        \vspace{-10pt}
        \caption*{\texttt{CVaR(0.25)}}
        \label{fig:mcc-1.0-cvar}
        \vspace{4pt}
    \end{subfigure}~
    \begin{subfigure}[b]{0.18\textwidth}
        \includegraphics[width=\textwidth]{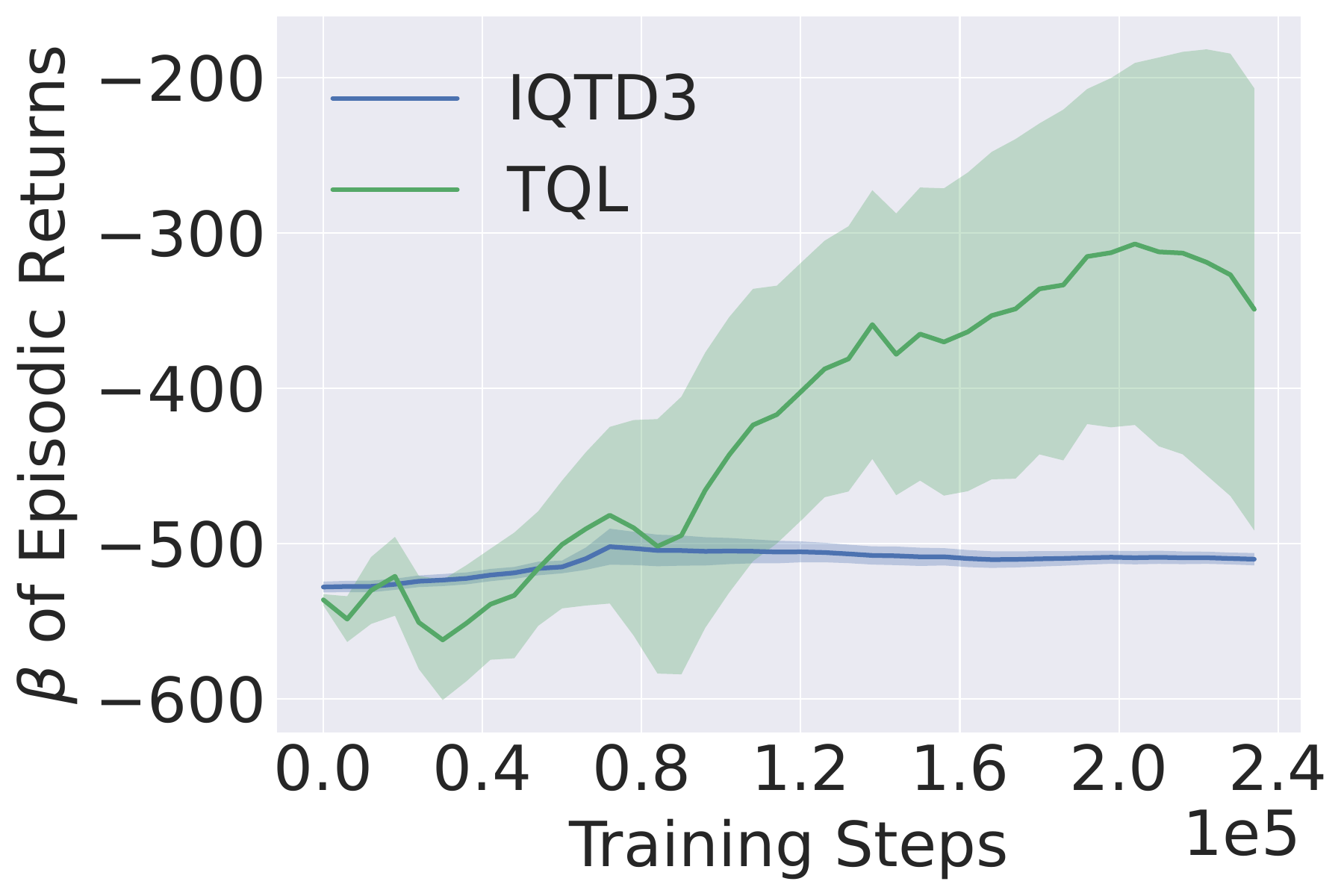}
        \vspace{-10pt}
        \caption*{\texttt{CPW(0.71)}}
        \label{fig:mcc-1.0-cpw}
        \vspace{4pt}
    \end{subfigure}~
    \begin{subfigure}[b]{0.18\textwidth}
        \includegraphics[width=\textwidth]{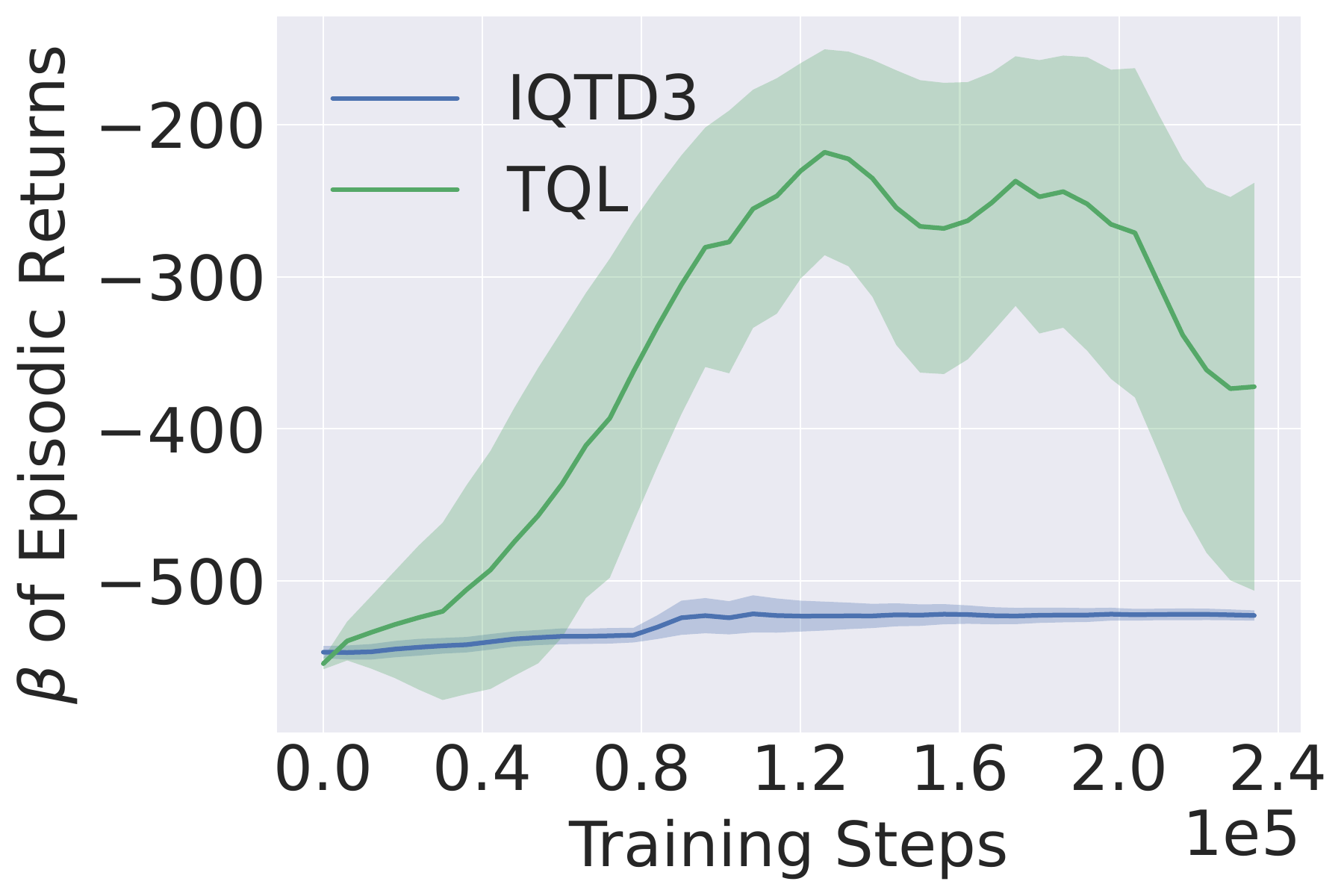}
        \vspace{-10pt}
        \caption*{\texttt{POW(-2.0)}}
        \label{fig:mcc-1.0-pow}
        \vspace{4pt}
    \end{subfigure}~
    \begin{subfigure}[b]{0.18\textwidth}
        \includegraphics[width=\textwidth]{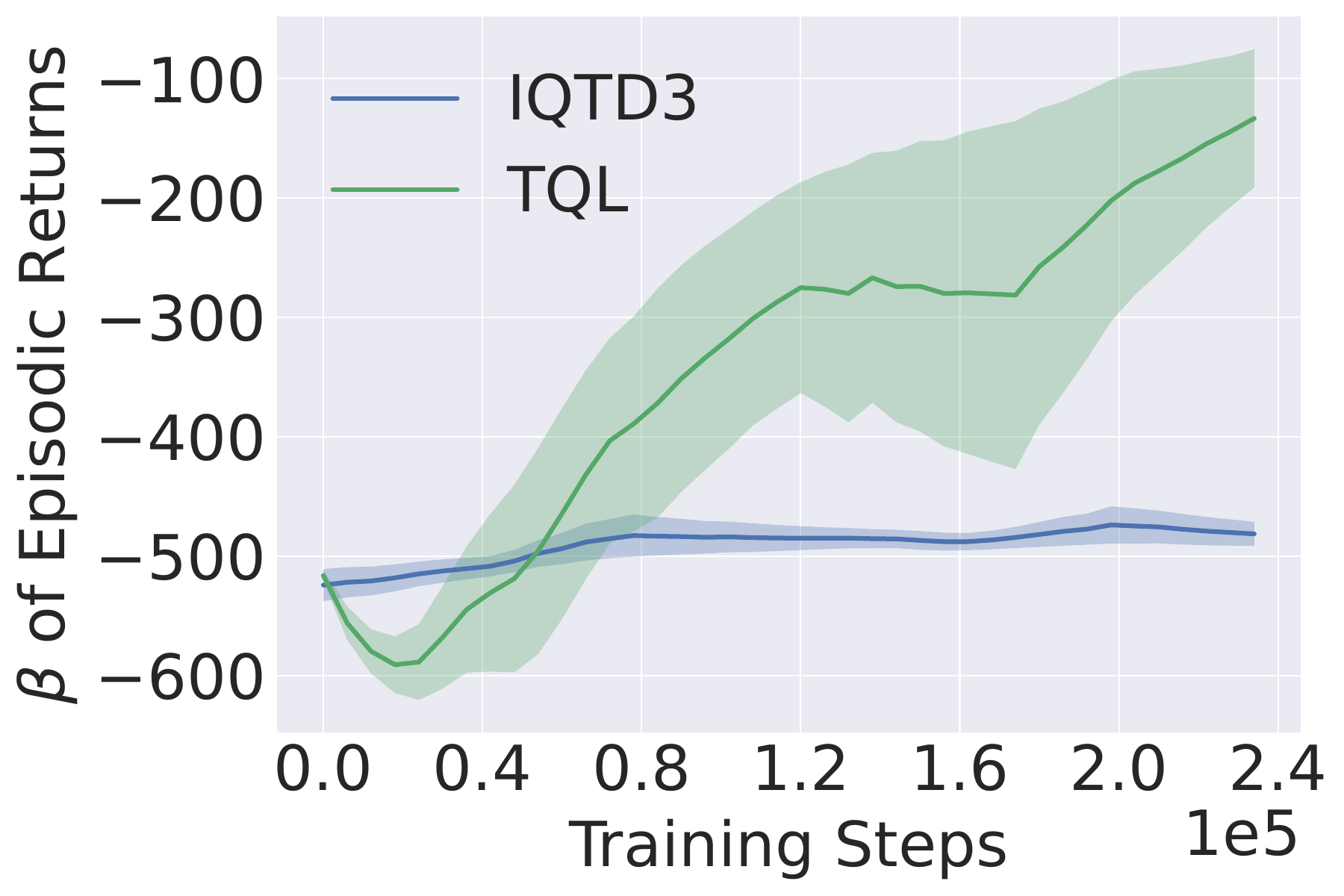}
        \vspace{-10pt}
        \caption*{\texttt{Wang(0.75)}}
        \label{fig:mcc-1.0-wang-pos}
        \vspace{4pt}
    \end{subfigure}~
    \begin{subfigure}[b]{0.18\textwidth}
        \includegraphics[width=\textwidth]{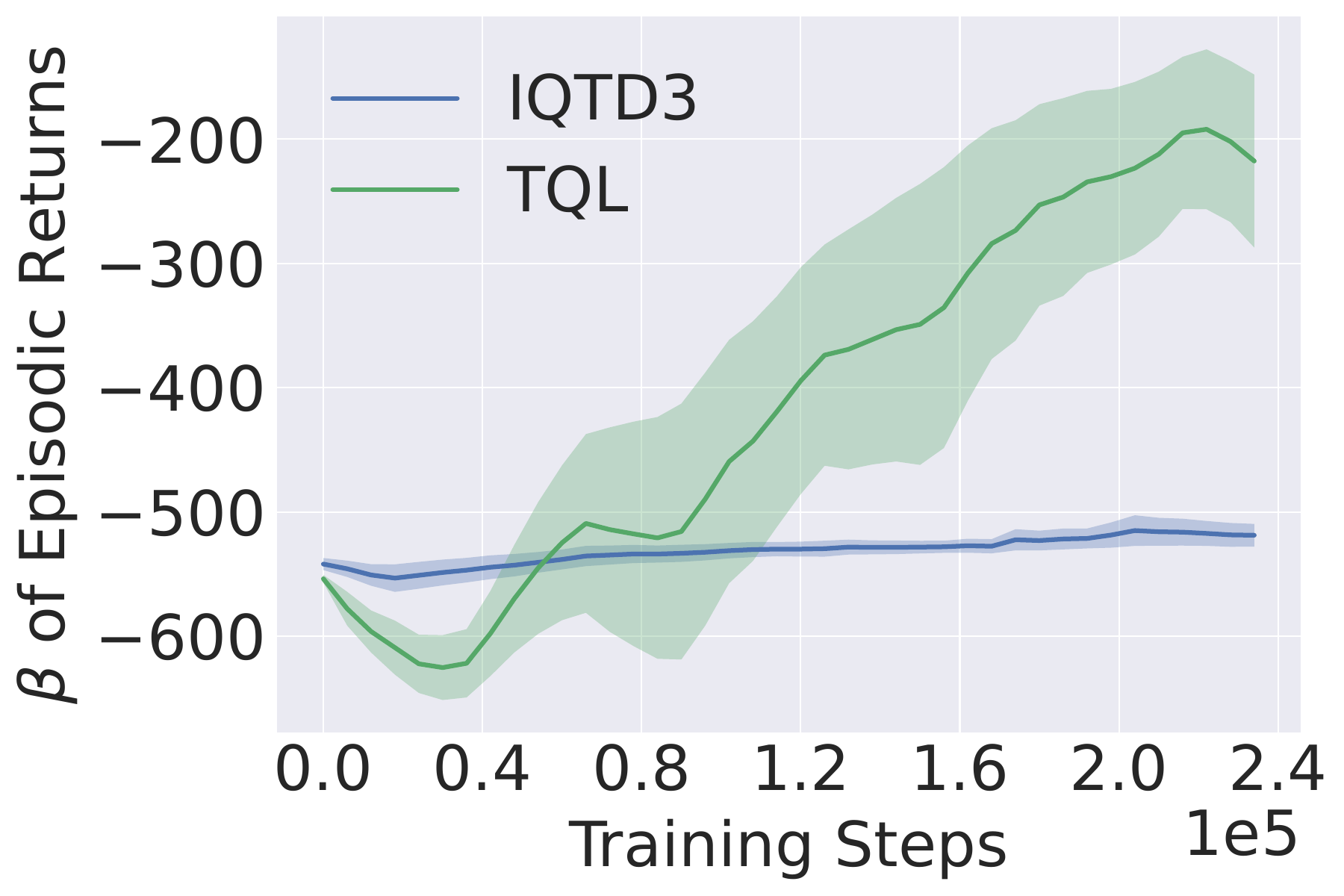}
        \vspace{-10pt}
        \caption*{\texttt{Wang(-0.75)}}
        \label{fig:mcc-1.0-wang-neg}
        \vspace{4pt}
    \end{subfigure}
    \end{minipage}\\
    \vspace{-10pt}
    \caption{Learning curves on modified Mountain-Car environment with different risk measures as objective, measured by risk measures. TQL shows a significant performance advantage over the IQTD3 baseline when exposed to larger risks ($c\in\{0.5, 0.75, 1.0\}$), and exhibits a comparable performance to baselines given smaller risks ($c\in\{0.25, 0.1, 0.0\}$).}
    \vspace{-4pt}
    \label{fig:mcc-results}
\end{figure*}

\minisection{Continuous control evaluations.}
The learning curves of our proposed TQL and baseline IQTD3 algorithm towards different distortion risk measure as objective and under different risk coefficient $c$ are shown in \fig{fig:mcc-results}. In general, when the potential risk is larger (i.e., risky penalty with larger coefficient $c\in\{0.5, 0.75, 1.0\}$), TQL obtains significantly higher risk-sensitive return than IQTD3. The Markovian policy learned by IQTD3 can hardly find out how to gain high risk-sensitive return within the control task due to its short-sighted decision-making, while TQL consistently learns a better policy. When the risk is smaller, namely $c\in\{0.25, 0.1, 0.0\}$, the difference between TQL and IQTD3 gradually diminishes, and both algorithms can learn a decent risk-sensitive policy.

\section{Limitations}
In this paper, we are the first to pose the biased objective problem in the optimization of previous risk-sensitive reinforcement learning (RSRL) algorithms. Accordingly, we propose Trajectory Q-Learning (TQL), as a general solution for RSRL, and verify its effectiveness, both theoretically and empirically. Although we believe this solution is general to a large set of RSRL problems and can be inspiring to the following study on RSRL, it still has limitations mainly in its assumption and efficiency.

\vspace{6pt}\minisection{Deterministic transition assumption.} Our solution assumes deterministic environmental dynamics, which implies a limitation the uncertainty only comes from the randomness of the reward function. In some cases, it is possible to translate a problem with stochastic dynamics structures into one with deterministic dynamics and random rewards. However, RSRL with stochastic dynamics is valuable and worth studying. Regarding this, we point out that a potential solution can be building a transition model and sampling transitions for policy optimization similar to \cite{rigter20231r2r}, combined with the proposed method in this work. Due to the workload and the value of the problem, we leave it as future work and make this work a good start for a more general solution for RSRL.

\vspace{6pt}\minisection{Efficiency and scalability.} To achieve theoretical optimality, our solution has to include historical trajectory in the input of value and policy function, which results in much larger search space than sub-optimal DP style algorithms. To mitigate the effect of increased complexity on TQL's efficiency and scalability, we apply a rolling window of length $T=10$ in the long-horizon MountainCar task to get a finite-step input sequence (more details can be referred to \ap{ap:rnn-value}), and more implementation tricks can be taken to further improve practical performance. Nonetheless, our TQL algorithm still suffers heavily from the dependence on history input, and is generally limited to evaluation on simple and toy environments. In the future we will try our best to apply the method on more complex and realistic problems.

\section{Conclusion and Future Work}
In this paper, we first present an in-depth analysis on the biased optimization issue of the existing distributional RSRL methods, and show that the issue comes from a lack of history return distribution in value and policy modeling.
Correspondingly, we introduce the historical value function that incorporate the history return distribution, and define risk-sensitive HR optimality for value and policy update.
Upon our theoretical results, we propose Trajectory Q-Learning (TQL), a distributional RL algorithm for learning the optimal policy in RSRL. We justify the theoretical properties of TQL and prove it converges to the optimal solution. 
Our experiments and the detailed analysis on both discrete and continuous control tasks validate the advantage of TQL in risk-sensitive settings. 
In future work, we plan to extend TQL to more complex tasks and real-world applications.

\begin{acks}
We thank Zhengyu Yang, Ming Zhou and Zheyuan Hu for their helpful discussions. The SJTU team is supported by ``New Generation of AI 2030'' Major Project (2018AAA0100900), Shanghai Municipal Science and Technology Major Project (2021SHZDZX0102) and National Natural Science Foundation of China (62076161).
The author Minghuan Liu is also supported by Wu Wen Jun Honorary Doctoral Scholarship, AI Institute, Shanghai Jiao Tong University. 
We sincerely thank all anonymous reviewers for their helpful feedback to revise our first manuscript.
\end{acks}

\bibliographystyle{ACM-Reference-Format}
\bibliography{reference}


\newpage
\onecolumn
\appendix
\newpage
\appendix

\section{Additional Backgrounds}
\subsection{Typical Distortion Risk Measures}\label{ap:exp-vs-distortion}
Distortion risk measure is a large family including various measures. Here we briefly introduce how we derive those used in our experiments from its framework and their properties.
\begin{itemize}[leftmargin=*]
    \item \texttt{Mean} is obtained by an identity distortion function:
    \begin{equation}
        h_{\tt mean}^{-1}\left(\tau\right)=\tau,\quad\forall\tau\in\left[0,1\right]~.
    \end{equation}
    \texttt{Mean} treats each quantile equally and serves as a risk-neutral measure, which indicates the unconditioned overall performance of a random variable.
    \item $\eta$-\texttt{CVaR} is obtained by a linear projection of fractions:
    \begin{equation}
        h_{\eta-{\tt CVaR}}^{-1}\left(\tau\right)=\eta\tau~.
    \end{equation}
    \texttt{CVaR} is always risk-averse, and smaller $\eta$ makes it more conservative.
    \item $\eta$-\texttt{Wang} is a simple measure that controls risk preference by translating the c.d.f. of standard Gaussian:
    \begin{equation}
        h_{\eta-{\tt Wang}}^{-1}\left(\tau\right)=\Phi(\Phi^{-1}(\tau)+\eta)~.
    \end{equation}
    where $\Phi$ is the c.d.f. of standard Gaussian distribution. Positive $\eta$ corresponds to risk-seeking, and negative $\eta$ corresponds to risk-aversity.
    \item $\eta$-\texttt{CPW} is introduced in \citet{tversky1992advances}, it is neither globally risk-averse nor risk-seeking.
    \begin{equation}
        h_{\eta-{\tt CPW}}^{-1}\left(\tau\right)=\frac{\tau^\eta}{\left(\tau^\eta+(1-\tau)^\eta\right)^{1/\eta}}~.
    \end{equation}
    \citet{wu1996curvature} proposed that $\eta=0.71$ matches human subjects well.
    \item $\eta$-\texttt{POW} is proposed in \citet{dabney2018iqn}. It is a simple power formula for risk-averse ($\eta<0$) or risk-seeking ($\eta>0$) policies:
    \begin{equation}
        h_{\eta-{\tt POW}}^{-1}\left(\tau\right)=\left\{\begin{array}{ll}
            \tau^{\frac{1}{1+|\eta|}}, & \text{if }\eta\ge 0 \\
            1-(1-\tau)^{\frac{1}{1+|\eta|}} & \text{otherwise }
        \end{array}\right..
    \end{equation}
\end{itemize}

\subsection{Quantile Regression and Quantile Huber Loss}\label{ap:qr-loss}
\minisection{Quantile regression.} \citet{dabney2018qrdqn} first to approximate the return distribution with quantiles and optimize the value function with quantile regression \citep{koenker2001qr}, which uses the quantile regression loss to estimate quantile functions of a distribution.

The quantile regression loss is an asymmetric convex loss function that penalizes overestimation and underestimation errors with different weights. For a distribution $Z$ with c.d.f. $F_Z(z)$, and a given quantile $\tau$, the value of quantile function $F_Z^{-1}(\tau)$ can be characterized as the minimizer of the following \emph{quantile regression loss}:
\begin{align}
    \caL_{\rm QR}^\tau(\theta):=\bbE_{\hat{z}\sim Z}\left[\rho_\tau(\hat{z}-\theta)\right]~,\text{ where}&\nonumber\\
    \rho_\tau(u)=u(\tau-\delta_{u<0})~,~\forall u\in\bbR&
\end{align}

\minisection{Quantile Huber loss.} \citet{dabney2018qrdqn} found that the non-smoothness at zero could limit performance when using non-linear function approximation, hence proposed a modified quantile loss, namely the \emph{quantile Huber loss}.

The Huber loss is given by \citet{huber1964robust}:
\begin{equation}
    \caL_\kappa(u)=\left\{
    \begin{array}{ll}
        \frac{1}{2}u^2 & \text{if }|u|\le\kappa \\
        \kappa(|u|-\frac{1}{2}\kappa) & \text{otherwise}
    \end{array}
    \right.~,
\end{equation}
and the quantile Huber loss is then an asymmetric version of the Huber loss:
\begin{equation}
    \rho^\kappa_\tau(u)=|\tau-\delta_{u<0}|\caL_\kappa(u)~.
\end{equation}

\subsection{Comparison of Our Work to \citep{lim2022distributional}}
\label{ap:compare}

The recent work of \cite{lim2022distributional} imposes a similar issue of optimizing \texttt{CVaR} in previous RSRL methods, and proposes modifications to existing algorithm methods that use a moving threshold for episodic \texttt{CVaR} estimation and include a new distributional Bellman operator. They show that the optimal \texttt{CVaR} policy corresponds to the fixed point of their proposed distributional Bellman operator.

However, \cite{lim2022distributional} does not show the convergence property of their proposed distributional Bellman operator, and their solution is available specifically for \texttt{CVaR} policy optimization. Furthermore, their proposed moving threshold is approximated with the learned return distribution, which also impacts the ability to find the optimal policy. In contrast, our proposed TQL is provided with convergence property and enables optimization under all kinds of risk measures. Experiments in \se{subsec:exp-results} also show the advantage of TQL over \citep{lim2022distributional} in practical tasks.

\subsection{Static Risk vs. Dynamic Risk}
\label{ap:dynamic-vs-static}
In this paper, we consider static risk measures $\beta$ over the whole trajectory. However, some other works~\cite{bauerle2011markov,chow2014cvar,tamar2015policy} consider dynamic risk measures $\rho$, which is defined recursively over a trajectory $\tau=(s_0,a_0,s_1,a_1,\cdots)$:
\begin{equation}
    \rho[\tau]=\rho\left[R(s_0,a_0)+\gamma\rho\left[R(s_1,a_1)+\gamma\rho[R(s_2,a_2)+\cdots]\right]\right]
\end{equation}

Dynamic risk measures have the advantage of time consistency, and \cite{bauerle2011markov,chow2014cvar,tamar2015policy} has conducted an in-depth analysis of the policy optimization based on dynamic risk measures. However, dynamic risk measures are more short-sighted and hard to estimate in practical tasks due to their per-step definition, and in the real-world people care more about the overall risk over the whole decision process. Therefore, static risk measures are most commonly used for evaluation in fields such as finance and medical treatments. Furthermore, the control problem over static risk measures have been shown to be non-Markovian and non-stationary \cite{bellemare2023distributional}, hence it can be hardly resolved by DP style RSRL algorithms, which motivates the use of historical return distribution and history-dependent policy in our TQL algorithm.

\section{Algorithms}
\label{ap:algo}
\minisection{Discrete control.} We present the step-by-step algorithm of TQL for the policy with discrete actions in 
\alg{alg:tql-disc}.

\minisection{Continuous control.} We present the step-by-step algorithm of TQL for continuous control in \alg{alg:tql-cont}.

\vspace{-6pt}
\begin{minipage}[t]{0.48\textwidth}
\begin{algorithm}[H]
   \caption{Trajectory Q-Learning (TQL) (Discrete)}
   \label{alg:tql-disc}
\begin{algorithmic}
    \STATE {\bfseries Input:} Parameter vectors $\theta$, $\psi$, and Risk Measure $\beta$
    \STATE Initialize parameters $\theta$, $\psi$, and replay buffer $\caD$.
    \FOR{each iteration}
        \FOR{each environment step}
        \STATE $a_t\sim\mathop{\arg\max}_a \beta[Z(h_t,a)]$
        \STATE $s_{t+1}\sim p\left(s_{t+1}|s_t,a_t\right)$
        \STATE $\caD\leftarrow\caD\cup\left\{\left(h_t,a_t,r\left(s_t,a_t\right),s_{t+1}\right)\right\}$
        \ENDFOR
        \FOR{each gradient step}
        \STATE $\psi\leftarrow\psi-\nabla_\psi J_Q(\psi)$ ~~~~(\eq{eqn:loss-Q-markovian})
        \STATE $\theta\leftarrow\theta-\nabla_\theta J_Q(\theta)$ ~~~~~~~(\eq{eqn:loss-Q}) 
        \ENDFOR
    \ENDFOR
\end{algorithmic}
\end{algorithm}
\end{minipage}\hfill
\begin{minipage}[t]{0.48\textwidth}
\begin{algorithm}[H]
   \caption{Trajectory Q-Learning (TQL) (Continuous)}
   \label{alg:tql-cont}
\begin{algorithmic}
    \STATE {\bfseries Input:} Parameter vectors $\phi$, $\theta$, $\psi$, and Risk Measure $\beta$
    \STATE Initialize parameters $\phi$, $\theta$, $\psi$, and replay buffer $\caD$.
    \FOR{each iteration}
        \FOR{each environment step}
        \STATE $a_t\sim\pi_\phi\left(a_t|h_t\right)$
        \STATE $s_{t+1}\sim p\left(s_{t+1}|s_t,a_t\right)$
        \STATE $\caD\leftarrow\caD\cup\left\{\left(h_t,a_t,r\left(s_t,a_t\right),s_{t+1}\right)\right\}$
        \ENDFOR
        \FOR{each gradient step}
        \STATE $\psi\leftarrow\psi-\nabla_\psi J_Q(\psi)$ ~~~~(\eq{eqn:loss-Q-markovian})
        \STATE $\theta\leftarrow\theta-\nabla_\theta J_Q(\theta)$ ~~~~~~~(\eq{eqn:loss-Q})
        \STATE $\phi\leftarrow\phi-\nabla_\phi J_\pi(\phi)$ ~~~~~~(\eq{eqn:loss-pi})
        \ENDFOR
    \ENDFOR
\end{algorithmic}
\end{algorithm}
\end{minipage}

\section{Proof}\label{ap:proof}
\subsection{Proof of Lemma \ref{lemma:value-iteration}}\label{ap:value-iteration}
\begin{lemma}[Value iteration theorem]
Recursively applying the distributional Bellman optimality operator $\caT^*Z_{k+1}=Z_{k}$ on an arbitrary value distribution $Z_0$ solves \eq{eqn:risk-sensitive-policy} when $\beta$ is exactly \texttt{mean} where the optimal policy is obtained via \eq{eqn:conventional-mean-policy}, and for $Z_1, Z_2 \in \caZ$, we have:
\begin{equation}
    \|\bbE[\caT^*Z_{1}]-\bbE[\caT^*Z_{2}]\|_\infty\le\gamma\|\bbE[Z_{1}]-\bbE[Z_{2}]\|_\infty~,
\end{equation}
and in particular $\bbE[Z_k] \rightarrow \bbE[Z^*]$ exponentially quickly.

\begin{proof}
The RSRL objective for \texttt{mean} policy is
\begin{equation}
    \mathop{\arg\max}_{a_0,\cdots,a_{T-1}}~\mathbb{E}\left[Z(s_0,a_0)\right]~.
\end{equation}
From the Bellman equation, we can unfold $Z(s_0,a_0)$ and get an equivalent form:
\begin{equation}\label{eqn:value-iteration-1}
    \mathop{\arg\max}_{a_0,\cdots,a_{T-1}}~\mathbb{E}\left[R(s_0,a_0)+\gamma Z(s_1,a_1)\right]~,\text{ where }S_1\sim\caP(s_1|s_0,a_0).
\end{equation}
Notice that \texttt{mean} is linearly additive, hence \eq{eqn:value-iteration-1} can be further divided into two parts of optimization:
\begin{equation}
    \mathop{\arg\max}_{a_0}~\mathbb{E}\left[R(s_0,a_0)+\gamma\max_{a_1,\cdots,a_{T-1}}~\mathbb{E}\left[Z(s_1,a_1)\right]\right]~.
\end{equation}
Repeating this process on $Z(s_1,a_1)$ and further, we will obtain the dynamic programming objective as following:
\begin{equation}
\begin{aligned}
&\mathop{\arg\max}_{a_0}~\mathbb{E}\left[R(s_0,a_0)+\gamma\max_{a_1,\cdots,a_{T-1}}~\mathbb{E}\left[Z(s_1,a_1)\right]\right]\\
\Leftrightarrow~&\mathop{\arg\max}_{a_0}~\mathbb{E}\left[R(s_0,a_0)+\gamma Z\left(S_1,\mathop{\arg\max}_{a_1}\mathbb{E}\left[R(s_1,a_1)\right.\right.\right.\\
&~~~~+\left.\left.\left.\gamma\max_{a_2,\cdots,a_{T-1}}\mathbb{E}\left[Z(s_2,a_2)\right]\right]\right)\right]\\
\Leftrightarrow~&\cdots
\end{aligned}
\end{equation}
which will finally get to the single-step target.

For $Z_1, Z_2 \in \caZ$, using the linearly additive property of \texttt{mean}, we have
\begin{align}
    \|\bbE~[\caT^*_D Z_{1}]-\bbE~[\caT^*_D Z_{2}]\|_\infty&=\|\caT^*_E~\bbE[Z_{1}]-\caT^*_E~\bbE[Z_{2}]\|_\infty\nonumber\\
    &\le\gamma\|\bbE[Z_{1}]-\bbE[Z_{2}]\|_\infty
\end{align}
where $\caT^*_D$ denotes the distributional operator and $\caT^*_E$ denotes the usual operator.
\end{proof}
\end{lemma}

\subsection{Proof of Theorem \ref{theo:no-contraction-theorem}}\label{ap:no-contraction-theorem}
\begin{theorem}
Recursively applying risk-sensitive Bellman optimality operator $\caT_\beta^*$ w.r.t. risk measure $\beta$ does not solve the 
RSRL objective \eq{eqn:risk-sensitive-policy} and $\beta [Z_k]$ is not guaranteed to converge to $\beta [Z^*]$ if $\beta$ is not \texttt{mean}.
\begin{proof}
    
    We want to show that $$\left\|\beta\left[\caT_{\beta}^*Z_1(s,a)\right]-\beta\left[\caT_{\beta}^*Z_2(s,a)\right]\right\|_{\infty}\le\gamma\left\|\beta\left[Z_1(s,a)\right]-\beta\left[Z_2(s,a)\right]\right\|_{\infty}$$ is not guaranteed to be true.
    We prove this by a counterexample. 
    Given a risk measure $\beta$, consider two value distributions $Z_1$ and $Z_2$ where $Z_1(s,a)=Z_2(s,a),~\forall s\in\caS,a\in\caA$. Assume there are more than one optimal actions $a\in\caA^*_1\subset\caA$ ($|\caA^*_1|>1$) at state $s_1$ in terms of risk measure $\beta[Z_i(s_1,\cdot)],~i\in\{1,2\}$, i.e.,
    \begin{equation*}    \forall~a^*\in\caA^*_1,~a^{\prime}\in\caA,~\beta[Z_i(s_1,a^*)]\ge\beta[Z_i(s_1,a^{\prime})]~.
    \end{equation*}
    and these optimal actions correspond to different value distributions, i.e. 
    \begin{equation*}    
    Z_i(s_1,a)\mathop{\neq}\limits^D Z_i(s,a^\prime),~\forall~a,a^\prime\in\caA^*_1,~i\in\{1,2\}~.
    \end{equation*}
    In case where we do not have a strict preference over $\caA^*_1$, the update formula for $Z_i(s_0,a_0),~i\in\{1,2\}$ where $M(s_0,a_0)=s_1$ will be
    \begin{equation}\label{eq:not-contraction-update}
        Z_i(s_0,a_0)\leftarrow R(s_0,a_0)+\gamma Z_i(s_1,a),~\forall a\in\caA^*_1~.
    \end{equation}
    When \eq{eq:not-contraction-update} uses $a=a_1\in\caA^*_1$ in the right-hand side, it holds that 
    \begin{equation*}
    Z_i(s_0,a_0)\mathop{\neq}\limits^D R(s_0,a_0)+\gamma Z_i(s_1,a_1^\prime),~a_1^\prime\in\caA^*_1\text{ and }a_1^\prime\not=a_1~.
    \end{equation*}
    Although $\beta[Z_i(s_1,a)],~\forall~a\in\caA^*_1,~i\in\{1,2\}$ are all the same, consider one update as follows:
    \begin{align*}
        Z_1^\prime(s_0,a_0)&\leftarrow R(s_0,a_0)+\gamma Z_1(s_1,a),~a\in\caA^*_1~,\\
        Z_2^\prime(s_0,a_0)&\leftarrow R(s_0,a_0)+\gamma Z_2(s_1,a^\prime),~a^\prime\in\caA^*_1\text{ and }a^\prime\not=a~.
    \end{align*}
    Notice that $\|\beta[Z_1]-\beta[Z_2]\|_\infty=0$. Since $R(s_0,a_0)$ and $Z_i(s_1,\cdot)$ can be arbitrary distributions, we can always find such $R(s_0,a_0)$ and $Z_i(s_1,\cdot)$ that $Z_1^\prime\not=Z_2^\prime$ when $\beta$ is not \texttt{mean}, as we show an example for \texttt{CVaR} as follows.
    \begin{equation*}
    \begin{small}
        \begin{array}[c]{l|ccc}
        \multicolumn{1}{c}{\beta=\texttt{CVaR}(\eta=0.1)} & \multicolumn{1}{c}{s_0,a_0} & \multicolumn{1}{c}{s_1, a_0} & \multicolumn{1}{c}{s_1, a_1} \\
        \midrule
        R & \makecell[c]{\left\{\begin{matrix}
         & 100, p=0.9 \\
         & -10, p=0.1 \\
        \end{matrix}\right.} & \makecell[c]{\left\{\begin{matrix}
         & 100, p=0.9 \\
         & -10, p=0.1 \\
        \end{matrix}\right.} & -10 \\
        \midrule
        Z_i & \makecell[c]{\left\{\begin{matrix}
         & 90, p=0.9 \\
         & -20, p=0.1 \\
        \end{matrix}\right.} & \makecell[c]{\left\{\begin{matrix}
         & 100, p=0.9 \\
         & -10, p=0.1 \\
        \end{matrix}\right.} & -10 \\
        \midrule
        \beta[Z_i] & -20 & -10 & -10 \\
        \midrule
        \midrule
        \multicolumn{4}{l}{
        Z_1^\prime(s_0,a_0)\leftarrow R(s_0,a_0)+\gamma Z_1(s_1,a_0),~~
        Z_2^\prime(s_0,a_0)\leftarrow R(s_0,a_0)+\gamma Z_2(s_1,a_1)
        }\\
        \midrule
        \midrule
        Z_1^\prime & \cellcolor{lightsunflower} \makecell[c]{\left\{\begin{matrix}
         & 200, p=0.81 \\
         & 90, p=0.18 \\
         & -20, p=0.01 \\
        \end{matrix}\right.} & \makecell[c]{\left\{\begin{matrix}
         & 100, p=0.9 \\
         & -10, p=0.1 \\
        \end{matrix}\right.} & -10 \\
        \midrule
        Z_2^\prime & \cellcolor{lightsunflower} \makecell[c]{\left\{\begin{matrix}
         & 90, p=0.9 \\
         & -20, p=0.1 \\
        \end{matrix}\right.} & \makecell[c]{\left\{\begin{matrix}
         & 100, p=0.9 \\
         & -10, p=0.1 \\
        \end{matrix}\right.} & -10 \\
        \midrule
        \beta[Z_1^\prime] & \cellcolor{lightsunflower} 79 & -10 & -10 \\
        \midrule
        \beta[Z_2^\prime] & \cellcolor{lightsunflower} -20 & -10 & -10 \\
        \bottomrule
        \end{array}
    \end{small}
    \end{equation*}
    In this case $\|\beta[Z_1^\prime]-\beta[Z_2^\prime]\|_\infty>\gamma\|\beta[Z_1]-\beta[Z_2]\|_\infty=0$.
    Therefore, $\|\beta[Z_1^\prime]-\beta[Z_2^\prime]\|_\infty\le\gamma\|\beta[Z_1]-\beta[Z_2]\|_\infty$ is not guaranteed to be true.
    
    Finally, a straight forward deduction is that, repeating using the following two formula stochastically to update $Z$ will never reach a convergence.
    \begin{align*}
        Z^\prime(s_0,a_0)&\leftarrow R(s_0,a_0)+\gamma Z(s_1,a),~a\in\caA^*_1~,\\
        Z^\prime(s_0,a_0)&\leftarrow R(s_0,a_0)+\gamma Z(s_1,a^\prime),~a^\prime\in\caA^*_1\text{ and }a^\prime\not=a~.
    \end{align*}
\end{proof}
\end{theorem}

\subsection{Proof of Theorem \ref{theo:pe-contraction}}\label{ap:pe-contraction-theorem}
\begin{theorem}[Policy Evaluation for $\caT^\pi_h$]
$\caT^\pi_h:\caZ\rightarrow\caZ$ is a $\gamma$-contraction in the metric of the maximum form of $p$-Wasserstein distance $\bar{d}_p$.

\begin{proof} 
Let's first define the transition operator $P^\pi :\caS \times \caA \to \caS$, and then extend it to $P^\pi_h:\caH\times\caA\to\caS$ by setting its output to a single-state history including only the next state as
\begin{equation*}
    P^\pi_h(h_t,a_t) = \{s_{t+1}\},~\text{where}~s_{t+1}=M(s_t,a_t)~,
\end{equation*}
and then we have
\begin{equation*}
P^\pi_h Z(h_t, a_t) \triangleq Z(\{s_{t+1}\}, a_{t+1})~,~\text{where}~a_{t+1} \sim \pi(\cdot | h_t \cup \{s_{t+1}\})~,
\end{equation*}
and $\{s_{t+1}\}$ that also falls in the space of $\caH$. Then, we have:
\begin{equation}
\begin{aligned}
&d_p\left (\caT^\pi_h Z_1(h_t,a_t), \caT^\pi_h Z_2(h_t,a_t)\right )\\
=~&d_p\left (R_{0:t}+\gamma^{t+1}P^\pi_h Z_1(h_t,a_t),
R_{0:t}+\gamma^{t+1}P^\pi_h Z_2(h_t,a_t)\right )\\
\le~&\gamma^{t+1} d_p\left (P^\pi_h Z_1(h_t,a_t),P^\pi_h Z_2(h_t,a_t)\right )\\
\le~&\gamma^{t+1}\mathop{\rm sup}_{h,a}d_p\left (P^\pi_h Z_1(h,a),P^\pi_h Z_2(h,a)\right )\\
\le~&\gamma\mathop{\rm sup}_{h,a}d_p\left (P^\pi_h Z_1(h,a),P^\pi_h Z_2(h,a)\right )~.
\end{aligned}
\end{equation}
Then it is easy to see
\begin{equation}
\begin{aligned}
&~\bar{d}_p\left (\caT^\pi_{h} Z_1(h,a), \caT^\pi Z_2(h,a)\right )\\
=&~\mathop{\rm sup}_{h,a} d_p\left (\caT^\pi_h Z_1(h,a), \caT^\pi_h Z_2(h,a)\right )\\
\le&~\gamma\mathop{\rm sup}_{h,a}d_p\left (P^\pi_h Z_1(h,a),P^\pi_h Z_2(h,a)\right )\\
=&~\gamma\bar{d}_p(Z_1, Z_2)~.
\end{aligned}
\end{equation}
\end{proof}
\end{theorem}

\subsection{Proof of Theorem \ref{theo:risk-policy-improvement}}\label{ap:risk-policy-improvement-theorem}
\begin{theorem}[Policy Improvement for $\caT^*_{h,\beta}$]
For two deterministic policies $\pi$ and $\pi^\prime$, if $\pi^\prime$ is obtained by $\caT^{*}_{h,\beta}$:
\begin{equation} 
\pi^\prime(h_{t})=\mathop{\arg\max}_{a\in\mathcal{A}}~\beta\left[Z^\pi(h_{t},a)\right]\nonumber~,
\end{equation}
then the following inequality holds
\begin{equation}
\beta\left[Z^\pi(h_{t},\pi(h_{t}))\right]\le
\beta\left[Z^{\pi^\prime}(h_{t},\pi^\prime(h_{t}))\right]\nonumber~.
\end{equation}

\begin{proof}
As $\pi^\prime$ is a greedy policy w.r.t. $\beta\left[Z^\pi\right]$, we have
\begin{equation}
\beta\left[Z^\pi(h_{t},\pi(h_{t}))\right]\le
\beta\left[Z^{\pi}(h_{t},\pi^\prime(h_{t}))\right]\nonumber~.
\end{equation}
Since $\pi^\prime$ is a deterministic policy, we can denote $a_t^\prime=\pi^\prime(h_{t})$, and unfold $Z^\pi(h_{t},a_t^\prime)$, we have
\begin{equation}
\begin{aligned}
Z^\pi(h_{t},a_t^\prime)&=
R_{0:t-1}+\gamma^t R(s_t,a_t^\prime)\\
&+\gamma^{t+1} Z^\pi\left(\{\tilde{s}_{t+1}\},\pi(\cdot|h_{t}\cup\{a_t^\prime,\tilde{s}_{t+1}\})\right)\\
&=Z^\pi\left(h_{t}\cup\{a_t^\prime,\tilde{s}_{t+1}\},\pi(\cdot|h_{t}\cup\{a_t^\prime,\tilde{s}_{t+1}\})\right)\nonumber~,
\end{aligned}
\end{equation}
where $\tilde{s}_{t+1}=M(s_t,a_t^\prime)$. Denoting $\tilde{h}_{t+1}=h_{t}\cup\{a_t^\prime,\tilde{s}_{t+1}\}$, $\tilde{a}_{t+1}=\pi(\tilde{h}_{t+1})$ and $\tilde{a}_{t+1}^\prime=\pi^\prime(\tilde{h}_{t+1})$, we further have
\begin{equation}
\begin{aligned}
&Z^\pi(h_{t},a_t^\prime)=Z^\pi\left(\tilde{h}_{t+1},\tilde{a}_{t+1})\right)\\
&\beta\left[Z^\pi(\tilde{h}_{t+1},\tilde{a}_{t+1}))\right]\le
\beta\left[Z^\pi(\tilde{h}_{t+1},\tilde{a}_{t+1}^\prime)\right]\nonumber~.
\end{aligned}
\end{equation}
Putting them together, we obtain
\begin{equation}
\begin{aligned}
\beta\left[Z^\pi(h_{t},a_{t}))\right]
\le~&\beta\left[Z^\pi(h_{t},a_t^\prime)\right]\\
=~&\beta\left[Z^\pi\left(\tilde{h}_{t+1},\tilde{a}_{t+1})\right)\right]\\
\le~&\beta\left[Z^\pi(\tilde{h}_{t+1},\tilde{a}_{t+1}^\prime)\right]\\
=~&\beta\left[Z^\pi(\tilde{h}_{t+1}\cup\{a_{t+1}^\prime,\tilde{s}_{t+2}\},\tilde{a}_{t+2}^\prime)\right]\\
\le~&\cdots\cdots\\
\le~&\beta\left[Z^{\pi^\prime}(h_{t},a_t^\prime)\right]\\
\end{aligned}
\end{equation}
\end{proof}
\end{theorem}

\subsection{Proof of \eq{eq:hr-optimal-equation} implying \eq{eqn:risk-sensitive-policy-form2}}
\label{ap:optimal-operator}
\begin{proof}
    We only consider episodic tasks\footnote{Note such tasks can also be represented in a uniform notation of infinite horizon by adding an absorbing state after termination states, see \citet{suttonreinforcement}.}, where each episode terminates at a certain termination state, say, at time step $T+1$. Hereby, we show we can obtain \eq{eqn:risk-sensitive-policy-form2} given \eq{eq:hr-optimal-equation} by induction:
    \begin{itemize}
        \item Considering timestep $T$, where $\forall\pi,~Z^\pi(h_T,a_T) = R_{0:T}$, \eq{eqn:risk-sensitive-policy-form2} holds, i.e., $\beta[Z^*(h_T,a_T)]=\beta[Z^{\pi^*}(h_T,a_T)]$.
        \item Given that \eq{eqn:risk-sensitive-policy-form2} holds at timestep $t+1$, namely
        \begin{equation*}
            \beta[Z^*(h_{t+1},\cdot)]=\beta[Z^{\pi^*}(h_{t+1},\cdot)]~.
        \end{equation*}
        Consider \eq{eq:hr-optimal-equation} at timestep $t$:
        \begin{align*}
            \beta[Z^*(h_t,a_t)] &= \beta\left[R_{0:t}+\gamma^{t+1} Z^*(\{s_{t+1}\},a^*_{t+1})\right]\\
            &= \beta\left[Z^*(h_t\cup\{a_t,s_{t+1}\},a^*_{t+1})\right]\\
            &= \beta\left[Z^{\pi^*}(h_t\cup\{a_t,s_{t+1}\},a^*_{t+1})\right]\\
            &= \beta[Z^{\pi^*}(h_t,a_t)]~.
        \end{align*}
        This indicates \eq{eqn:risk-sensitive-policy-form2} still holds at timestep $t$.
        \item Given the two statements, \eq{eq:hr-optimal-equation} is sufficient for \eq{eqn:risk-sensitive-policy-form2}.
    \end{itemize}
\end{proof}

\subsection{Proof of Theorem \ref{theo:risk-value-iteration}}\label{ap:risk-value-iteration}
\begin{theorem}
For $Z_1, Z_2 \in \caZ_h$,
\operator~optimality operator $\caT^{*}_{h,\beta}$ has the following property:
\begin{equation}
    \|\beta[\caT^*Z_{1}]-\beta[\caT^*Z_{2}]\|_\infty\le\|\beta[Z_{1}]-\beta[Z_{1}]\|_\infty~,
\end{equation}
\begin{proof}
$\forall s\in \caS, a \in \caA, 
s'=M(s,a),$ 
\begin{equation}
\begin{aligned}\label{eq:hr-optimal-contraction}
&\|\beta\left[\caT_{\beta}^*Z_1(h,a)\right]-\beta\left[\caT_{\beta}^*Z_2(h,a)\right]\|_{\infty}\\
=~&\|\max_{a'}\beta\left[R_{0:t}+\gamma^{t+1} Z_1(\{s'\},a')]\right]\\
&~~~~-\max_{a'}\beta\left[R_{0:t}+\gamma^{t+1} Z_2(\{s'\},a')]\right]\|_\infty\\
\le~&\|\beta\left[R_{0:t}+\gamma^{t+1} Z_1(\{s'\},a')]\right]\\
&~~~~-\beta\left[R_{0:t}+\gamma^{t+1} Z_2(\{s'\},a')]\right]\|_\infty\\
=~&\|\beta\left[Z_1(h,a)\right]-\beta\left[Z_2(h,a)\right]\|_{\infty}
\end{aligned}
\end{equation}
\end{proof}
\end{theorem}

\section{Implementation Details}
\label{ap:implementation}
\subsection{Practical Historical Value Function}
\label{ap:rnn-value}
We propose to involve the history information in \se{sec:tql}, which enables the learning of historical value distribution. In practice, the summarization of history information can be achieved via any sequence model, e.g. GRU, LSTM, and Transformer.

For our experiments on both MiniGrid and MountainCar tasks, we use a single-layer GRU network for simplicity. We first encode the observations $\{s_t\}$ and $\{a_t\}$ into representations of the same dimension $d$ with two encoders ${\rm enc}_{s}:\caS\rightarrow\bbR^d$ and ${\rm enc}_{a}:\caA\rightarrow\bbR^d$. Then we concatenate the representations ${\rm enc}_{s}(s_t)$ and ${\rm enc}_{a}(a_t)$ into $[{\rm enc}_{s}(s_t),{\rm enc}_{a}(a_t)]\in\bbR^{2d}$, and stack the representations until current timestep as the input for the GRU network.

The GRU network produces the summarization of history $h_t$ as 
\begin{equation*}
    repr_t={\rm GRU}\left(\left[[{\rm enc}_{s}(s_0),{\rm enc}_{a}(a_0)],\cdots,[{\rm enc}_{s}(s_t),{\rm enc}_{a}(a_t)]\right]\right)
\end{equation*}
and we finally model the historical value distribution and history-dependent policy via MLPs with two hidden layers, taking $repr_t$ and $a_t$ as input. It is worth noting that, as conditioning the policy and value function on history will inflate the searching space, we use a rolling window of fixed length $L=10$ on the history trajectory when computing $repr_t$ on continuous Mountain-Car tasks, i.e.
\begin{equation*}
    repr_t={\rm GRU}\left(\left[[{\rm enc}_{s}(s_{t-L+1}),{\rm enc}_{a}(a_{t-L+1})],\cdots,[{\rm enc}_{s}(s_t),{\rm enc}_{a}(a_t)]\right]\right)~,
\end{equation*}
which will significantly reduce the computation cost and complexity.

\subsection{Hyperparameters}

We list the hyperparameters of TQL for discrete MiniGrid tasks and continuous Mountain-Car tasks in \tb{tab:hyper-param-discrete} and \tb{tab:hyper-param-continuous} respectively.

\begin{minipage}[t]{0.48\textwidth}
\begin{table}[H]
\centering
\caption{Hyperparameters on Discrete MiniGrid}
\label{tab:hyper-param-discrete}
\begin{tabular}{c|c}
\toprule
$\epsilon_{\rm init}$ & 0.25\\
$\epsilon_{\rm final}$ & 0.001\\
$\epsilon_{\rm test}$ & 0.0\\
Buffer size & 3e5\\
Batch size & 32\\
Learning rate & 1e-3\\
$\gamma$ & 0.99\\
Sample size & 128\\
Online sample size & 64\\
Target sample size & 64\\
Hidden sizes & [512]\\
Start timesteps & 5000\\
Target update frequency & 500\\
\bottomrule
\end{tabular}
\end{table}
\end{minipage}\hfill
\begin{minipage}[t]{0.48\textwidth}
\begin{table}[H]
\centering
\caption{Hyperparameters on Continuous Mountain-Car}
\label{tab:hyper-param-continuous}
\begin{tabular}{c|c}
\toprule
Buffer size & 5e4\\
Batch size & 128\\
Policy lr & 3e-4\\
$\gamma$ & 0.99\\
Sample size & 32\\
Online sample size & 64\\
Target sample size & 64\\
Hidden sizes & [256, 256]\\
Start timesteps & 2.5e4\\
Soft update weight & 0.005\\
\bottomrule
\end{tabular}
\end{table}
\end{minipage}

\end{document}